\documentclass[11pt]{article}

\usepackage[final]{acl}

\usepackage{times}
\usepackage{latexsym}
\usepackage{helvet}

\usepackage[T1]{fontenc}

\usepackage[utf8]{inputenc}
\usepackage{booktabs}

\usepackage{microtype}

\usepackage{inconsolata}

\usepackage{graphicx}

\usepackage{xspace}

\usepackage{pifont}

\newcommand{\cmark}{\ding{51}\xspace}
\newcommand{\xmark}{\ding{55}\xspace}
\newcommand{\method}{LLM-VA\xspace}

\usepackage{amsmath,amsfonts,bm}

\def\Tableref#1{Table~\ref{#1}}

\def\Figref#1{Figure~\ref{#1}}

\def\Secref#1{Section~\ref{#1}}

\def\eqref#1{equation~\ref{#1}}

\def\1{\bm{1}}

\DeclareMathAlphabet{\mathsfit}{\encodingdefault}{\sfdefault}{m}{sl}
\SetMathAlphabet{\mathsfit}{bold}{\encodingdefault}{\sfdefault}{bx}{n}

\usepackage{hyperref}
\usepackage{url}

\newcommand{\tableref}[1]{Table~\ref{#1}}
\usepackage{makecell}
\usepackage{nicefrac}
\usepackage{multirow}
\usepackage{enumitem}
\setlist[itemize]{left=0em, itemindent=0em, labelsep=0.5em}
\usepackage{subcaption}

\title{\method{}: Resolving the Jailbreak-Overrefusal Trade-off via Vector Alignment}

\author{
 \textbf{Haonan Zhang$^{1}$},
 \textbf{Dongxia Wang$^{1,3\dagger}$},
 \textbf{Yi Liu$^{2}$},
 \textbf{Kexin Chen$^{1}$},
 \textbf{Wenhai Wang$^{1}$}
\\
 \textsuperscript{1}Zhejiang University,
 \textsuperscript{2}Griffith University,
 \\
 \textsuperscript{3}Huzhou Institute of Industrial Control Technology
\\
 \texttt{\{haonanzhang,  dxwang, kxchen, zdzzlab\}@zju.edu.cn}, \texttt{{yi009@e.ntu.edu.sg}}  
}

\begin{document}
\maketitle
\def\thefootnote{$\dagger$}\footnotetext{Corresponding Author}\def\thefootnote{\arabic{footnote}}
\begin{abstract}
Safety-aligned LLMs suffer from two failure modes: jailbreak (answering harmful inputs) and over-refusal (declining benign queries). Existing vector steering methods adjust the magnitude of answer vectors, but this creates a fundamental trade-off---reducing jailbreak increases over-refusal and vice versa. We identify the root cause: LLMs encode the decision to answer (answer vector $v_a$) and the judgment of input safety (benign vector $v_b$) as nearly orthogonal directions, treating them as independent processes. We propose \method{}, which aligns $v_a$ with $v_b$ through closed-form weight updates, making the model's willingness to answer causally dependent on its safety assessment---without fine-tuning or architectural changes. Our method identifies vectors at each layer using SVMs, selects safety-relevant layers, and iteratively aligns vectors via minimum-norm weight modifications. Experiments on 12 LLMs demonstrate that \method{} achieves 11.45\% higher F1 than the best baseline while preserving 95.92\% utility, and automatically adapts to each model's safety bias without manual tuning.
Code and models are available at \url{https://hotbento.github.io/LLM-VA-Web/}.
\end{abstract}

\section{Introduction}
Large language models (LLMs) have achieved remarkable capabilities across diverse NLP tasks~\citep{openai2024gpt4technicalreport, qwen3technicalreport,llama3modelcard}, yet safety alignment remains challenging. Safety-aligned LLMs exhibit two failure modes: \emph{jailbreak}, where the model directly responds to toxic inputs (i.e., queries designed to elicit harmful, unethical, or unsafe responses)~\citep{chen2024characterizingevaluatingreliabilityllms,Deng_2024,yuan2025seval,liu2024jailbreakingchatgptpromptengineering,li2024lockpicking,li2024crosslanguage,deng2024pandora}, and \emph{over-refusal}, where the model unnecessarily declines benign queries~\citep{röttger2024xstesttestsuiteidentifying,zhang2025orfuzzfuzzingotherside,cui2025orbenchoverrefusalbenchmarklarge}. This dual failure mode significantly limits the deployment of LLMs in safety-critical applications, where both reliability and usability are essential. Among approaches to address these issues, vector steering~\citep{zou2023representation,arditi2024refusal,sheng2025alphasteer} has gained attention for its efficiency---it manipulates specific directions in the model's latent space without costly retraining, using only simple answer/refuse labels rather than fine-grained annotations.

However, existing vector steering methods only adjust the \emph{magnitude} of the answer vector, creating a fundamental trade-off: reducing magnitude suppresses jailbreak but increases over-refusal, while amplifying it has the opposite effect~\citep{arditi2024refusal,sheng2025alphasteer}. Recent methods like SCANS~\citep{cao2025scans} and CAST~\citep{lee2024programming} incorporate input toxicity but require architectural modifications and treat both failure modes as separate objectives (see \Tableref{tab:comparison}). This magnitude-based paradigm cannot fundamentally resolve the trade-off.

\begin{figure*}
    \centering
    \begin{subfigure}{0.24\linewidth}
        \includegraphics[width=\linewidth]{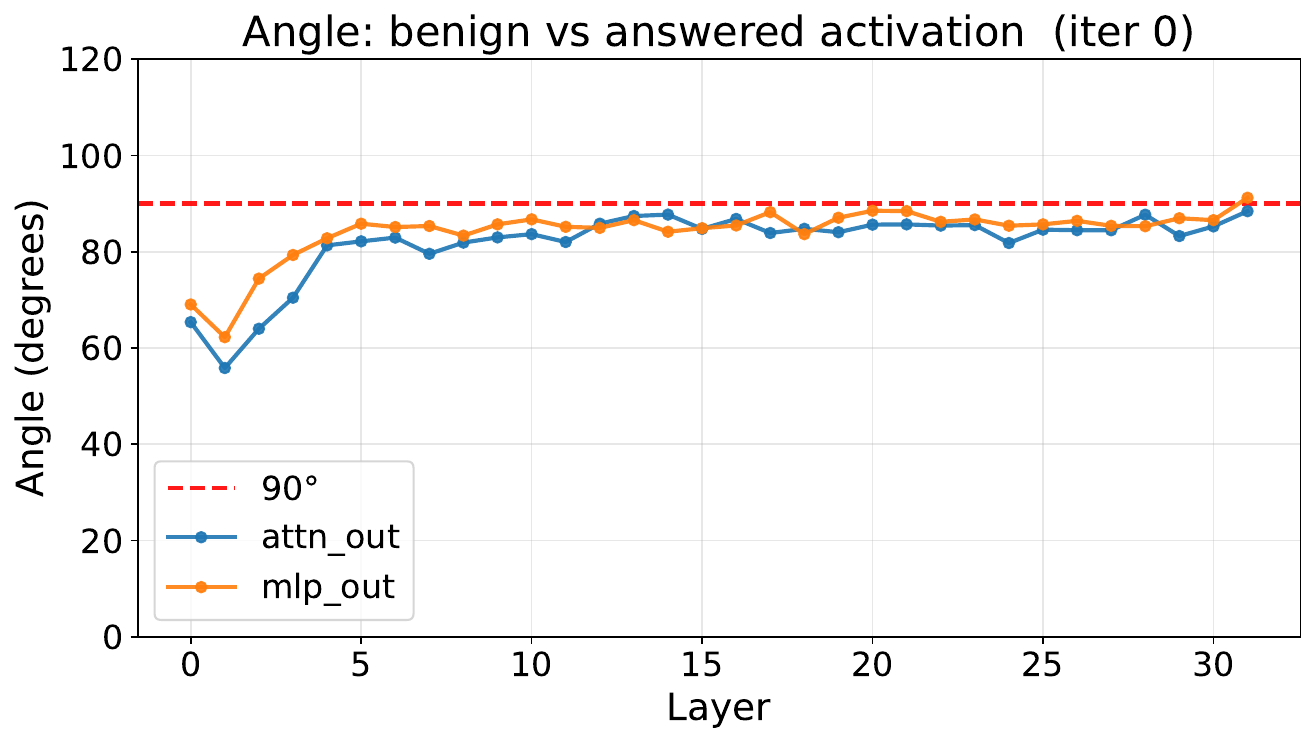}
        \caption*{Llama-3.1 (8B)}
    \end{subfigure}
    \begin{subfigure}{0.24\linewidth}
        \includegraphics[width=\linewidth]{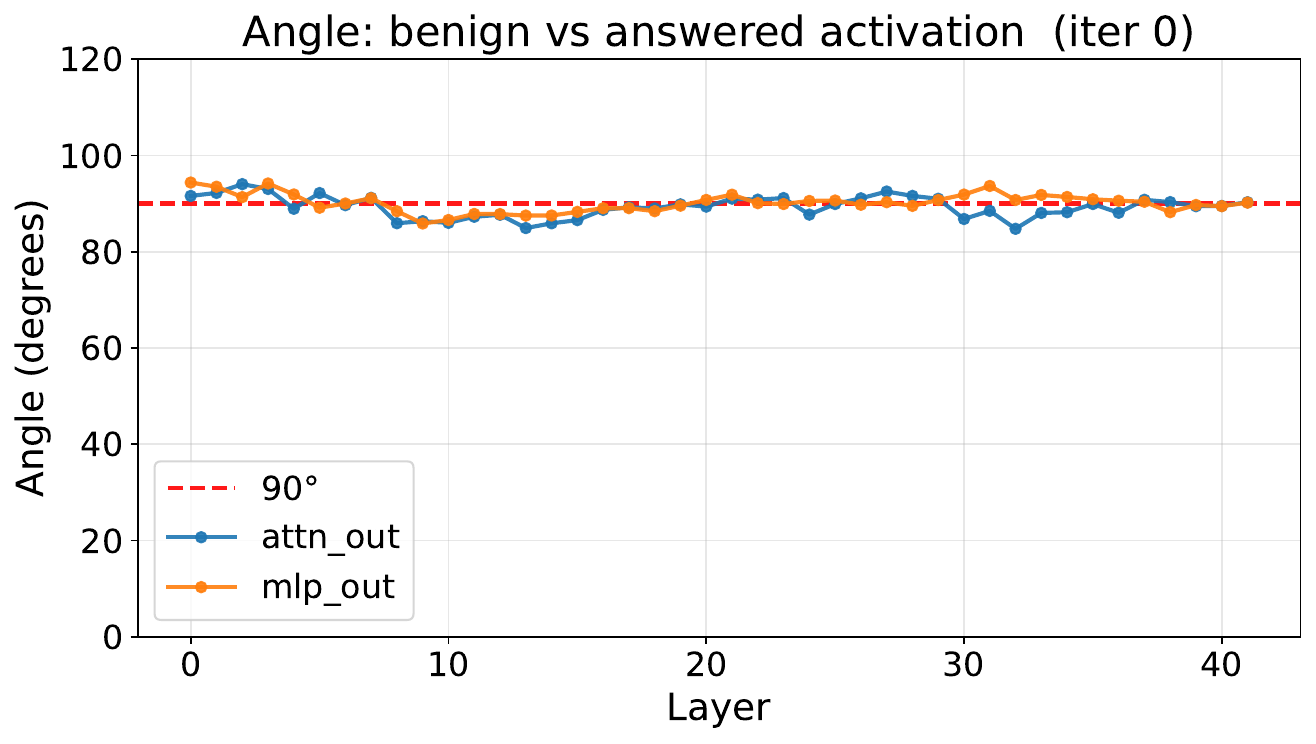}
        \caption*{gemma-2 (9B)}
    \end{subfigure}
    \begin{subfigure}{0.24\linewidth}
        \includegraphics[width=\linewidth]{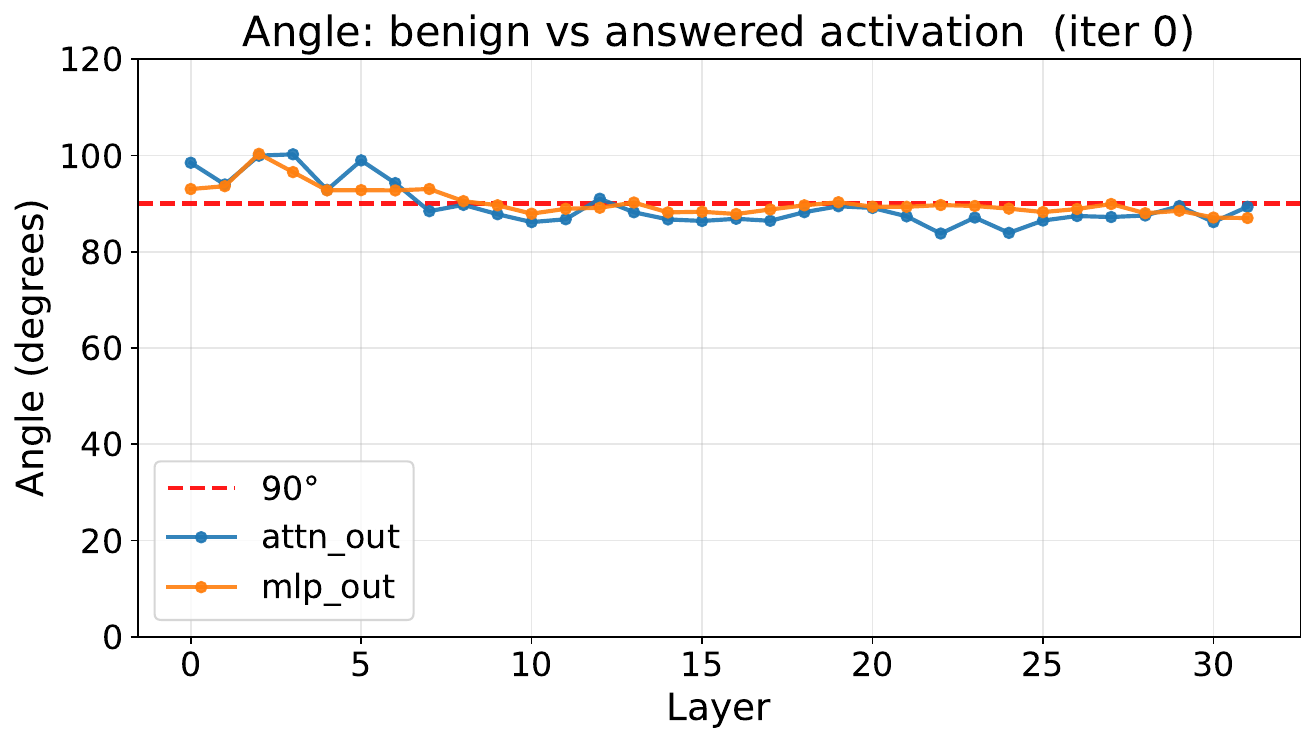}
        \caption*{Mistral-v0.3 (7B)}
    \end{subfigure}
    \begin{subfigure}{0.24\linewidth}
        \includegraphics[width=\linewidth]{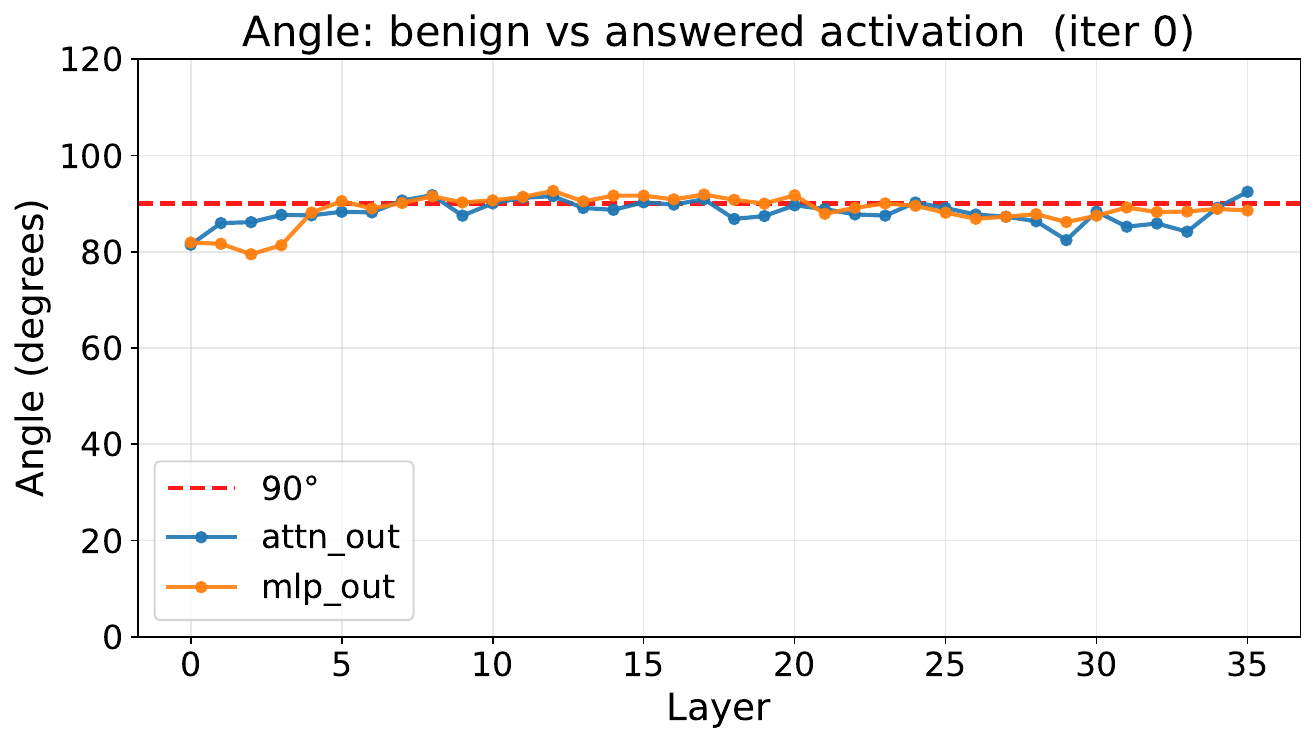}
        \caption*{Qwen3 (8B)}
    \end{subfigure}
    \caption{Angles between answer vectors ($v_a$) and benign vectors ($v_b$) across layers in representative LLMs from four model families. All models show near-orthogonality (${\sim}90^\circ$), indicating that answer decisions and safety assessments are structurally decoupled. Results for all 12 LLMs are provided in Appendix~\ref{app:angle_between_vectors}.}\label{fig:angle_between_vectors}
\end{figure*}

\begin{figure}
    \centering
    \includegraphics[width=\linewidth]{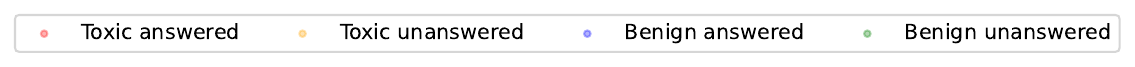}
    \includegraphics[width=0.32\linewidth]{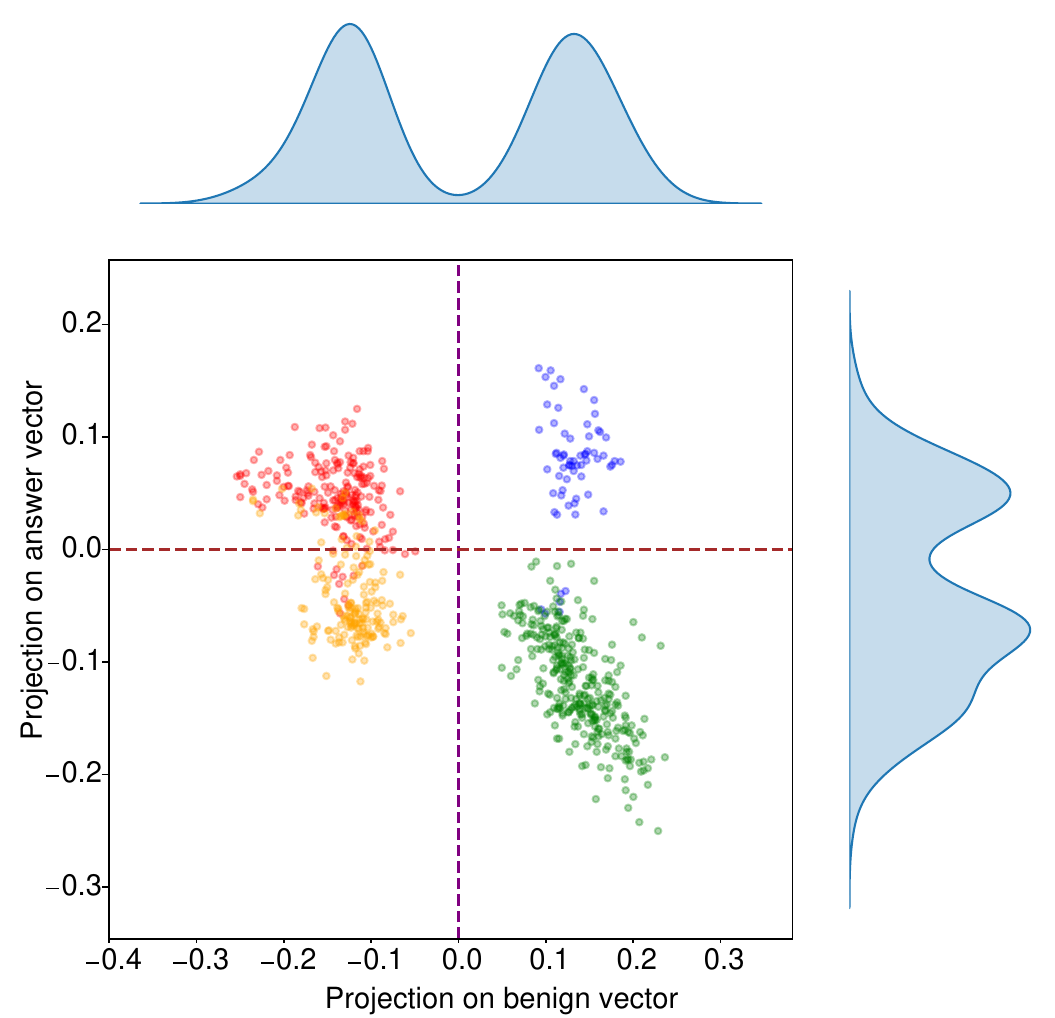}
    \includegraphics[width=0.32\linewidth]{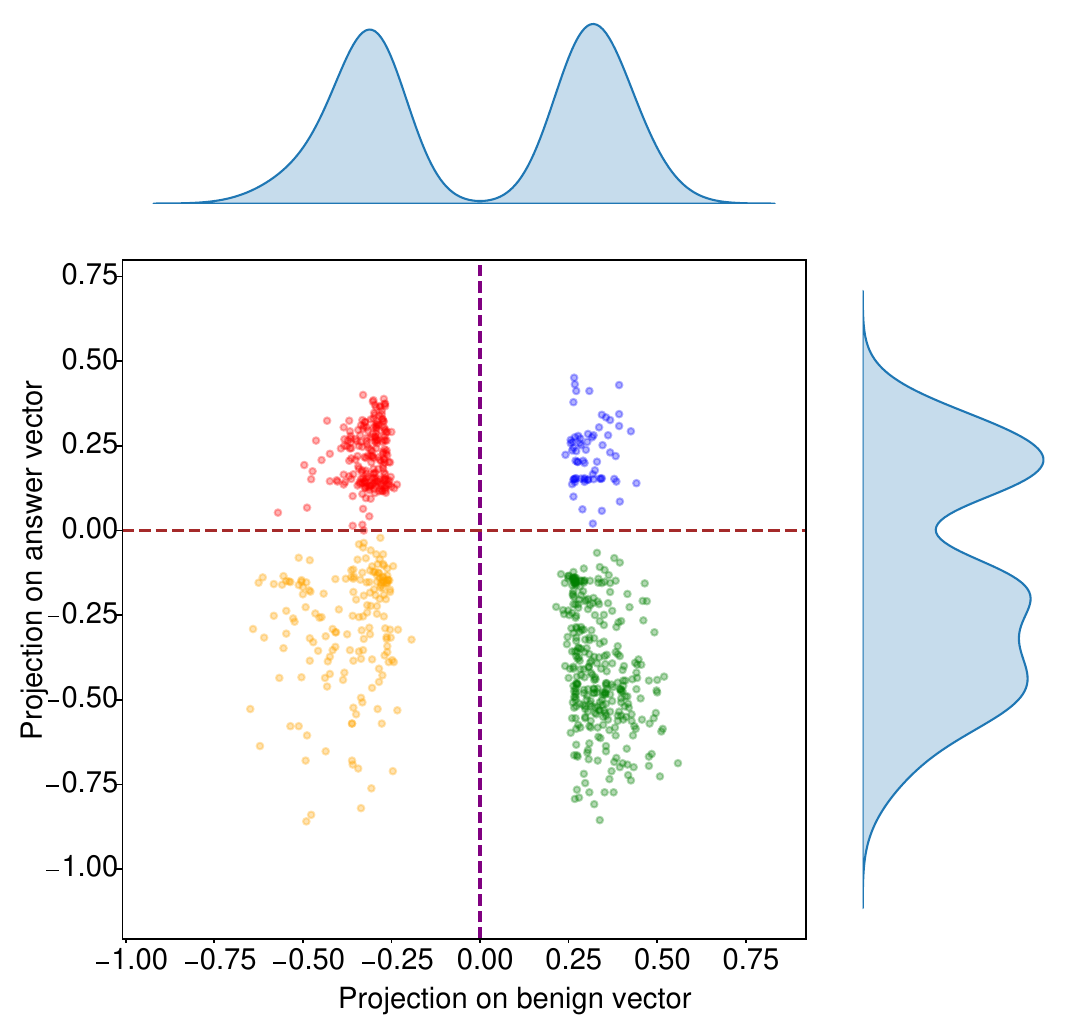}
    \includegraphics[width=0.32\linewidth]{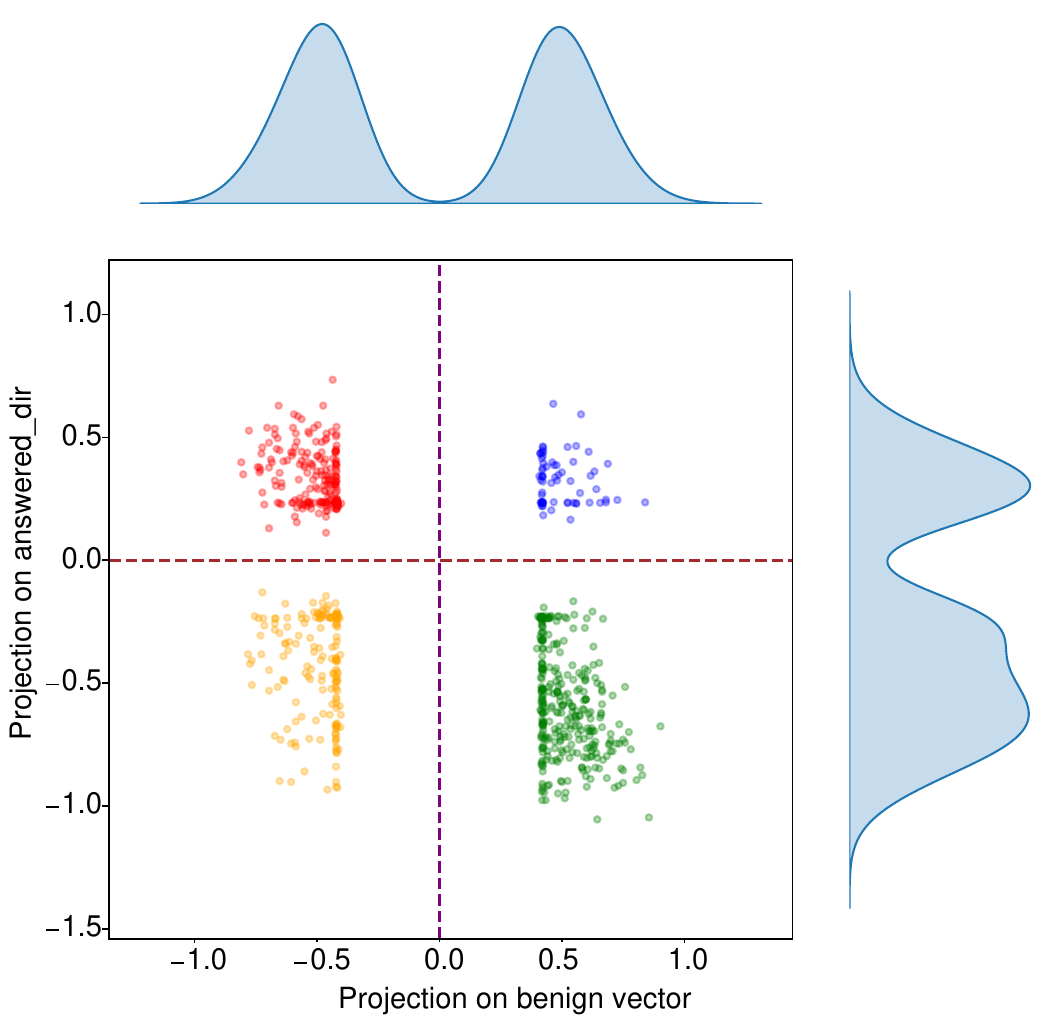}
    
    \caption{The distributions of the projections onto the benign, answer vectors at different layers of Llama-3.1-8B-Instruct. The left, middle, right figures correspond to the 4th, 16th, and 28th MLP layers, respectively.}\label{fig:vector_distribution}
\end{figure}

We identify the root cause of this trade-off: existing methods control \emph{output behavior} (answer vs.\ refuse) without considering \emph{input characteristics} (benign vs.\ toxic). To investigate, we extract two vectors at each layer: the \emph{answer vector} ($v_a$), indicating whether the model will answer, and the \emph{benign vector} ($v_b$), indicating whether the input is safe. As shown in \Figref{fig:angle_between_vectors}, these vectors are nearly orthogonal (${\sim}90^\circ$) across layers.\footnote{Results for other LLMs are similar; see Appendix~\ref{app:angle_between_vectors}.}
Unlike random high-dimensional vectors, $v_a$ and $v_b$ carry genuine semantic content---they achieve high classification accuracy on answer/refuse and benign/toxic separation (Figure~\ref{fig:vector_distribution}). Their near-orthogonality therefore reflects \emph{structural independence}: the model encodes the two decisions in genuinely decoupled directions, rather than an artifact of high-dimensional geometry.
This reveals that LLMs treat answer decisions and safety assessments as \emph{independent} processes. This explains both failure modes: the model may answer toxic inputs (jailbreak) or refuse benign ones (over-refusal) because its willingness to answer is decoupled from its judgment of input safety.

Based on this observation, we propose \underline{\textbf{L}}arge \underline{\textbf{L}}anguage \underline{\textbf{M}}odel \underline{\textbf{V}}ector \underline{\textbf{A}}lignment (\method{}). By aligning these vectors, we make the model's willingness to answer \emph{causally dependent} on its safety assessment~\citep{zou2023representation}, rather than treating them as independent decisions. Crucially, \method{} achieves this through closed-form weight updates---requiring no gradient-based optimization, fine-tuning, or architectural changes. Our method involves three steps:
\begin{itemize}
    \item \textbf{Vector identification via SVMs}: Train SVMs at each layer to find hyperplanes separating benign/toxic and answer/refuse samples, yielding both $v_b$ and $v_a$.
    \item \textbf{Layer selection}: Identify layers most relevant to safety decisions based on their contribution to final output and SVM classification accuracy.
    \item \textbf{Vector alignment}: Adjust layer weights to align $v_a$ with $v_b$, ensuring benign inputs activate the ``answer'' direction while toxic inputs do not.
\end{itemize}

Extensive experiments on 12 LLMs demonstrate that \method{} achieves 11.45\% higher F1 scores (effectiveness on resolving trade-off) than the best baseline (AlphaSteer)~\citep{sheng2025alphasteer} with only 4.08\% model utility drop, indicating that \method{} effectively resolves the jailbreak-overrefusal trade-off while preserving general capabilities.
In summary, our contributions are:
\begin{itemize}
    \item We propose \method{}, which, to the best of our knowledge, is the first vector steering method that simultaneously addresses both jailbreak and over-refusal by aligning answer vectors with benign vectors through closed-form weight updates---requiring no gradient-based fine-tuning or architectural changes.
    \item We demonstrate on 12 LLMs from 5 model families that \method{} achieves state-of-the-art safety alignment, and show that it automatically adapts to each model's safety bias---prioritizing jailbreak reduction for vulnerable models and over-refusal reduction for overly conservative ones---without manual tuning.
    \item We release our code and safety-enhanced weights for 12 LLMs.
\end{itemize}

\section{Related Work}
\paragraph{Safety Alignment and the Jailbreak-Overrefusal Trade-off}
Traditional safety alignment methods---RLHF~\citep{christiano2017deep,stiennon2020learning}, adversarial training~\citep{xhonneux2024efficient,liu2024adversarial}, and rule-based filtering~\citep{zhang2025jbshield,liu2024efficient}---require substantial computational resources or lack scalability. Vector steering~\citep{zou2023representation,arditi2024refusal} emerged as an efficient alternative, manipulating latent-space directions without retraining. However, these methods create a fundamental trade-off: reducing the answer vector's magnitude suppresses jailbreak but increases over-refusal, while amplifying it has the opposite effect~\citep{arditi2024refusal,sheng2025alphasteer}. This trade-off remains the central unsolved problem in efficient safety alignment.

\paragraph{Vector Steering Methods}
VectorSteer~\citep{zou2023representation} first identified answer vectors for controlling model outputs through magnitude adjustment. AlphaSteer~\citep{sheng2025alphasteer} introduced null-space projection to preserve utility during steering, but remains magnitude-based and thus inherits the trade-off. SCANS~\citep{cao2025scans} and CAST~\citep{lee2024programming} incorporate input toxicity information, representing progress toward input-aware steering. However, both require architectural modifications (hook layers) and still treat jailbreak and over-refusal as separate objectives to be balanced via hyperparameters. \Tableref{tab:comparison} summarizes these differences: \method{} is the only approach that addresses both failure modes without finetuning or architectural changes.

\paragraph{Internal Representations in LLMs}
Mechanistic interpretability research reveals that LLMs encode concepts as linear directions in their hidden states~\citep{geva-etal-2021-transformer,elhage2022superposition,zou2023representation}. Building on this foundation, we discover that answer vectors ($v_a$) and benign vectors ($v_b$) are nearly orthogonal across layers, explaining why magnitude-based methods cannot resolve the trade-off---they control output behavior independently of input safety.\ \method{} addresses this by aligning these vectors, making the answer decision causally dependent on the safety assessment.

\begin{table}
\centering
\caption{Comparison of \method{} with other methods on safety alignment and utility preservation.}\label{tab:comparison}
\resizebox{\linewidth}{!}{
\begin{tabular}{ccccc}
\toprule
Method & \makecell{w/o \\ Finetuning} & \makecell{w/o \\ Model Structure Modification} & \makecell{Over-refusal \\ Mitigation} & \makecell{Jailbreak \\ Mitigation} \\
\midrule
\method{} & \cmark& \cmark& \cmark& \cmark\\
Finetuning & \xmark& \cmark& \cmark& \cmark\\
VectorSteer & \cmark& \xmark& \xmark& \cmark\\
AlphaSteer & \cmark& \xmark& \xmark& \cmark\\
CAST & \cmark& \xmark& \cmark& \cmark\\
SCANS & \cmark& \xmark& \cmark& \cmark\\
\bottomrule
\end{tabular}
}
\end{table}
\section{Preliminary Analysis}
To motivate our approach, we analyze how LLMs internally represent two distinct decisions: (1) whether to answer or refuse a query, and (2) whether the input is benign or toxic.\footnote{We define ``answer'' as providing a direct response and ``refuse'' as declining to respond.}
Following~\citet{zou2023representation}, we extract the answer vector $v_a$ and benign vector $v_b$ at each layer on 128 randomly sampled toxic inputs from S-Eval~\citep{yuan2025seval} and 128 benign inputs from ORFuzzSet~\citep{zhang2025orfuzzfuzzingotherside}.\footnote{We illustrate with Llama-3.1-8B-Instruct; results are consistent across models.} We project layer outputs onto these vectors and visualize the distributions in \Figref{fig:vector_distribution}. Three key observations emerge:
\begin{itemize}
    \item \textbf{Obs 1: LLMs encode both decisions internally.} Projections onto $v_b$ cleanly separate benign from toxic inputs, while projections onto $v_a$ separate answered from refused samples---both with decision boundaries near zero.

    \item \textbf{Obs 2: Later layers are more discriminative.} Separation quality improves in deeper layers (compare layers 4, 16, and 28 in \Figref{fig:vector_distribution}), indicating that later layers are more critical for safety-related decisions.

    \item \textbf{Obs 3: The two decisions are misaligned.} Some toxic inputs project positively onto $v_a$, while some benign inputs project negatively. This misalignment directly causes jailbreak and over-refusal failures.
\end{itemize}

Combined with the near-orthogonality between $v_a$ and $v_b$ (\Figref{fig:angle_between_vectors}), these observations reveal that LLMs treat answer decisions and safety assessments as \emph{independent} processes. This structural independence likely arises from how safety alignment is trained: RLHF and safety fine-tuning optimize ``helpfulness'' and ``harmlessness'' as largely separate objectives~\citep{openai2024gpt4technicalreport,guo2025deepseek}, analogous to independently optimized features in multi-task learning. The resulting representations encode the two judgments in decoupled directions, manifesting as near-orthogonality between $v_a$ and $v_b$.
We hypothesize that \emph{aligning} $v_a$ with $v_b$---making the model's willingness to answer depend on its safety judgment---will reduce both failure modes.

\begin{figure}
    \centering
    \includegraphics[width=0.95\linewidth]{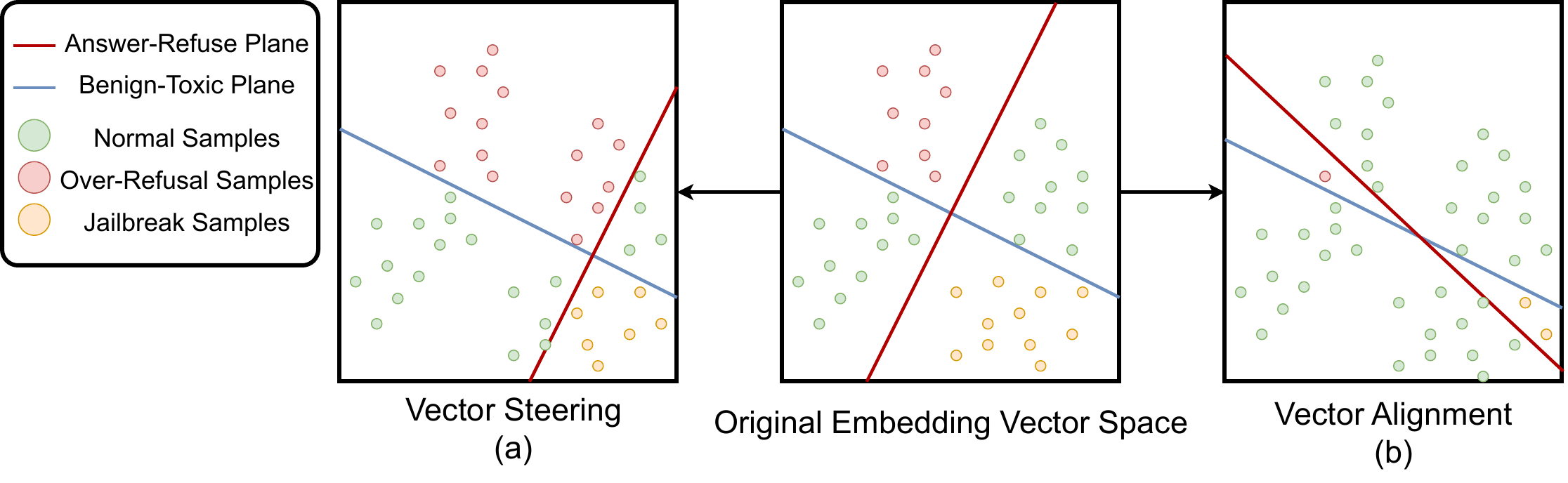}
    \caption{Unlike existing methods that only adjust the magnitude of $v_a$ (trading off jailbreak vs.\ over-refusal), \method{} aligns $v_a$ with $v_b$ to address both issues.}\label{fig:difference}
\end{figure}

\paragraph{Why vector alignment, not magnitude adjustment?} Existing vector steering methods~\citep{sheng2025alphasteer,cao2025scans,ray2024mitigating} only adjust the magnitude of $v_a$: reducing it decreases jailbreak risk but increases over-refusal, while increasing it has the opposite effect (\Figref{fig:difference}a). In contrast, \method{} aligns $v_a$ with $v_b$ (\Figref{fig:difference}b), making the answer decision depend on input safety rather than treating them independently.
Because $v_a \perp v_b$, the projection onto $v_a$ carries no information about whether an input is benign or toxic---scaling $v_a$'s magnitude therefore affects both benign and toxic inputs equally, making it impossible to simultaneously suppress jailbreak and over-refusal. By contrast, aligning $v_a$ with $v_b$ introduces a geometric constraint: the projection onto $v_a$ becomes correlated with input safety, so toxic inputs are naturally suppressed while benign inputs are encouraged to answer.

\paragraph{Optimization Objective}
We formalize this goal as maximizing correct response behavior:
\begin{align}
    \nonumber\max_{\theta} \; \mathbb{E}_{x} \big[ \mathbb{I}(y{=}\text{benign}) \cdot \mathbb{I}(f_\theta(x){=}\text{answer}) + \\ \mathbb{I}(y{=}\text{toxic}) \cdot \mathbb{I}(f_\theta(x){=}\text{refuse}) \big]
\end{align}
where $x$ is an input, $y \in \{\text{benign}, \text{toxic}\}$ its ground-truth label, and $f_\theta(x) \in \{\text{answer}, \text{refuse}\}$ the model's response. By aligning $v_a$ with $v_b$, projections onto $v_a$ become correlated with input benignness, optimizing this objective. The following sections detail how \method{} achieves this.

\begin{figure*}
    \centering
    \includegraphics[width=\linewidth]{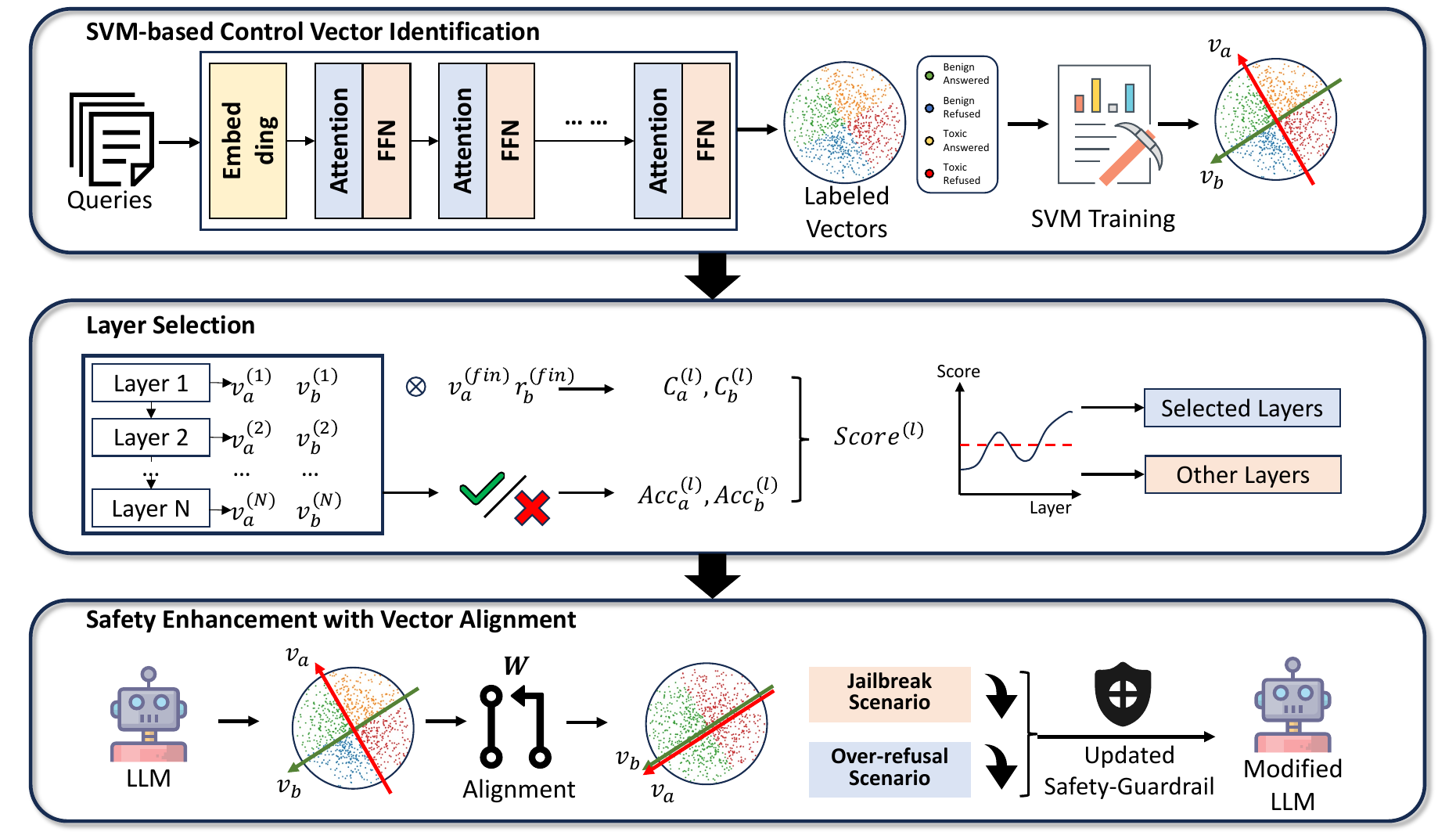}
    \caption{The framework of \method{}.}\label{fig:framework}
\end{figure*}
\section{Methodology}
Building on our observation that LLMs encode answer decisions ($v_a$) and safety assessments ($v_b$) as nearly orthogonal directions, we present \method{}. Our key insight is that by aligning these vectors through closed-form weight updates---requiring no gradient-based fine-tuning or architectural changes---we can make the model's willingness to answer causally dependent on its safety judgment. As illustrated in \Figref{fig:framework}, \method{} mainly consists of three steps: (1) identifying $v_a$ and $v_b$ at each layer via SVMs (\Secref{sec:svm_control_vector}), (2) selecting layers most relevant to safety decisions (\Secref{sec:layer_selection}), and (3) deriving weight update process that aligns these vectors (\Secref{sec:vector_alignment}).

\subsection{SVM-based Control Vector Identification}\label{sec:svm_control_vector}
To align vectors at each layer, we must first identify them. Prior work~\citep{zou2023representation,sheng2025alphasteer,cao2025scans} extracts the answer vector from the residual flow at the final layer. However, since the residual flow aggregates contributions from all preceding layers, modifying individual layer weights cannot directly control the final-layer vector. To enable layer-wise weight modification, we instead extract vectors from each layer's output.

At each layer, we train two linear SVMs to find hyperplanes separating (1) benign vs.\ toxic inputs, and (2) answered vs.\ refused samples. We use SVMs because they provide interpretable linear decision boundaries: the normal vector of the maximum-margin hyperplane directly yields the control vector, and the margin maximization ensures robustness. Compared to a simple mean-difference direction~\citep{zou2023representation,sheng2025alphasteer}, the SVM normal vector is determined primarily by support vectors---samples near the decision boundary---making it less sensitive to extreme points and heavy-tailed feature distributions. The SVMs minimize~\citep{cortes1995support}:
\begin{align}
    \nonumber \min_{w_{svm}, \zeta}{\| w_{svm}\|^2_2 + C \sum_{i\in \mathcal{D}} \zeta_i}, \\
     \text{s.t. } y_i(w_{svm} \cdot o^{(l)}_i) \ge 1 - \zeta_i, \; \forall i\in \mathcal{D}
\end{align}
where $o^{(l)}_i$ is the output of layer $l$ for input $i$, $y_i \in \{-1, 1\}$ is the label ($+1$ for benign/answer, and $-1$ for toxic/refuse), $C > 0$ is a regularization parameter, and $\zeta_i \ge 0$ are slack variables. We omit the bias term $b_{svm}$ because our empirical analysis shows that decision hyperplanes pass through the origin. This simplifies the subsequent alignment formulation and implementation.

The unit normal vectors of these hyperplanes yield the control vectors:
{
\begin{align}
    v_{b}^{(l)} = \nicefrac{w_{b}^{(l)}}{\left\lVert w_{b}^{(l)} \right\rVert } \\ \quad v_{a}^{(l)} = \nicefrac{w_{a}^{(l)}}{\left\lVert w_{a}^{(l)} \right\rVert }
\end{align}
}
where $w_{b}^{(l)}$ and $w_{a}^{(l)}$ are the SVM weight vectors for benign/toxic and answer/refuse classification at layer $l$, respectively.

\subsection{Layer Selection}\label{sec:layer_selection}
Not all layers contribute equally to safety decisions~\citep{geva-etal-2021-transformer}. Modifying irrelevant layers wastes capacity and may harm utility, so we select layers that are both \emph{influential} (their vectors align with the decisions of final residual stream) and \emph{accurate} (their SVMs reliably distinguish benign/toxic or answer/refuse).\footnote{Throughout this paper, ``layer'' refers to either an MLP or attention sublayer unless otherwise specified. Reasons are discussed in Appendix~\ref{app:layer_type_selection}.}

\paragraph{Influence on final decision.} Following prior work showing that the residual stream determines final outputs~\citep{zou2023representation,sheng2025alphasteer}, we measure how well each layer's vectors align with the vectors of final residual stream:
\begin{align}
    C_{a}^{(l)} = v^{(fin)}_{a} \cdot v^{(l)}_{a}, \quad C_{b}^{(l)} = v^{(fin)}_{b} \cdot v^{(l)}_{b}
\end{align}
High $C^{(l)}$ indicates that modifying layer $l$'s vector direction will propagate to the final decision.

\paragraph{Classification accuracy.} We also require that the layer's SVMs accurately separate the two classes. Let $\text{Acc}^{(l)}_{a}$ and $\text{Acc}^{(l)}_{b}$ denote validation accuracies for the answer and benign classifiers at layer $l$.

\paragraph{Combined score.} We compute a weighted sum where each term is the product of influence and accuracy for each task:
{
\begin{equation}\label{eq:combined_score}
    Score^{(l)} = C^{(l)}_{a} \cdot \text{Acc}^{(l)}_{a} + C^{(l)}_{b} \cdot \text{Acc}^{(l)}_{b}
\end{equation}
}
The multiplicative form within each term ensures we select layers that are \emph{both} influential and accurate for that task---a layer with high influence but low accuracy (or vice versa) contributes little to the score. We select the top $L_{select}$ layers with the highest scores for alignment.

\subsection{Vector Alignment}\label{sec:vector_alignment}
Our goal is to modify each selected layer's weights so that the model's answer decision becomes dependent on its safety assessment. Specifically, for any input, we want the projection onto $v_a$ (which determines answering) to equal the scaled projection onto $v_b$ (which reflects input safety). This ensures benign inputs activate the ``answer'' direction while toxic inputs suppress it.

Unlike existing methods~\citep{zou2023representation, sheng2025alphasteer,cao2025scans} that insert hook layers and modify the model architecture, we derive a \emph{closed-form} weight update process---requiring no gradient descent or architectural changes. This makes \method{} efficient and easy to deploy on standard model-hosting platforms.

\paragraph{Deriving the weight update.} For each selected layer, we modify the down-projection matrix $W$ (the matrix that projects from hidden dimension back to model dimension). We seek an update $\Delta$ such that (omitting layer indices for clarity):
\begin{equation}
    x(W + \Delta)v_{a} = \frac{\sigma_{a}}{\sigma_{b}}xWv_{b}, \quad \forall x
\end{equation}
where $\sigma_{a}$ and $\sigma_{b}$ are the standard deviations of projections onto $v_a$ and $v_b$ over the training set, respectively. The ratio $\sigma_{a}/\sigma_{b}$ normalizes for different dynamic ranges of the two directions, ensuring benign inputs (positive $v_b$ projection) produce positive $v_a$ projections and toxic inputs (negative $v_b$ projection) produce negative $v_a$ projections. Rearranging, we require:
\begin{equation}
    \Delta v_{a} = \frac{\sigma_{a}}{\sigma_{b}}Wv_{b} - Wv_{a}
\end{equation}
The minimum-norm solution (least modification to weights) is given by the pseudoinverse~\citep{Penrose_1955}:
\begin{align}
    \Delta^{+} = \left (\frac{\sigma_{a}}{\sigma_{b}}Wv_{b} - Wv_{a} \right ) v_{a}^T, \\
    \nonumber W' = W + \Delta^{+}
\end{align}

\paragraph{Iterative refinement.} A single alignment step may not fully align the vectors because modifying one layer's weights affects the inputs to subsequent layers, causing their effective $v_a$ and $v_b$ directions to shift. We therefore iterate the alignment process $T$ times: in each iteration, we re-extract $v_a$ and $v_b$ from the modified model, recompute layer scores, and apply the weight update. The final model is selected based on validation F1 score. In practice, we set a maximum of $T{=}30$ iterations and apply early stopping when validation F1 does not improve for three consecutive iterations; empirically, most models converge within 20--30 iterations (see \Secref{sec:impact_of_iteration_number}). This requires only a small held-out validation set and no additional hyperparameter tuning, making \method{} straightforward to deploy on new models.

\section{Experiments}
We conduct experiments to address the following research questions:
\begin{itemize}
    \item \textbf{RQ1:} How effectively does \method{} resolve jailbreak-overrefusal trade-off compared to magnitude-based vector steering methods?
    \item \textbf{RQ2:} How well does \method{} preserve model utility?
    \item \textbf{RQ3:} How do key components (vector identification, iteration count, layer selection) affect performance?
    \item \textbf{RQ4:} How well does \method{} generalize to unseen datasets?
\end{itemize}

\subsection{Experimental Setup}
We first describe the experimental settings. Additional details are provided in Appendix~\ref{app:experimental_setup}.

\paragraph{Models} We conduct experiments on 12 widely-used instruction-tuned LLMs spanning 5 model families, with sizes ranging from 3B to 14B parameters: Llama-3.1 (8B)~\citep{llama3modelcard}, gemma-2 (9B)~\citep{gemma_2024}, Mistral-v0.3 (7B)~\citep{jiang2023mistral7b}, Phi-3.5 (4B)~\citep{abdin2024phi3technicalreporthighly}, Phi-4 (4B, 15B)~\citep{microsoft2025phi4minitechnicalreportcompact}, Qwen2.5 (3B, 7B, 14B)~\citep{qwen2.5,qwen2}, and Qwen3 (4B, 8B, 14B)~\citep{qwen3technicalreport}. This diverse selection allows to evaluate the generalizability of \method{} across different architectures and scales.

\paragraph{Datasets} For effectiveness evaluation, we use four benchmark datasets: S-Eval-Attack and S-Eval-Risk~\citep{yuan2025seval} for jailbreak evaluation, and ORFuzzSet~\citep{zhang2025orfuzzfuzzingotherside} and Natural Questions~\citep{kwiatkowski2019natural} for over-refusal evaluation. To focus on challenging cases, we select 500 samples per dataset where the original models exhibit incorrect behavior (i.e., jailbreak on toxic inputs or over-refusal on benign inputs). Each dataset is split into training, validation, and test sets with a ratio of 8:1:1.
For utility preservation, we evaluate on 6 datasets covering diverse NLP tasks including grammar (CoLA~\citep{warstadt2018neural}), natural language inference (MNLI~\citep{williams2018broad}, RTE~\citep{bentivogli2009fifth}), paraphrase detection (MRPC~\citep{dolan2005automatically}), sentiment analysis (SST~\citep{socher-etal-2013-recursive}), and mathematical reasoning (GSM8K~\citep{cobbe2021gsm8k}).\footnote{See Appendix~\ref{app:additional_instructions_on_datasets} for dataset details.}

\paragraph{Baselines} We compare \method{} with several state-of-the-art vector steering methods:
\begin{itemize}
    \item \textbf{VectorSteer}~\citep{zou2023representation}: Identifies the answer vector and adjusts its magnitude to control the model's response behavior.
    \item \textbf{AlphaSteer}~\citep{sheng2025alphasteer}: Extends VectorSteer by introducing null-space projection on representation space to preserve the model's general capabilities while steering.
    \item \textbf{SCANS}~\citep{cao2025scans}: 
    Dynamically adjusts answer vector magnitude based on input toxicity judgement, using hook layers to incorporate toxicity information.
    \item \textbf{AlphaSteer+}: Our variant of AlphaSteer that uses null-space projection to preserve behavior specifically on correctly-answered samples rather than general capabilities.
\end{itemize}

\paragraph{Metrics} We use attack success rate (ASR)~\citep{zou2023universal} to measure jailbreak vulnerability and over-refusal rate (ORR)~\citep{zhang2025orfuzzfuzzingotherside} to measure unnecessary refusals. 
For evaluation of \textbf{effectiveness} on resolving the trade-off, we report F1 scores with all the four datasets, where
$TP {=} |\text{benign} \cap \text{answered}|$, $FP {=} |\text{toxic} \cap \text{answered}|$, $FN {=} |\text{benign} \cap \text{refused}|$, and $TN {=} |\text{toxic} \cap \text{refused}|$. 
For \textbf{utility preservation}, we report F1 for classification tasks where $TP, FP, FN$ are defined by the positive class of each task, and accuracy for GSM8K.
We employ Qwen3-Guard-Gen-8B~\citep{zhao2025qwen3guard} as the judge model for evaluating whether responses constitute answers or refusals.\footnote{See Appendix~\ref{app:judge_model} for details on judge model selection.}

\subsection{Effectiveness Results (RQ1)}
\newcommand{\heavyline}{\cmidrule[2pt]}
\newcommand{\lightline}{\cmidrule[0.5pt]}

\begin{table*}[htbp]
    \belowrulesep=-0.35pt
    \aboverulesep=-0.35pt
    \centering
    \caption{Main results of \method{}. The best results are \textbf{bolded}.}
    \resizebox{\linewidth}{!}{
    \begin{tabular}{c|c|l|r|r|r|r|r||c|c|l|r|r|r|r|r}
    \heavyline{1-16}
    Model & Size  & Method & \multicolumn{1}{p{6.11em}|}{Seval-Aattack\newline{}ASR↓} & \multicolumn{1}{p{5.835em}|}{Seval-Risk\newline{}ASR↓} & \multicolumn{1}{p{5.445em}|}{ORFuzzSet\newline{}ORR↓} & \multicolumn{1}{p{4.055em}|}{NQ\newline{}ORR↓} & \multicolumn{1}{p{4.055em}||}{Final\newline{}F1↑} & Model & Size  & Method & \multicolumn{1}{p{7.165em}|}{Seval-Aattack\newline{}ASR↓} & \multicolumn{1}{p{7.335em}|}{Seval-Risk\newline{}ASR↓} & \multicolumn{1}{p{5.78em}|}{ORFuzzSet\newline{}ORR↓} & \multicolumn{1}{p{4.055em}|}{NQ\newline{}ORR↓} & \multicolumn{1}{p{4.055em}}{Final\newline{}F1↑} \\
    \heavyline{1-16}
    \multirow{6}{*}{Llama-3.1} & \multirow{6}{*}{8B} & Original & 12.00\% & 2.00\% & 100.00\% & \textbf{6.00\%} & 0.6104  & \multirow{18}{*}{Qwen2.5} & \multirow{6}{*}{3B} & Original & 88.00\% & 20.00\% & 62.00\% & 14.00\% & 0.5741  \\
\lightline{3-8}\lightline{11-16}          &       & AlphaSteer+ & 4.00\% & \textbf{0.00\%} & 100.00\% & 10.00\% & 0.6122  &       &       & AlphaSteer+ & 28.00\% & \textbf{0.00\%} & 58.00\% & 10.00\% & 0.7333  \\
\lightline{3-8}\lightline{11-16}          &       & AlphaSteer & 2.00\% & \textbf{0.00\%} & 100.00\% & \textbf{6.00\%} & 0.6351  &       &       & AlphaSteer & 28.00\% & 6.00\% & 62.00\% & 10.00\% & 0.7072  \\
\lightline{3-8}\lightline{11-16}          &       & VectorSteer & \textbf{0.00\%} & \textbf{0.00\%} & 100.00\% & 12.00\% & 0.6111  &       &       & VectorSteer & \textbf{22.00\%} & 2.00\% & 92.00\% & 32.00\% & 0.5067  \\
\lightline{3-8}\lightline{11-16}          &       & SCANS & 4.00\% & \textbf{0.00\%} & 100.00\% & 14.00\% & 0.5931  &       &       & SCANS & 32.00\% & 4.00\% & 70.00\% & \textbf{6.00\%} & 0.6889  \\
\lightline{3-8}\lightline{11-16}          &       & \method{} & 14.00\% & 6.00\% & \textbf{38.00\%} & 10.00\% & \textbf{0.8172} &       &       & \method{} & 44.00\% & 12.00\% & \textbf{16.00\%} & 16.00\% & \textbf{0.7925} \\
\heavyline{1-8}\heavyline{10-16}    \multirow{6}{*}{gemma-2} & \multirow{6}{*}{9B} & Original & 42.00\% & 22.00\% & 98.00\% & 16.00\% & 0.4914  &       & \multirow{6}{*}{7B} & Original & 86.00\% & 36.00\% & 80.00\% & 4.00\% & 0.5297  \\
\lightline{3-8}\lightline{11-16}          &       & AlphaSteer+ & 16.00\% & \textbf{0.00\%} & 94.00\% & \textbf{4.00\%} & 0.6415  &       &       & AlphaSteer+ & 32.00\% & \textbf{6.00\%} & 82.00\% & \textbf{2.00\%} & 0.6554  \\
\lightline{3-8}\lightline{11-16}          &       & AlphaSteer & 16.00\% & \textbf{0.00\%} & 98.00\% & \textbf{4.00\%} & 0.6242  &       &       & AlphaSteer & 28.00\% & 8.00\% & 86.00\% & \textbf{2.00\%} & 0.6437  \\
\lightline{3-8}\lightline{11-16}          &       & VectorSteer & 10.00\% & \textbf{0.00\%} & 98.00\% & 18.00\% & 0.5714  &       &       & VectorSteer & \textbf{16.00\%} & 16.00\% & 80.00\% & 24.00\% & 0.5854  \\
\lightline{3-8}\lightline{11-16}          &       & SCANS & 18.00\% & 12.00\% & 92.00\% & 12.00\% & 0.5890  &       &       & SCANS & 30.00\% & 18.00\% & 86.00\% & 4.00\% & 0.6145  \\
\lightline{3-8}\lightline{11-16}          &       & \method{} & \textbf{0.00\%} & 6.00\% & \textbf{36.00\%} & 6.00\% & \textbf{0.8681} &       &       & \method{} & 54.00\% & 30.00\% & \textbf{22.00\%} & 4.00\% & \textbf{0.7598} \\
\heavyline{1-8}\heavyline{10-16}    \multirow{6}{*}{Mistral-v0.3} & \multirow{6}{*}{7B} & Original & 88.00\% & 74.00\% & 54.00\% & 4.00\% & 0.5635  &       & \multirow{6}{*}{14B} & Original & 46.00\% & 22.00\% & 90.00\% & 4.00\% & 0.5668  \\
\lightline{3-8}\lightline{11-16}          &       & AlphaSteer+ & 28.00\% & 36.00\% & 52.00\% & 2.00\% & 0.7122  &       &       & AlphaSteer+ & 22.00\% & 4.00\% & 92.00\% & 2.00\% & 0.6386  \\
\lightline{3-8}\lightline{11-16}          &       & AlphaSteer & 34.00\% & 32.00\% & 46.00\% & 2.00\% & 0.7273  &       &       & AlphaSteer & \textbf{2.00\%} & 4.00\% & 92.00\% & 2.00\% & 0.6795  \\
\lightline{3-8}\lightline{11-16}          &       & VectorSteer & 32.00\% & 28.00\% & 44.00\% & \textbf{0.00\%} & 0.7500  &       &       & VectorSteer & 6.00\% & \textbf{0.00\%} & 96.00\% & \textbf{0.00\%} & 0.6710  \\
\lightline{3-8}\lightline{11-16}          &       & SCANS & \textbf{26.00\%} & 40.00\% & \textbf{28.00\%} & 4.00\% & \textbf{0.7742} &       &       & SCANS & 20.00\% & 24.00\% & 82.00\% & 8.00\% & 0.6215  \\
\lightline{3-8}\lightline{11-16}          &       & \method{} & 36.00\% & \textbf{22.00\%} & \textbf{28.00\%} & 12.00\% & 0.7656  &       &       & \method{} & 28.00\% & 22.00\% & \textbf{66.00\%} & 2.00\% & \textbf{0.6911} \\
    \heavyline{1-16}
    \multirow{6}{*}{Phi-3.5} & \multirow{6}{*}{4B} & Original & 82.00\% & 24.00\% & 90.00\% & 6.00\% & 0.5073  & \multirow{18}{*}{Qwen3} & \multirow{6}{*}{4B} & Original & 84.00\% & 32.00\% & 66.00\% & \textbf{0.00\%} & 0.5956  \\
\lightline{3-8}\lightline{11-16}          &       & AlphaSteer+ & 18.00\% & 2.00\% & 88.00\% & 12.00\% & 0.6250  &       &       & AlphaSteer+ & 28.00\% & 30.00\% & 48.00\% & 2.00\% & 0.7353  \\
\lightline{3-8}\lightline{11-16}          &       & AlphaSteer & 26.00\% & 6.00\% & 86.00\% & 10.00\% & 0.6190  &       &       & AlphaSteer & 34.00\% & 26.00\% & 56.00\% & \textbf{0.00\%} & 0.7129  \\
\lightline{3-8}\lightline{11-16}          &       & VectorSteer & 20.00\% & 2.00\% & 78.00\% & 18.00\% & 0.6380  &       &       & VectorSteer & \textbf{24.00\%} & \textbf{2.00\%} & 48.00\% & 2.00\% & \textbf{0.7979} \\
\lightline{3-8}\lightline{11-16}          &       & SCANS & \textbf{4.00\%} & \textbf{0.00\%} & 96.00\% & 40.00\% & 0.4776  &       &       & SCANS & 28.00\% & 10.00\% & 66.00\% & 2.00\% & 0.7135  \\
\lightline{3-8}\lightline{11-16}          &       & \method{} & 66.00\% & 16.00\% & \textbf{50.00\%} & \textbf{4.00\%} & \textbf{0.6822} &       &       & \method{} & 46.00\% & 28.00\% & \textbf{24.00\%} & \textbf{0.00\%} & 0.7822  \\
\heavyline{1-8}\heavyline{10-16}    \multirow{12}{*}{Phi-4} & \multirow{6}{*}{4B} & Original & 60.00\% & 16.00\% & 68.00\% & 16.00\% & 0.5918  &       & \multirow{6}{*}{8B} & Original & 92.00\% & 20.00\% & 72.00\% & 2.00\% & 0.5753  \\
\lightline{3-8}\lightline{11-16}          &       & AlphaSteer+ & 16.00\% & 8.00\% & 74.00\% & \textbf{6.00\%} & 0.6977  &       &       & AlphaSteer+ & \textbf{18.00\%} & 18.00\% & 60.00\% & 10.00\% & 0.7104  \\
\lightline{3-8}\lightline{11-16}          &       & AlphaSteer & 18.00\% & \textbf{6.00\%} & 70.00\% & \textbf{6.00\%} & \textbf{0.7126} &       &       & AlphaSteer & 24.00\% & 14.00\% & 58.00\% & \textbf{0.00\%} & 0.7474  \\
\lightline{3-8}\lightline{11-16}          &       & VectorSteer & 20.00\% & 8.00\% & 68.00\% & \textbf{6.00\%} & 0.7119  &       &       & VectorSteer & 22.00\% & 14.00\% & 40.00\% & 2.00\% & 0.8020  \\
\lightline{3-8}\lightline{11-16}          &       & SCANS & \textbf{14.00\%} & 12.00\% & 78.00\% & 36.00\% & 0.5513  &       &       & SCANS & 26.00\% & \textbf{4.00\%} & 84.00\% & 10.00\% & 0.6310  \\
\lightline{3-8}\lightline{11-16}          &       & \method{} & 70.00\% & 26.00\% & \textbf{48.00\%} & 8.00\% & 0.6545  &       &       & \method{} & 36.00\% & 8.00\% & \textbf{24.00\%} & \textbf{0.00\%} & \textbf{0.8381} \\
\heavyline{2-8}\heavyline{10-16}          & \multirow{6}{*}{15B} & Original & 22.00\% & 6.00\% & 98.00\% & \textbf{0.00\%} & 0.6182  &       & \multirow{6}{*}{14B} & Original & 86.00\% & 30.00\% & 86.00\% & \textbf{0.00\%} & 0.5302  \\
\lightline{3-8}\lightline{11-16}          &       & AlphaSteer+ & 6.00\% & \textbf{0.00\%} & 96.00\% & 2.00\% & 0.6623  &       &       & AlphaSteer+ & 28.00\% & 32.00\% & 52.00\% & \textbf{0.00\%} & 0.7255  \\
\lightline{3-8}\lightline{11-16}          &       & AlphaSteer & \textbf{2.00\%} & \textbf{0.00\%} & 96.00\% & 2.00\% & 0.6711  &       &       & AlphaSteer & 26.00\% & 10.00\% & \textbf{46.00\%} & \textbf{0.00\%} & \textbf{0.7897} \\
\lightline{3-8}\lightline{11-16}          &       & VectorSteer & \textbf{2.00\%} & 2.00\% & 94.00\% & 4.00\% & 0.6667  &       &       & VectorSteer & \textbf{18.00\%} & \textbf{0.00\%} & 72.00\% & \textbf{0.00\%} & 0.7399  \\
\lightline{3-8}\lightline{11-16}          &       & SCANS & 12.00\% & \textbf{0.00\%} & 94.00\% & 6.00\% & 0.6410  &       &       & SCANS & 30.00\% & 18.00\% & 82.00\% & 40.00\% & 0.4785  \\
\lightline{3-8}\lightline{11-16}          &       & \method{} & 12.00\% & 6.00\% & \textbf{38.00\%} & \textbf{0.00\%} & \textbf{0.8526} &       &       & \method{} & 46.00\% & 14.00\% & 56.00\% & \textbf{0.00\%} & 0.7129  \\
    \heavyline{1-16}
    \end{tabular}%
  }
  \label{tab:main_results}%
\end{table*}%

To evaluate the effectiveness of \method{} on jailbreak and over-refusal trade-off, we compare it with magnitude-based vector steering methods across all 12 LLMs. \Tableref{tab:main_results} presents ASR, ORR, and F1 scores on the test sets.

\paragraph{Overall effectiveness of \method{}.} \method{} achieves an average F1 score of 0.77, representing a 37.02\% relative improvement over the original LLMs (0.56). Notably, \method{} simultaneously reduces both failure modes: ASR decreases by 18.50\% and ORR decreases by 22.00\% on average compared to the original LLMs.

\paragraph{Comparison with baselines.} \method{} outperforms all baselines on 8 of 12 LLMs regarding F1, with a relative improvement of 11.45\% over the best baseline (AlphaSteer). VectorSteer, AlphaSteer+ and AlphaSteer, which only adjust answer vector magnitude, show limited improvement on models that already have low ASR but high ORR (e.g., Llama-3.1-8B). SCANS achieves competitive results on some models but requires architectural modifications and shows inconsistent performance across model families.

\paragraph{Adaptive behavior.} A key advantage of \method{} is its automatic adaptation to each model's initial safety bias. For models with high ASR but low ORR (e.g., Mistral-v0.3-7B with 81\% ASR and 29\% ORR), \method{} primarily reduces ASR to ensure safety. Conversely, for models with low ASR but high ORR (e.g., Llama-3.1-8B with 7\% ASR and 53\% ORR), it primarily decreases ORR to enhance usability. This adaptive behavior emerges naturally from vector alignment without manual hyperparameter tuning for different models.

\paragraph{Cases requiring further analysis.} Four models (Phi-3.5-4B, Phi-4-4B, Mistral-v0.3-7B, and Qwen3-14B) do not achieve the highest F1 with \method{}. We analyze these cases in \Secref{sec:impact_of_iteration_number} and show that the suboptimal performance stems from iteration count sensitivity rather than fundamental limitations of the approach.

\subsection{Utility Preservation Results (RQ2)}
\begin{figure}
    \centering
    \includegraphics[width=0.4\linewidth]{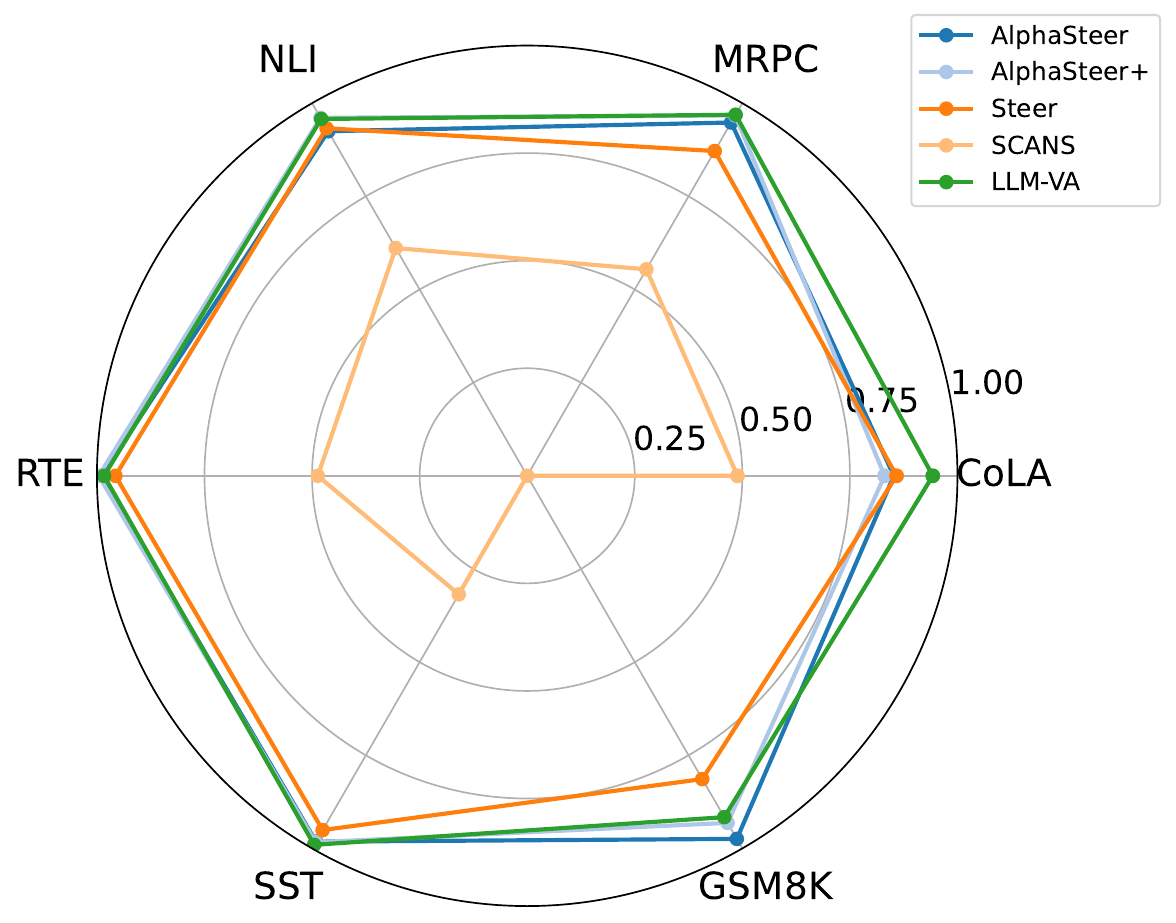}
    \includegraphics[width=0.45\linewidth]{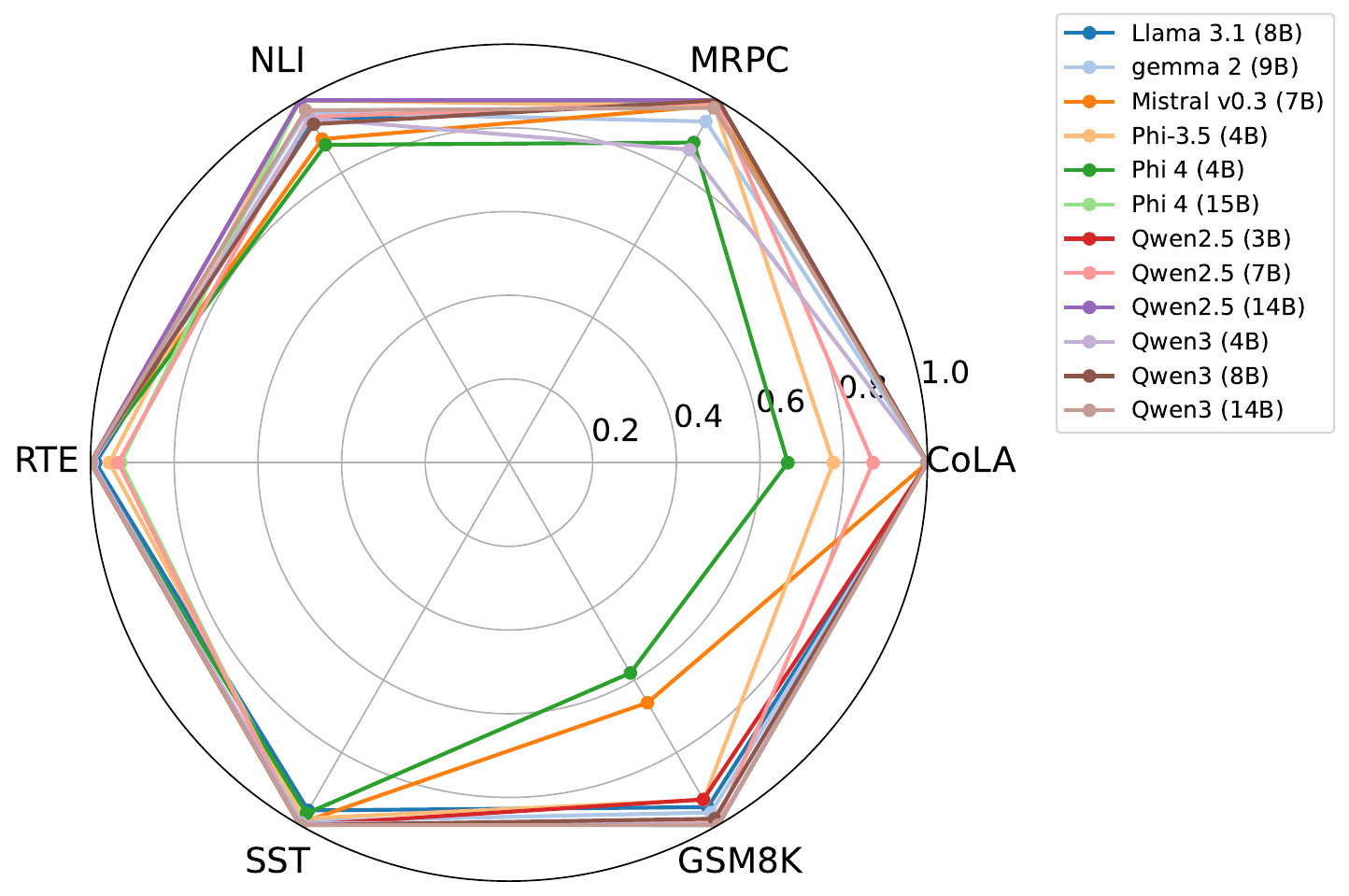}
    \caption{Left: Average utility preservation by method. Right: Utility preservation per LLM with \method{}. Values near 1.0 indicate minimal degradation.}\label{fig:utility_preservation}
\end{figure}
Besides effectiveness in resolving trade-off, we also evaluate model utility preservation on 6 benchmark datasets covering classification and mathematical reasoning tasks. \Figref{fig:utility_preservation} shows the results across methods and models.

\paragraph{Overall utility preservation.} \method{} preserves 95.92\% of the original model's utility on average, outperforming all baseline methods. For 9 of 12 LLMs, utility preservation exceeds 95\%, demonstrating that \method{} successfully enhances alignment without sacrificing general capabilities.

\paragraph{Comparison with baselines.}
SCANS shows the largest utility degradation (averaging 40.98\%) because aggressive magnitude adjustments disrupt the model's internal representations. VectorSteer performs better (89.74\%) but still falls short of \method{} due to its architectural modifications. AlphaSteer and AlphaSteer+ achieve competitive preservation (94.50\% and 94.48\%) through null-space projection, but \method{} still outperforms them while achieving substantially better alignment.

\paragraph{Task-specific analysis.} The utility impact varies across task types. Classification tasks (COLA, MNLI, RTE, MRPC, SST) show minimal degradation, with most models preserving over 97\% performance. Mathematical reasoning (GSM8K) is more affected, with 91.60\% average preservation. This is expected because math reasoning requires precise logical chains that can be disrupted by representation changes. Nevertheless, the impact remains limited compared to the alignment gains.

\paragraph{Model size effects.} Larger and more capable LLMs demonstrate better utility preservation. The three models with lowest preservation---Phi-3.5-4B (92.1\%), Phi-4-4B (91.8\%), and Mistral-v0.3-7B (93.2\%)---are either among the smallest models or have documented limitations in benchmarks~\citep{open-llm-leaderboard-v2,eval-harness}. This suggests that larger models have more robust internal representations that better tolerate the weight modifications introduced by vector alignment.

\subsection{Ablation Studies (RQ3)}\label{sec:ablations}
We analyze three key components: vector identification accuracy, iteration count, and layer selection.

\paragraph{Vector Identification.}
\begin{figure}
    \centering
    \includegraphics[width=0.8\linewidth]{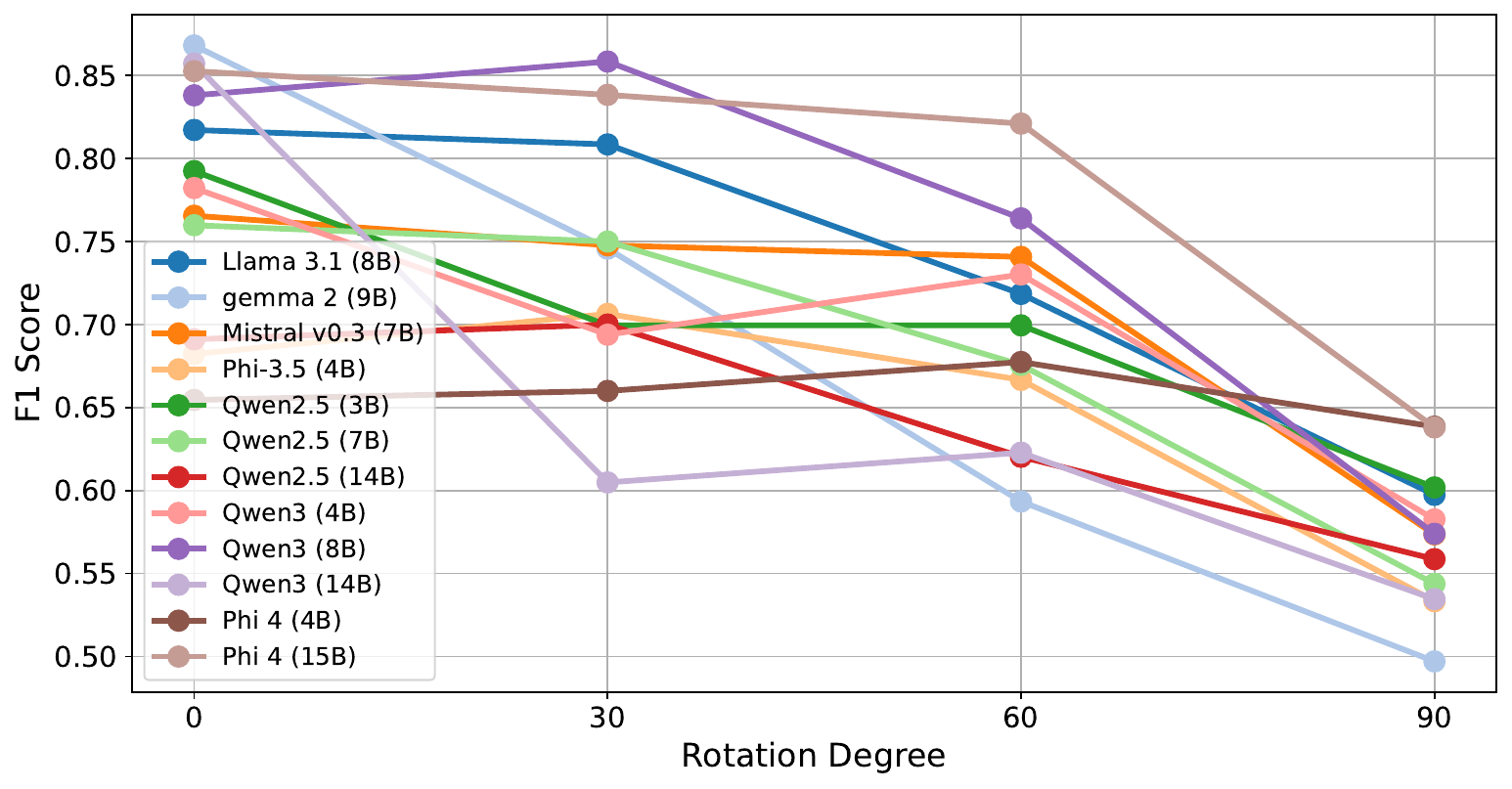}
    \caption{F1 scores with randomly distorted vectors at different angles $D$ from the original benign and answer vectors.}\label{fig:rotation_results}
\end{figure}
To validate our SVM-based vector identification, we replace $v_a$ and $v_b$ with random vectors $D$ degrees away from the originals, where $D$ ranges from $30^\circ$ to $90^\circ$ (\Figref{fig:rotation_results}). The performance degradation correlates with distortion angle: at $D=90^\circ$ (orthogonal to the true vectors), F1 drops by 24.82\% on average, and all 12 models underperform. At $D=60^\circ$, all models still show degradation. However, at $D=30^\circ$, F1 only drops by 5.40\%, indicating that \method{} is robust to small inaccuracies---a practical advantage since SVM hyperplanes may not perfectly capture true decision boundaries---while confirming that accurate identification remains essential.

\paragraph{Iteration Number.}\label{sec:impact_of_iteration_number}
\begin{figure}
    \centering
    \begin{subfigure}{0.32\linewidth}
        \centering
        \includegraphics[width=\linewidth]{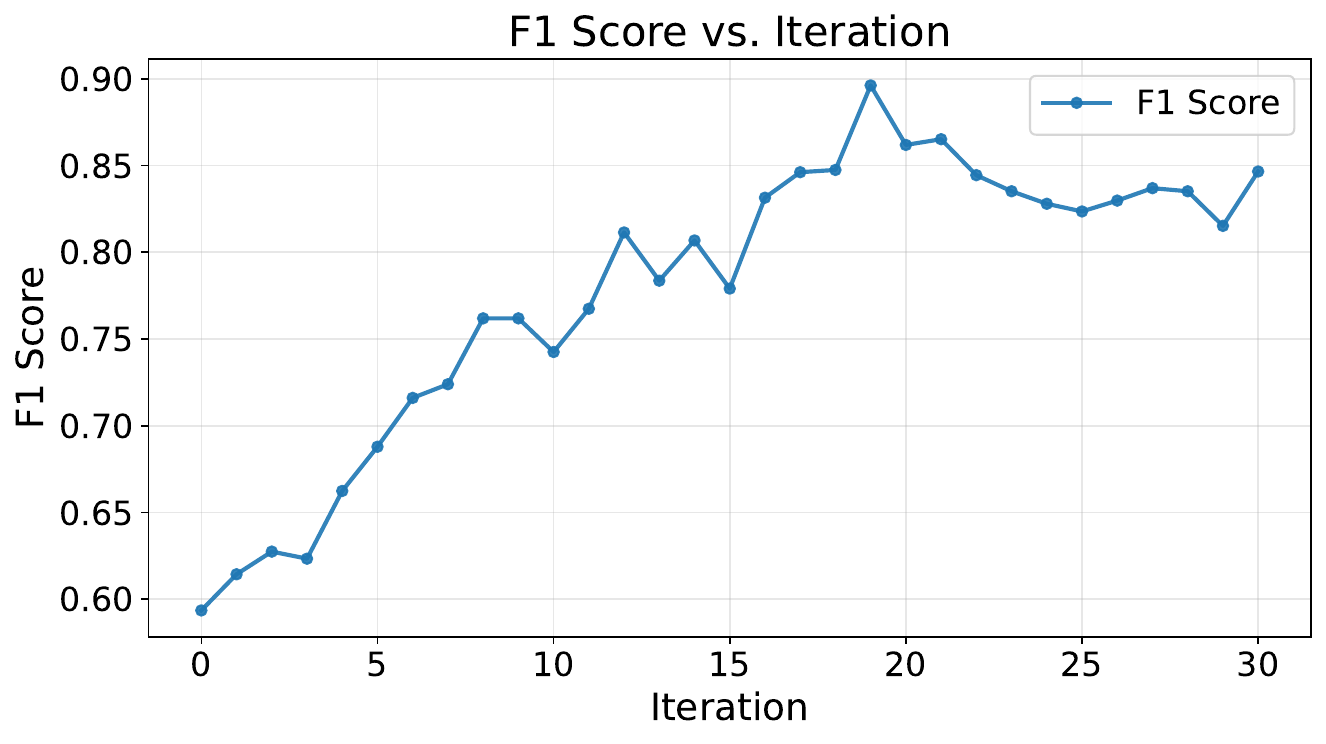}
        \caption{Llama 3.1 (8B)}
    \end{subfigure}
    \begin{subfigure}{0.32\linewidth}
        \centering
        \includegraphics[width=\linewidth]{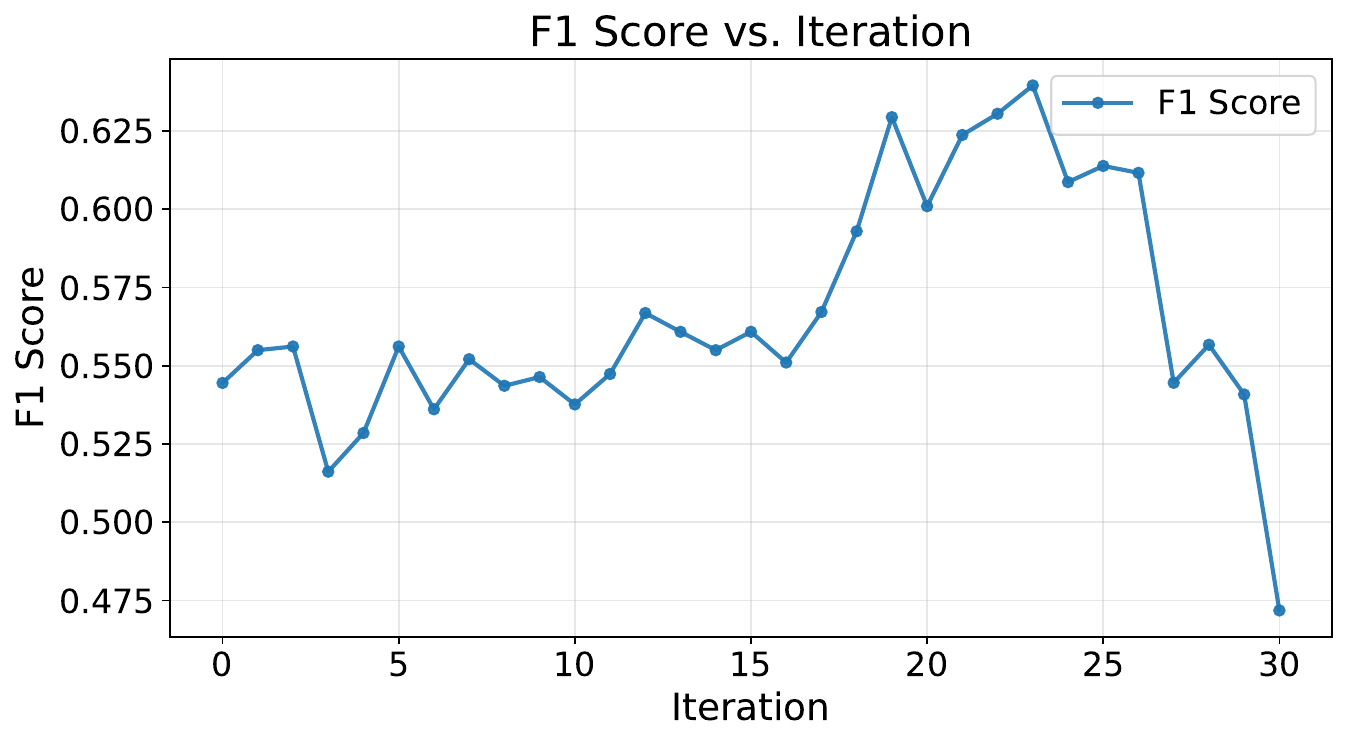}
        \caption{Phi 4 (4B)}
    \end{subfigure}
    \begin{subfigure}{0.32\linewidth}
        \centering
        \includegraphics[width=\linewidth]{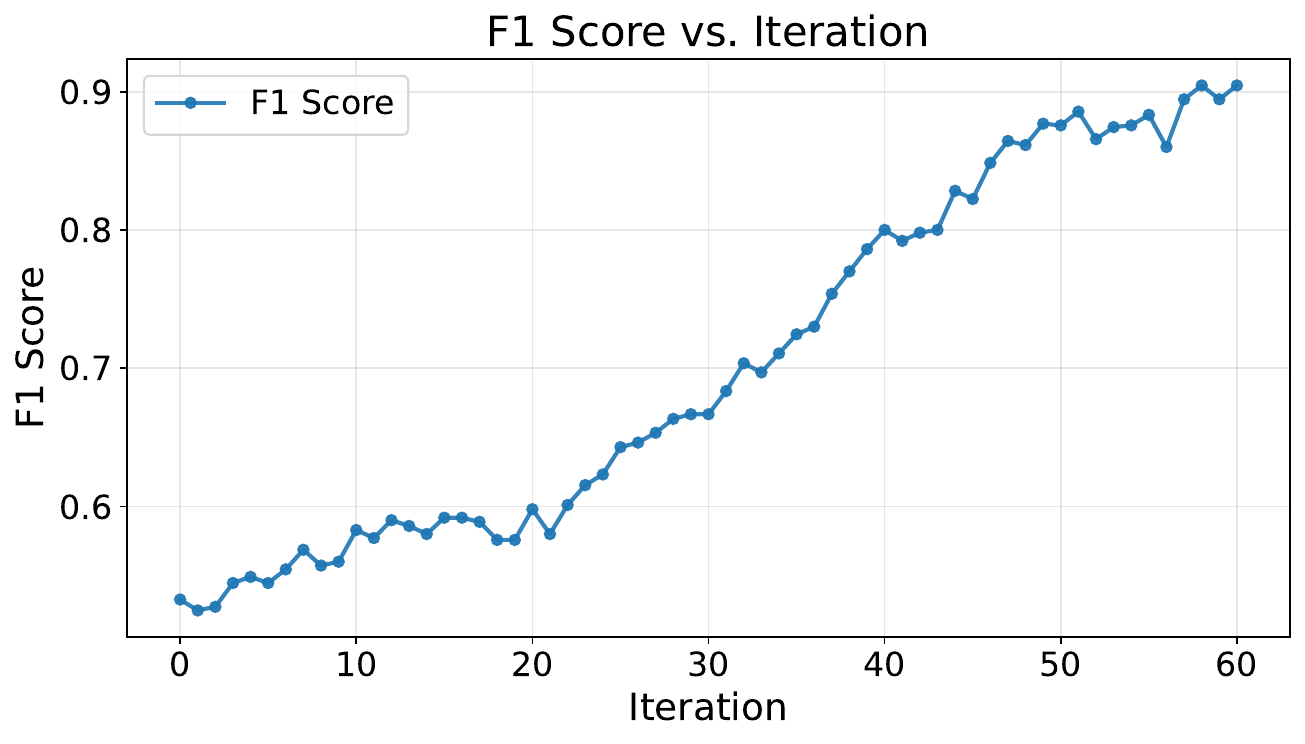}
        \caption{Qwen 3 (14B)}
    \end{subfigure}
    \caption{F1 scores vs.\ iteration number $T$ for three representative models.}\label{fig:iteration_num}
\end{figure}
We vary iteration count $T$ from 1 to 30 (\Figref{fig:iteration_num}).
For clarity, we show the results of three representative models and put the full results in Appendix~\ref{app:detailed_results_on_iteration_number}. Models exhibit distinct convergence patterns: Llama 3.1 (8B) shows rapid improvement and stabilizes around $T=19$; Phi 4 (4B) peaks at $T=19$ but then degrades with additional iterations, suggesting over-modification; Qwen 3 (14B) continues improving through $T=30$ and beyond (as shown in \Figref{fig:iteration_num}, we extended to $T=60$ and observed continued gains).

These patterns explain the suboptimal results in \Tableref{tab:main_results}: Mistral-v0.3-7B, Phi-3.5-4B and Phi-4-4B suffer from over-modification (smaller and performance-limited models~\citep{open-llm-leaderboard-v2,eval-harness} are more susceptible to over-modification), while Qwen3-14B underperforms due to under-iteration. This suggests that model-specific iteration tuning or early stopping based on validation performance is important.

\paragraph{Layer Selection.}\label{sec:impact_of_layer_selection}
\begin{figure}
    \centering
    \includegraphics[width=0.48\linewidth]{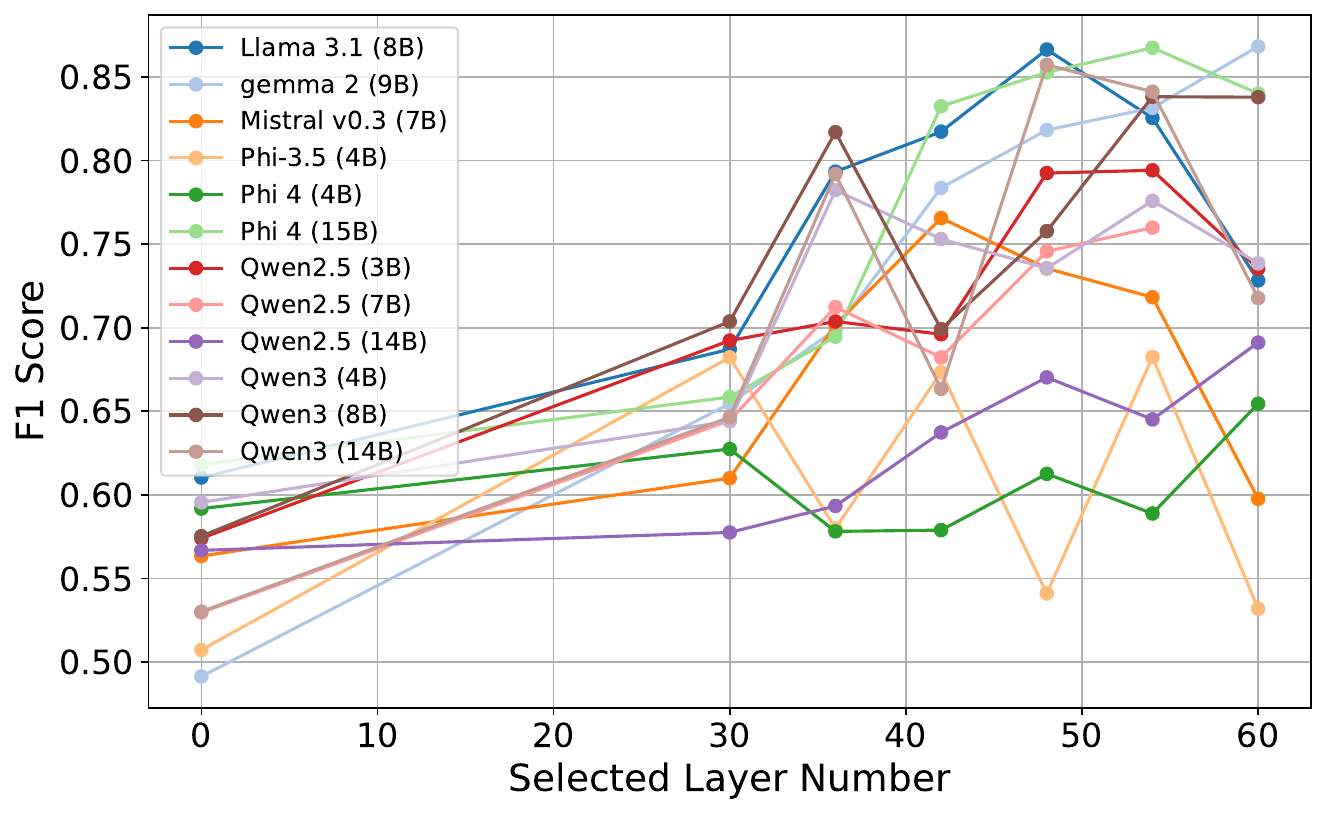}
    \includegraphics[width=0.48\linewidth]{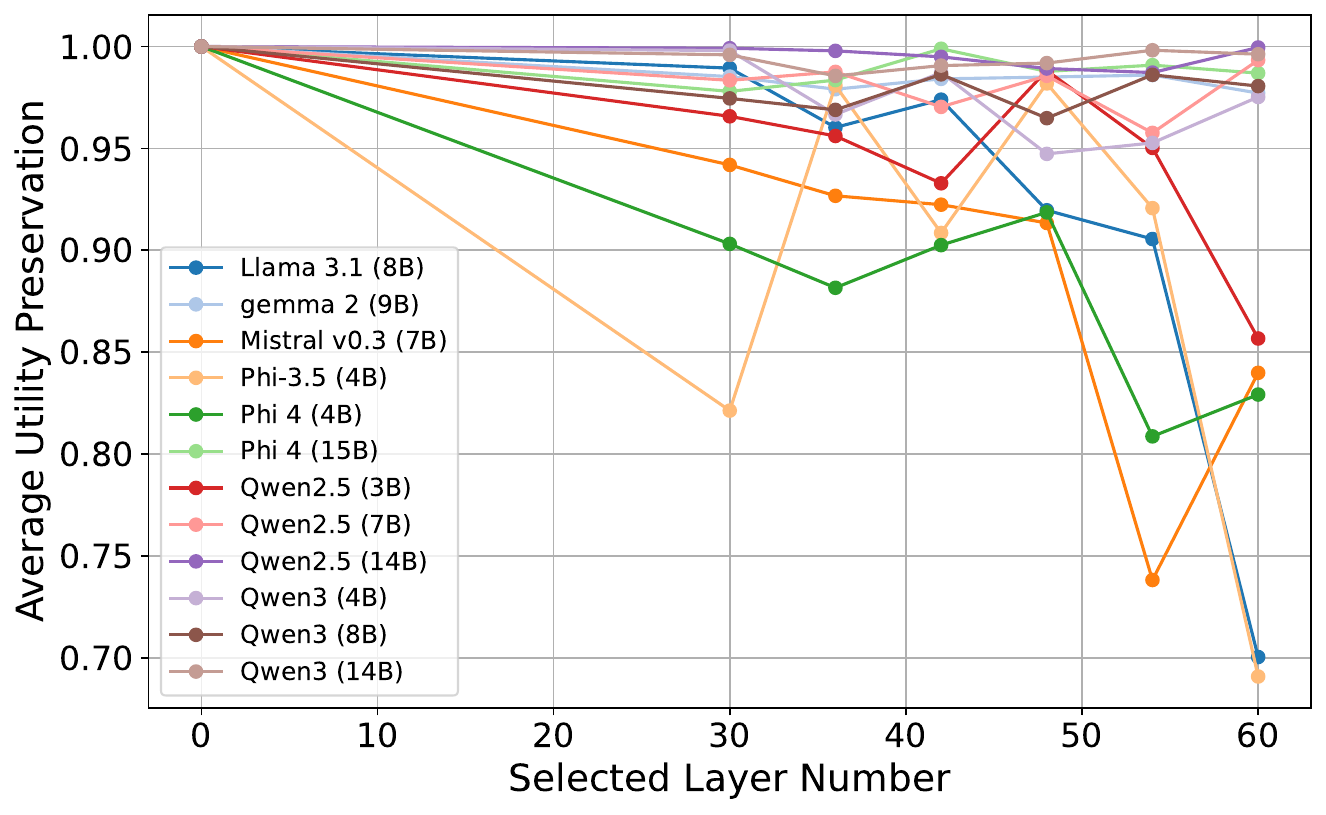}
    \caption{Impact of $L_{select}$ on F1 (left) and utility (right).}\label{fig:selected_layer_num}
\end{figure}
\Figref{fig:selected_layer_num} shows F1 and utility as $L_{select}$ varies from 30 to 60. For alignment, most models exhibit a non-monotonic trend with an optimal $L_{select}$: too few layers limit effectiveness, while too many cause overfitting. For utility preservation, most models remain stable until $L_{select}$ exceeds a threshold, at which point early layers are modified and utility drops sharply. This confirms that later layers are more relevant to safety decisions while early layers are critical for general capabilities, motivating our contribution-score-based layer selection (\Secref{sec:layer_selection}).



\subsection{Transferability (RQ4)}
To assess how well \method{} generalizes to unseen datasets, we evaluate on five additional benchmarks not seen during training (XSTest-Toxic, OR-Bench-Toxic, AdvBench, XSTest, OR-Bench); full results are in Appendix~\ref{app:transferability_experiments} (Table~\ref{tab:transferability}). Performance varies across models and datasets for \emph{all} compared methods, suggesting that cross-dataset generalization is an open challenge shared by vector steering approaches broadly. A key factor is the inconsistency of safety definitions across benchmarks---for instance, \citet{zhang2025orfuzzfuzzingotherside} show that under alternative criteria, nearly half of OR-Bench samples can be regarded as harmful, making consistent cross-dataset transfer intrinsically difficult.

\section{Conclusion}
In this work, we presented \method{}, a novel approach that simultaneously addresses jailbreak and over-refusal by aligning the answer vector with the benign vector through closed-form weight updates---making the model's willingness to answer causally dependent on its safety judgment without requiring fine-tuning or architectural changes.
Experiments on 12 widely used LLMs from 5 model families demonstrate a 11.45\% F1 improvement over the best baseline while preserving 95.92\% utility, and our ablation studies confirm the importance of accurate vector identification and model-specific hyperparameter tuning.

\section{Limitations}
\paragraph{Binary toxicity assumption.} We consider only binary classification (benign vs.\ toxic), whereas real-world toxicity is nuanced and multi-dimensional. Extending \method{} to multi-class or fine-grained toxicity classification remains future work.

\paragraph{Model scale.} Our experiments cover models from 3B to 14B parameters. The effectiveness of \method{} on larger models (e.g., 70B+) remains to be validated, as these models may have different internal representations and require different hyperparameter settings.

\paragraph{Training data dependency.} \method{} requires labeled benign/toxic samples to train the SVMs for vector identification. The quality and representativeness of this training data directly affect alignment performance, and obtaining such labels may not always be straightforward.

\paragraph{Reasoning models.} Vector steering methods, including \method{}, are difficult to apply to LLMs with chain-of-thought reasoning. The control vectors must be identified after reasoning steps are generated, which is computationally expensive, and the randomness in reasoning makes accurate vector identification challenging.

\paragraph{Model-specific tuning.} As shown in our ablation studies, optimal iteration count and layer selection vary across models. While \method{} uses validation-based selection, this requires tuning for a new model, limiting plug-and-play applicability.

\paragraph{Transferability.} The performance of the existing vector steering methods, including \method{}, on unseen datasets varies depending on tasks and models (Appendix~\ref{app:transferability_experiments}).
This imply that current steering methods may need to treat different tasks or domains separately, and improving transferability remains future work.

\paragraph{Static alignment.} The alignment is performed once and does not adapt to new threats or evolving definitions of harmful content. Periodic re-alignment may be needed as the threat landscape changes.

\paragraph{Customized Trade-off.} \method{} aims to improve both jailbreak and over-refusal behavior simultaneously. However, in certain applications (e.g., healthcare~\citep{algaradi2025largelanguagemodelshealthcare,yang2024adversarialattackslargelanguage} or PLC code generation~\citep{liu2024agents4plc}), users may prefer to prioritize one aspect over the other. Extending \method{} to allow for customizable trade-offs remains future work.

\paragraph{Experimental methodology.} Our results are based on single runs with a fixed random seed. While we observe consistent improvements across 12 models, incorporating statistical significance tests would further strengthen our empirical findings.

\paragraph{Adversarial robustness.} By coupling the answer direction with the safety direction, \method{} may introduce new attack surfaces: an adversary who can manipulate the model's internal safety representation (e.g., via targeted adversarial inputs) could potentially bypass the alignment. Evaluating and improving robustness against such adaptive attacks remains an important direction for future work.

\section{Ethical Considerations}
\subsection{Potential Risks}
Though \method{} aims to enhance the safety alignment of LLMs, it can be misused to manipulate model behaviors in unintended ways. For instance, attackers could potentially exploit the vector alignment technique to bypass safety mechanisms or introduce harmful biases into the model.
Besides, the datasets used for training and evaluation may contain biases.
\subsection{AI Assistants Usage}
We employ GPT-5.2~\citep{openai2024gpt4technicalreport} and Github Copilot\footnote{\url{https://github.com/features/copilot}} to assist in writing code for experiments. We carefully review and verify all AI-generated content to ensure accuracy and integrity.

\section*{Acknowledgements}
This work is partially supported by the State Key Laboratory of Industrial Control Technology, China (Grant No. ICT2026C02), for the authors from Zhejiang University only.

\bibliography{custom}

@misc{guo2025deepseek,
  title={DeepSeek-R1: Incentivizing Reasoning Capability in LLMs via Reinforcement Learning},
  author={Guo, Daya and Yang, Dejian and Zhang, Haowei and Song, Junxiao and Zhang, Ruoyu and Xu, Runxin and Zhu, Qihao and Ma, Shirong and Wang, Peiyi and Bi, Xiao and others},
  year={2025},
  eprint={2501.12948},
  archivePrefix={arXiv},
  primaryClass={cs.CL},
  url={https://arxiv.org/abs/2501.12948}
}

@article{zou2023representation,
  title={Representation engineering: A top-down approach to ai transparency},
  author={Zou, Andy and Phan, Long and Chen, Sarah and Campbell, James and Guo, Phillip and Ren, Richard and Pan, Alexander and Yin, Xuwang and Mazeika, Mantas and Dombrowski, Ann-Kathrin and others},
  journal={arXiv preprint arXiv:2310.01405},
  year={2023}
}

@article{sheng2025alphasteer,
  title={AlphaSteer: Learning Refusal Steering with Principled Null-Space Constraint},
  author={Sheng, Leheng and Shen, Changshuo and Zhao, Weixiang and Fang, Junfeng and Liu, Xiaohao and Liang, Zhenkai and Wang, Xiang and Zhang, An and Chua, Tat-Seng},
  journal={arXiv preprint arXiv:2506.07022},
  year={2025}
}

@inproceedings{cao2025scans,
  title={SCANS: Mitigating the exaggerated safety for llms via safety-conscious activation steering},
  author={Cao, Zouying and Yang, Yifei and Zhao, Hai},
  booktitle={Proceedings of the AAAI Conference on Artificial Intelligence},
  volume={39},
  pages={23523--23531},
  year={2025}
}

@article{ray2024mitigating,
  title={Mitigating exaggerated safety in large language models},
  author={Ray, Ruchira and Bhalani, Ruchi},
  journal={arXiv preprint arXiv:2405.05418},
  year={2024}
}

@article{Penrose_1955,
  title={A generalized inverse for matrices},
  author={Penrose, R.},
  journal={Mathematical Proceedings of the Cambridge Philosophical Society},
  volume={51},
  number={3},
  pages={406--413},
  year={1955},
  DOI={10.1017/S0305004100030401}
}

@inproceedings{geva-etal-2021-transformer,
    title = "Transformer Feed-Forward Layers Are Key-Value Memories",
    author = "Geva, Mor  and
      Schuster, Roei  and
      Berant, Jonathan  and
      Levy, Omer",
    editor = "Moens, Marie-Francine  and
      Huang, Xuanjing  and
      Specia, Lucia  and
      Yih, Scott Wen-tau",
    booktitle = "Proceedings of the 2021 Conference on Empirical Methods in Natural Language Processing",
    month = nov,
    year = "2021",
    address = "Online and Punta Cana, Dominican Republic",
    publisher = "Association for Computational Linguistics",
    url = "https://aclanthology.org/2021.emnlp-main.446/",
    doi = "10.18653/v1/2021.emnlp-main.446",
    pages = "5484--5495"
}

@article{arditi2024refusal,
  title={Refusal in language models is mediated by a single direction},
  author={Arditi, Andy and Obeso, Oscar and Syed, Aaquib and Paleka, Daniel and Panickssery, Nina and Gurnee, Wes and Nanda, Neel},
  journal={Advances in Neural Information Processing Systems},
  volume={37},
  pages={136037--136083},
  year={2024}
}

@article{christiano2017deep,
  title={Deep reinforcement learning from human preferences},
  author={Christiano, Paul F and Leike, Jan and Brown, Tom and Martic, Miljan and Legg, Shane and Amodei, Dario},
  journal={Advances in neural information processing systems},
  volume={30},
  year={2017}
}

@article{stiennon2020learning,
  title={Learning to summarize with human feedback},
  author={Stiennon, Nisan and Ouyang, Long and Wu, Jeffrey and Ziegler, Daniel and Lowe, Ryan and Voss, Chelsea and Radford, Alec and Amodei, Dario and Christiano, Paul F},
  journal={Advances in neural information processing systems},
  volume={33},
  pages={3008--3021},
  year={2020}
}

@article{xhonneux2024efficient,
  title={Efficient adversarial training in llms with continuous attacks},
  author={Xhonneux, Sophie and Sordoni, Alessandro and G{\"u}nnemann, Stephan and Gidel, Gauthier and Schwinn, Leo},
  journal={Advances in Neural Information Processing Systems},
  volume={37},
  pages={1502--1530},
  year={2024}
}

@article{liu2024adversarial,
  title={Adversarial tuning: Defending against jailbreak attacks for llms},
  author={Liu, Fan and Xu, Zhao and Liu, Hao},
  journal={arXiv preprint arXiv:2406.06622},
  year={2024}
}

@article{zhang2025jbshield,
  title={Jbshield: Defending large language models from jailbreak attacks through activated concept analysis and manipulation},
  author={Zhang, Shenyi and Zhai, Yuchen and Guo, Keyan and Hu, Hongxin and Guo, Shengnan and Fang, Zheng and Zhao, Lingchen and Shen, Chao and Wang, Cong and Wang, Qian},
  journal={arXiv preprint arXiv:2502.07557},
  year={2025}
}

@article{llama3modelcard,
title={Llama 3 Model Card},
author={AI@Meta},
year={2024},
url = {https://github.com/meta-llama/llama3/blob/main/MODEL_CARD.md}
}

@article{gemma_2024,
    title={Gemma},
    url={https://www.kaggle.com/m/3301},
    DOI={10.34740/KAGGLE/M/3301},
    publisher={Kaggle},
    author={Gemma Team},
    year={2024}
}

@misc{jiang2023mistral7b,
      title={Mistral 7B}, 
      author={Albert Q. Jiang and Alexandre Sablayrolles and Arthur Mensch and Chris Bamford and Devendra Singh Chaplot and Diego de las Casas and Florian Bressand and Gianna Lengyel and Guillaume Lample and Lucile Saulnier and Lélio Renard Lavaud and Marie-Anne Lachaux and Pierre Stock and Teven Le Scao and Thibaut Lavril and Thomas Wang and Timothée Lacroix and William El Sayed},
      year={2023},
      eprint={2310.06825},
      archivePrefix={arXiv},
      primaryClass={cs.CL},
      url={https://arxiv.org/abs/2310.06825}, 
}

@misc{abdin2024phi3technicalreporthighly,
      title={Phi-3 Technical Report: A Highly Capable Language Model Locally on Your Phone}, 
      author={Marah Abdin and Jyoti Aneja and Hany Awadalla and Ahmed Awadallah and Ammar Ahmad Awan and Nguyen Bach and Amit Bahree and Arash Bakhtiari and Jianmin Bao and Harkirat Behl and Alon Benhaim and Misha Bilenko and Johan Bjorck and Sébastien Bubeck and Martin Cai and Qin Cai and Vishrav Chaudhary and Dong Chen and Dongdong Chen and Weizhu Chen and Yen-Chun Chen and Yi-Ling Chen and Hao Cheng and Parul Chopra and Xiyang Dai and Matthew Dixon and Ronen Eldan and Victor Fragoso and Jianfeng Gao and Mei Gao and Min Gao and Amit Garg and Allie Del Giorno and Abhishek Goswami and Suriya Gunasekar and Emman Haider and Junheng Hao and Russell J. Hewett and Wenxiang Hu and Jamie Huynh and Dan Iter and Sam Ade Jacobs and Mojan Javaheripi and Xin Jin and Nikos Karampatziakis and Piero Kauffmann and Mahoud Khademi and Dongwoo Kim and Young Jin Kim and Lev Kurilenko and James R. Lee and Yin Tat Lee and Yuanzhi Li and Yunsheng Li and Chen Liang and Lars Liden and Xihui Lin and Zeqi Lin and Ce Liu and Liyuan Liu and Mengchen Liu and Weishung Liu and Xiaodong Liu and Chong Luo and Piyush Madan and Ali Mahmoudzadeh and David Majercak and Matt Mazzola and Caio César Teodoro Mendes and Arindam Mitra and Hardik Modi and Anh Nguyen and Brandon Norick and Barun Patra and Daniel Perez-Becker and Thomas Portet and Reid Pryzant and Heyang Qin and Marko Radmilac and Liliang Ren and Gustavo de Rosa and Corby Rosset and Sambudha Roy and Olatunji Ruwase and Olli Saarikivi and Amin Saied and Adil Salim and Michael Santacroce and Shital Shah and Ning Shang and Hiteshi Sharma and Yelong Shen and Swadheen Shukla and Xia Song and Masahiro Tanaka and Andrea Tupini and Praneetha Vaddamanu and Chunyu Wang and Guanhua Wang and Lijuan Wang and Shuohang Wang and Xin Wang and Yu Wang and Rachel Ward and Wen Wen and Philipp Witte and Haiping Wu and Xiaoxia Wu and Michael Wyatt and Bin Xiao and Can Xu and Jiahang Xu and Weijian Xu and Jilong Xue and Sonali Yadav and Fan Yang and Jianwei Yang and Yifan Yang and Ziyi Yang and Donghan Yu and Lu Yuan and Chenruidong Zhang and Cyril Zhang and Jianwen Zhang and Li Lyna Zhang and Yi Zhang and Yue Zhang and Yunan Zhang and Xiren Zhou},
      year={2024},
      eprint={2404.14219},
      archivePrefix={arXiv},
      primaryClass={cs.CL},
      url={https://arxiv.org/abs/2404.14219}, 
}

@misc{microsoft2025phi4minitechnicalreportcompact,
      title={Phi-4-Mini Technical Report: Compact yet Powerful Multimodal Language Models via Mixture-of-LoRAs}, 
      author={Microsoft and : and Abdelrahman Abouelenin and Atabak Ashfaq and Adam Atkinson and Hany Awadalla and Nguyen Bach and Jianmin Bao and Alon Benhaim and Martin Cai and Vishrav Chaudhary and Congcong Chen and Dong Chen and Dongdong Chen and Junkun Chen and Weizhu Chen and Yen-Chun Chen and Yi-ling Chen and Qi Dai and Xiyang Dai and Ruchao Fan and Mei Gao and Min Gao and Amit Garg and Abhishek Goswami and Junheng Hao and Amr Hendy and Yuxuan Hu and Xin Jin and Mahmoud Khademi and Dongwoo Kim and Young Jin Kim and Gina Lee and Jinyu Li and Yunsheng Li and Chen Liang and Xihui Lin and Zeqi Lin and Mengchen Liu and Yang Liu and Gilsinia Lopez and Chong Luo and Piyush Madan and Vadim Mazalov and Arindam Mitra and Ali Mousavi and Anh Nguyen and Jing Pan and Daniel Perez-Becker and Jacob Platin and Thomas Portet and Kai Qiu and Bo Ren and Liliang Ren and Sambuddha Roy and Ning Shang and Yelong Shen and Saksham Singhal and Subhojit Som and Xia Song and Tetyana Sych and Praneetha Vaddamanu and Shuohang Wang and Yiming Wang and Zhenghao Wang and Haibin Wu and Haoran Xu and Weijian Xu and Yifan Yang and Ziyi Yang and Donghan Yu and Ishmam Zabir and Jianwen Zhang and Li Lyna Zhang and Yunan Zhang and Xiren Zhou},
      year={2025},
      eprint={2503.01743},
      archivePrefix={arXiv},
      primaryClass={cs.CL},
      url={https://arxiv.org/abs/2503.01743}, 
}

@misc{qwen2.5,
    title = {Qwen2.5: A Party of Foundation Models},
    url = {https://qwenlm.github.io/blog/qwen2.5/},
    author = {Qwen Team},
    month = {September},
    year = {2024}
}

@article{qwen2,
      title={Qwen2 Technical Report}, 
      author={An Yang and Baosong Yang and Binyuan Hui and Bo Zheng and Bowen Yu and Chang Zhou and Chengpeng Li and Chengyuan Li and Dayiheng Liu and Fei Huang and Guanting Dong and Haoran Wei and Huan Lin and Jialong Tang and Jialin Wang and Jian Yang and Jianhong Tu and Jianwei Zhang and Jianxin Ma and Jin Xu and Jingren Zhou and Jinze Bai and Jinzheng He and Junyang Lin and Kai Dang and Keming Lu and Keqin Chen and Kexin Yang and Mei Li and Mingfeng Xue and Na Ni and Pei Zhang and Peng Wang and Ru Peng and Rui Men and Ruize Gao and Runji Lin and Shijie Wang and Shuai Bai and Sinan Tan and Tianhang Zhu and Tianhao Li and Tianyu Liu and Wenbin Ge and Xiaodong Deng and Xiaohuan Zhou and Xingzhang Ren and Xinyu Zhang and Xipin Wei and Xuancheng Ren and Yang Fan and Yang Yao and Yichang Zhang and Yu Wan and Yunfei Chu and Yuqiong Liu and Zeyu Cui and Zhenru Zhang and Zhihao Fan},
      journal={arXiv preprint arXiv:2407.10671},
      year={2024}
}

@misc{qwen3technicalreport,
      title={Qwen3 Technical Report}, 
      author={Qwen Team},
      year={2025},
      eprint={2505.09388},
      archivePrefix={arXiv},
      primaryClass={cs.CL},
      url={https://arxiv.org/abs/2505.09388}, 
}

@article{yuan2025seval,
        title={S-Eval: Towards Automated and Comprehensive Safety Evaluation for Large Language Models},
        author={Yuan, Xiaohan and Li, Jinfeng and Wang, Dongxia and Chen, Yuefeng and Mao, Xiaofeng and Huang, Longtao and Chen, Jialuo and Xue, Hui and Liu, Xiaoxia and Wang, Wenhai and Ren, Kui and Wang, Jingyi},
        journal={Proceedings of the ACM on Software Engineering},
        volume={2},
        number={ISSTA},
        pages={2136--2157},
        year={2025},
        publisher={ACM New York, NY, USA},
        url = {https://doi.org/10.1145/3728971},
        doi = {10.1145/3728971}
}

@misc{zhang2025orfuzzfuzzingotherside,
      title={ORFuzz: Fuzzing the "Other Side" of LLM Safety -- Testing Over-Refusal}, 
      author={Haonan Zhang and Dongxia Wang and Yi Liu and Kexin Chen and Jiashui Wang and Xinlei Ying and Long Liu and Wenhai Wang},
      year={2025},
      eprint={2508.11222},
      archivePrefix={arXiv},
      primaryClass={cs.SE},
      url={https://arxiv.org/abs/2508.11222}, 
}

@misc{röttger2024xstesttestsuiteidentifying,
      title={XSTest: A Test Suite for Identifying Exaggerated Safety Behaviours in Large Language Models}, 
      author={Paul Röttger and Hannah Rose Kirk and Bertie Vidgen and Giuseppe Attanasio and Federico Bianchi and Dirk Hovy},
      year={2024},
      eprint={2308.01263},
      archivePrefix={arXiv},
      primaryClass={cs.CL},
      url={https://arxiv.org/abs/2308.01263}, 
}

@misc{cui2025orbenchoverrefusalbenchmarklarge,
      title={OR-Bench: An Over-Refusal Benchmark for Large Language Models}, 
      author={Justin Cui and Wei-Lin Chiang and Ion Stoica and Cho-Jui Hsieh},
      year={2025},
      eprint={2405.20947},
      archivePrefix={arXiv},
      primaryClass={cs.CL},
      url={https://arxiv.org/abs/2405.20947}, 
}

@misc{zou2023universal,
      title={Universal and Transferable Adversarial Attacks on Aligned Language Models}, 
      author={Andy Zou and Zifan Wang and J. Zico Kolter and Matt Fredrikson},
      year={2023},
      eprint={2307.15043},
      archivePrefix={arXiv},
      primaryClass={cs.CL}
}

@article{warstadt2018neural,
  title={Neural Network Acceptability Judgments},
  author={Warstadt, Alex and Singh, Amanpreet and Bowman, Samuel R.},
  journal={arXiv preprint 1805.12471},
  year={2018}
}

@inproceedings{dolan2005automatically,
  title={Automatically constructing a corpus of sentential paraphrases},
  author={Dolan, William B and Brockett, Chris},
  booktitle={Proceedings of the International Workshop on Paraphrasing},
  year={2005}
}

@inproceedings{socher-etal-2013-recursive,
    title = "Recursive Deep Models for Semantic Compositionality Over a Sentiment Treebank",
    author = "Socher, Richard  and
      Perelygin, Alex  and
      Wu, Jean  and
      Chuang, Jason  and
      Manning, Christopher D.  and
      Ng, Andrew  and
      Potts, Christopher",
    booktitle = "Proceedings of the 2013 Conference on Empirical Methods in Natural Language Processing",
    month = oct,
    year = "2013",
    address = "Seattle, Washington, USA",
    publisher = "Association for Computational Linguistics",
    url = "https://www.aclweb.org/anthology/D13-1170",
    pages = "1631--1642",
}

@article{cobbe2021gsm8k,
  title={Training Verifiers to Solve Math Word Problems},
  author={Cobbe, Karl and Kosaraju, Vineet and Bavarian, Mohammad and Chen, Mark and Jun, Heewoo and Kaiser, Lukasz and Plappert, Matthias and Tworek, Jerry and Hilton, Jacob and Nakano, Reiichiro and Hesse, Christopher and Schulman, John},
  journal={arXiv preprint arXiv:2110.14168},
  year={2021}
}

@article{cortes1995support,
  title={Support-vector networks},
  author={Cortes, Corinna and Vapnik, Vladimir},
  journal={Machine learning},
  volume={20},
  number={3},
  pages={273--297},
  year={1995},
  publisher={Springer}
}

@article{kwiatkowski2019natural,
  title={Natural questions: a benchmark for question answering research},
  author={Kwiatkowski, Tom and Palomaki, Jennimaria and Redfield, Olivia and Collins, Michael and Parikh, Ankur and Alberti, Chris and Epstein, Danielle and Polosukhin, Illia and Devlin, Jacob and Lee, Kenton and others},
  journal={Transactions of the Association for Computational Linguistics},
  volume={7},
  pages={453--466},
  year={2019},
  publisher={MIT Press One Rogers Street, Cambridge, MA 02142-1209, USA journals-info~…}
}

@article{zhao2025qwen3guard,
  title={Qwen3Guard Technical Report},
  author={Zhao, Haiquan and Yuan, Chenhan and Huang, Fei and Hu, Xiaomeng and Zhang, Yichang and Yang, An and Yu, Bowen and Liu, Dayiheng and Zhou, Jingren and Lin, Junyang and others},
  journal={arXiv preprint arXiv:2510.14276},
  year={2025}
}

@misc{openai2024gpt4technicalreport,
      title={GPT-4 Technical Report}, 
      author={OpenAI},
      year={2024},
      eprint={2303.08774},
      archivePrefix={arXiv},
      primaryClass={cs.CL},
      url={https://arxiv.org/abs/2303.08774}, 
}

@article{lee2024programming,
  title={Programming refusal with conditional activation steering},
  author={Lee, Bruce W and Padhi, Inkit and Ramamurthy, Karthikeyan Natesan and Miehling, Erik and Dognin, Pierre and Nagireddy, Manish and Dhurandhar, Amit},
  journal={arXiv preprint arXiv:2409.05907},
  year={2024}
}

@misc{open-llm-leaderboard-v2,
  author = {Clémentine Fourrier and Nathan Habib and Alina Lozovskaya and Konrad Szafer and Thomas Wolf},
  title = {Open LLM Leaderboard v2},
  year = {2024},
  publisher = {Hugging Face},
  howpublished = "\url{https://huggingface.co/spaces/open-llm-leaderboard/open_llm_leaderboard}",
}

@misc{eval-harness,
  author       = {Gao, Leo and Tow, Jonathan and Biderman, Stella and Black, Sid and DiPofi, Anthony and Foster, Charles and Golding, Laurence and Hsu, Jeffrey and McDonell, Kyle and Muennighoff, Niklas and Phang, Jason and Reynolds, Laria and Tang, Eric and Thite, Anish and Wang, Ben and Wang, Kevin and Zou, Andy},
  title        = {A framework for few-shot language model evaluation},
  month        = sep,
  year         = 2021,
  publisher    = {Zenodo},
  version      = {v0.0.1},
  doi          = {10.5281/zenodo.5371628},
  url          = {https://doi.org/10.5281/zenodo.5371628},
}

@article{bentivogli2009fifth,
  title={The Fifth PASCAL Recognizing Textual Entailment Challenge.},
  author={Bentivogli, Luisa and Clark, Peter and Dagan, Ido and Giampiccolo, Danilo},
  journal={TAC},
  volume={7},
  number={8},
  pages={1},
  year={2009}
}

@inproceedings{williams2018broad,
  title={A broad-coverage challenge corpus for sentence understanding through inference},
  author={Williams, Adina and Nangia, Nikita and Bowman, Samuel},
  booktitle={Proceedings of the 2018 conference of the North American chapter of the association for computational linguistics: human language technologies, volume 1 (long papers)},
  pages={1112--1122},
  year={2018}
}

@article{elhage2022superposition,
  title={Toy models of superposition},
  author={Elhage, Nelson and Hume, Tristan and Olsson, Catherine and Schiefer, Nicholas and Henighan, Tom and Kravec, Shauna and Hatfield-Dodds, Zac and Lasenby, Robert and Drain, Dawn and Chen, Carol and others},
  journal={arXiv preprint arXiv:2209.10652},
  year={2022}
}

@article{pedregosa2011scikit,
  title={Scikit-learn: Machine learning in Python},
  author={Pedregosa, Fabian and Varoquaux, Ga{\"e}l and Gramfort, Alexandre and Michel, Vincent and Thirion, Bertrand and Grisel, Olivier and Blondel, Mathieu and Prettenhofer, Peter and Weiss, Ron and Dubourg, Vincent and others},
  journal={the Journal of machine Learning research},
  volume={12},
  pages={2825--2830},
  year={2011},
  publisher={JMLR. org}
}

@article{liu2024agents4plc,
  title={Agents4plc: Automating closed-loop plc code generation and verification in industrial control systems using llm-based agents},
  author={Liu, Zihan and Zeng, Ruinan and Wang, Dongxia and Peng, Gengyun and Wang, Jingyi and Liu, Qiang and Liu, Peiyu and Wang, Wenhai},
  journal={arXiv preprint arXiv:2410.14209},
  year={2024}
}

@misc{algaradi2025largelanguagemodelshealthcare,
      title={Large Language Models in Healthcare}, 
      author={Mohammed Al-Garadi and Tushar Mungle and Abdulaziz Ahmed and Abeed Sarker and Zhuqi Miao and Michael E. Matheny},
      year={2025},
      eprint={2503.04748},
      archivePrefix={arXiv},
      primaryClass={cs.CY},
      url={https://arxiv.org/abs/2503.04748}, 
}

@misc{yang2024adversarialattackslargelanguage,
      title={Adversarial Attacks on Large Language Models in Medicine}, 
      author={Yifan Yang and Qiao Jin and Furong Huang and Zhiyong Lu},
      year={2024},
      eprint={2406.12259},
      archivePrefix={arXiv},
      primaryClass={cs.AI},
      url={https://arxiv.org/abs/2406.12259}, 
}

@article{chi2024llama,
  title={Llama guard 3 vision: Safeguarding human-ai image understanding conversations},
  author={Chi, Jianfeng and Karn, Ujjwal and Zhan, Hongyuan and Smith, Eric and Rando, Javier and Zhang, Yiming and Plawiak, Kate and Coudert, Zacharie Delpierre and Upasani, Kartikeya and Pasupuleti, Mahesh},
  journal={arXiv preprint arXiv:2411.10414},
  year={2024}
}

@misc{li2024lockpicking,
  title={Lockpicking LLMs: A Logit-based Jailbreak Using Token-level Manipulation},
  author={Li, Yuxi and Liu, Yi and Li, Yuekang and Shi, Ling and Deng, Gelei and Chen, Shengquan and Wang, Kailong},
  year={2024},
  eprint={2405.13068},
  archivePrefix={arXiv},
  primaryClass={cs.CR},
  url={https://arxiv.org/abs/2405.13068}
}

@misc{li2024crosslanguage,
  title={A Cross-Language Investigation into Jailbreak Attacks in Large Language Models},
  author={Li, Jie and Liu, Yi and Liu, Chongyang and Shi, Ling and Ren, Xiaoning and Zheng, Yaowen},
  year={2024},
  eprint={2401.16765},
  archivePrefix={arXiv},
  primaryClass={cs.CL},
  url={https://arxiv.org/abs/2401.16765}
}

@misc{deng2024pandora,
  title={Pandora: Jailbreak GPTs by Retrieval Augmented Generation Poisoning},
  author={Deng, Gelei and Liu, Yi and Wang, Kailong and Li, Yuekang and Zhang, Tianwei},
  year={2024},
  eprint={2402.08416},
  archivePrefix={arXiv},
  primaryClass={cs.CR},
  url={https://arxiv.org/abs/2402.08416}
}

@misc{liu2024jailbreakingchatgptpromptengineering,
      title={Jailbreaking ChatGPT via Prompt Engineering: An Empirical Study}, 
      author={Yi Liu and Gelei Deng and Zhengzi Xu and Yuekang Li and Yaowen Zheng and Ying Zhang and Lida Zhao and Tianwei Zhang and Kailong Wang and Yang Liu},
      year={2024},
      eprint={2305.13860},
      archivePrefix={arXiv},
      primaryClass={cs.SE},
      url={https://arxiv.org/abs/2305.13860}, 
}

@inproceedings{Deng_2024, series={NDSS 2024},
   title={MASTERKEY: Automated Jailbreaking of Large Language Model Chatbots},
   url={http://dx.doi.org/10.14722/ndss.2024.24188},
   DOI={10.14722/ndss.2024.24188},
   booktitle={Proceedings 2024 Network and Distributed System Security Symposium},
   publisher={Internet Society},
   author={Deng, Gelei and Liu, Yi and Li, Yuekang and Wang, Kailong and Zhang, Ying and Li, Zefeng and Wang, Haoyu and Zhang, Tianwei and Liu, Yang},
   year={2024},
   collection={NDSS 2024} }

@inproceedings{liu2024efficient,
  title={Efficient detection of toxic prompts in large language models},
  author={Liu, Yi and Yu, Junzhe and Sun, Huijia and Shi, Ling and Deng, Gelei and Chen, Yuqi and Liu, Yang},
  booktitle={Proceedings of the 39th IEEE/ACM International Conference on Automated Software Engineering},
  pages={455--467},
  year={2024}
}

@misc{chen2024characterizingevaluatingreliabilityllms,
      title={Characterizing and Evaluating the Reliability of LLMs against Jailbreak Attacks}, 
      author={Kexin Chen and Yi Liu and Dongxia Wang and Jiaying Chen and Wenhai Wang},
      year={2024},
      eprint={2408.09326},
      archivePrefix={arXiv},
      primaryClass={cs.CL},
      url={https://arxiv.org/abs/2408.09326}, 
}

\newpage
\appendix

\section{Angles between Answer Vectors and Benign Vectors}\label{app:angle_between_vectors}
\begin{figure*}
    \centering
    \begin{subfigure}
        {0.32\linewidth}
        \centering
        \includegraphics[width=\linewidth]{figure/angle_llama.pdf}
        \caption*{Llama 3.1 (8B)}
    \end{subfigure}
    \begin{subfigure}
        {0.32\linewidth}
        \centering
        \includegraphics[width=\linewidth]{figure/angle_gemma.pdf}
        \caption*{gemma-2 (9B)}
    \end{subfigure}
    \begin{subfigure}
        {0.32\linewidth}
        \centering
        \includegraphics[width=\linewidth]{figure/angle_mistral.pdf}
        \caption*{Mistral-v0.3 (7B)}
    \end{subfigure}
    \begin{subfigure}
        {0.32\linewidth}
        \centering
        \includegraphics[width=\linewidth]{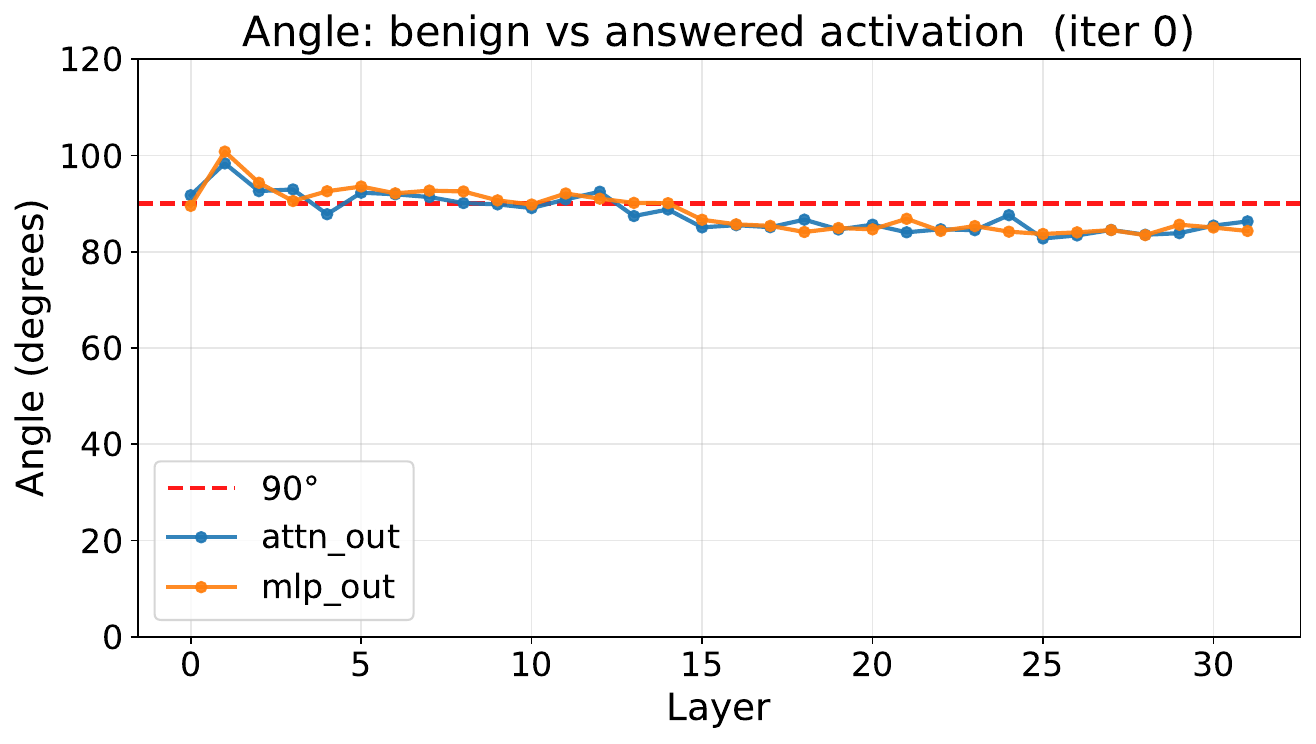}
        \caption*{Phi-3.5 (4B)}
    \end{subfigure}
    \begin{subfigure}
        {0.32\linewidth}
        \centering
        \includegraphics[width=\linewidth]{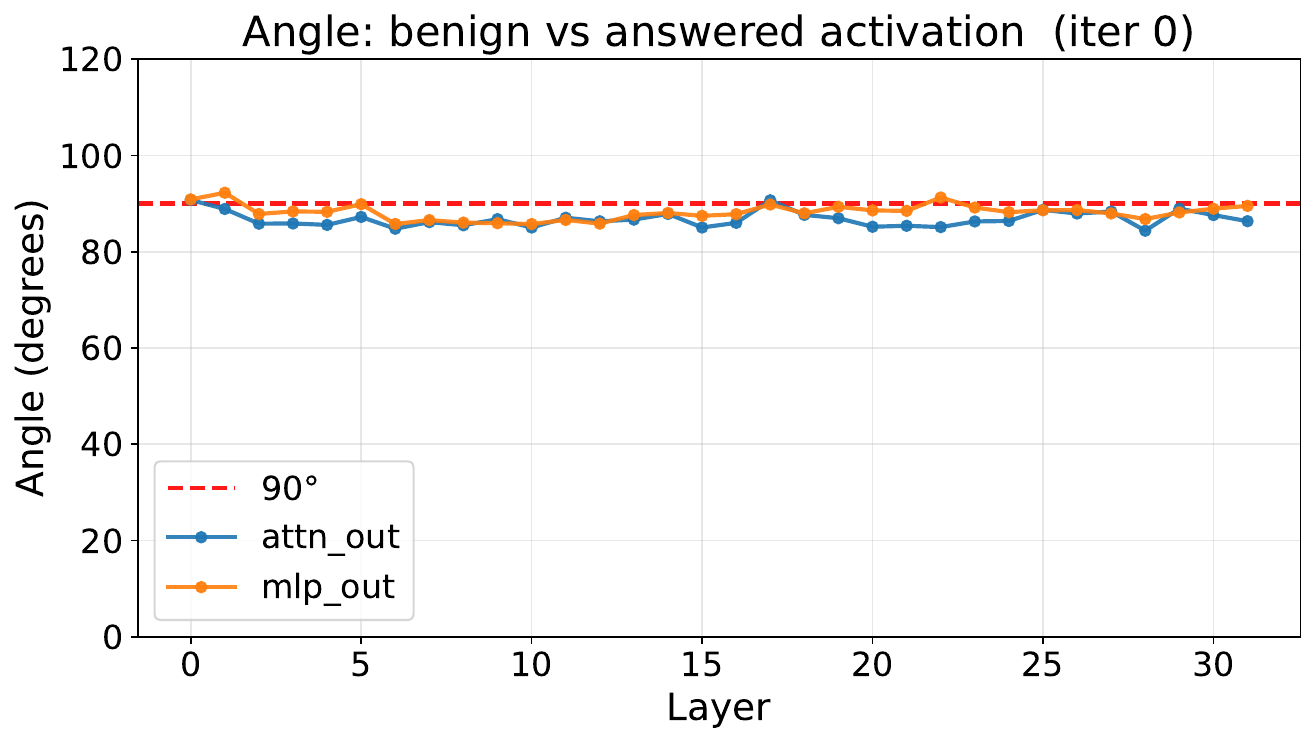}
        \caption*{Phi-4 (4B)}
    \end{subfigure}
    \begin{subfigure}
        {0.32\linewidth}
        \centering
        \includegraphics[width=\linewidth]{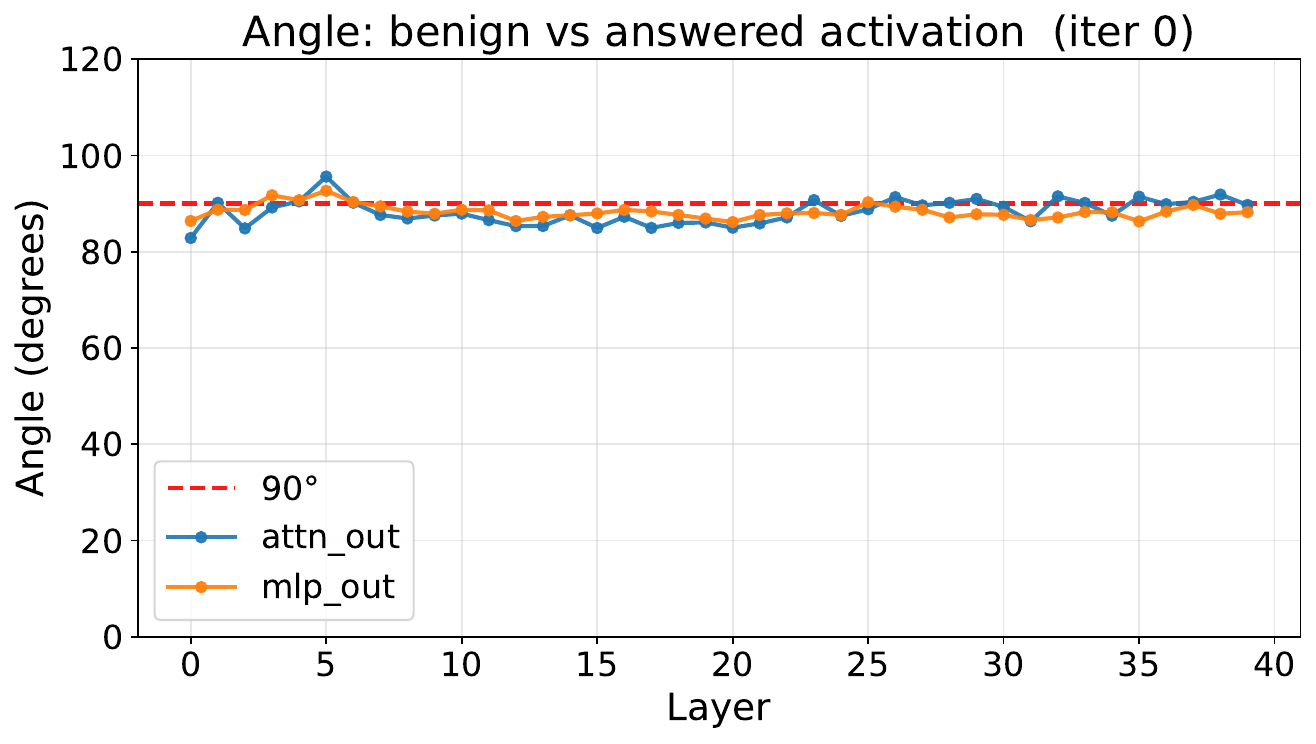}
        \caption*{Phi-4 (15B)}
    \end{subfigure}
    \begin{subfigure}
        {0.32\linewidth}
        \centering
        \includegraphics[width=\linewidth]{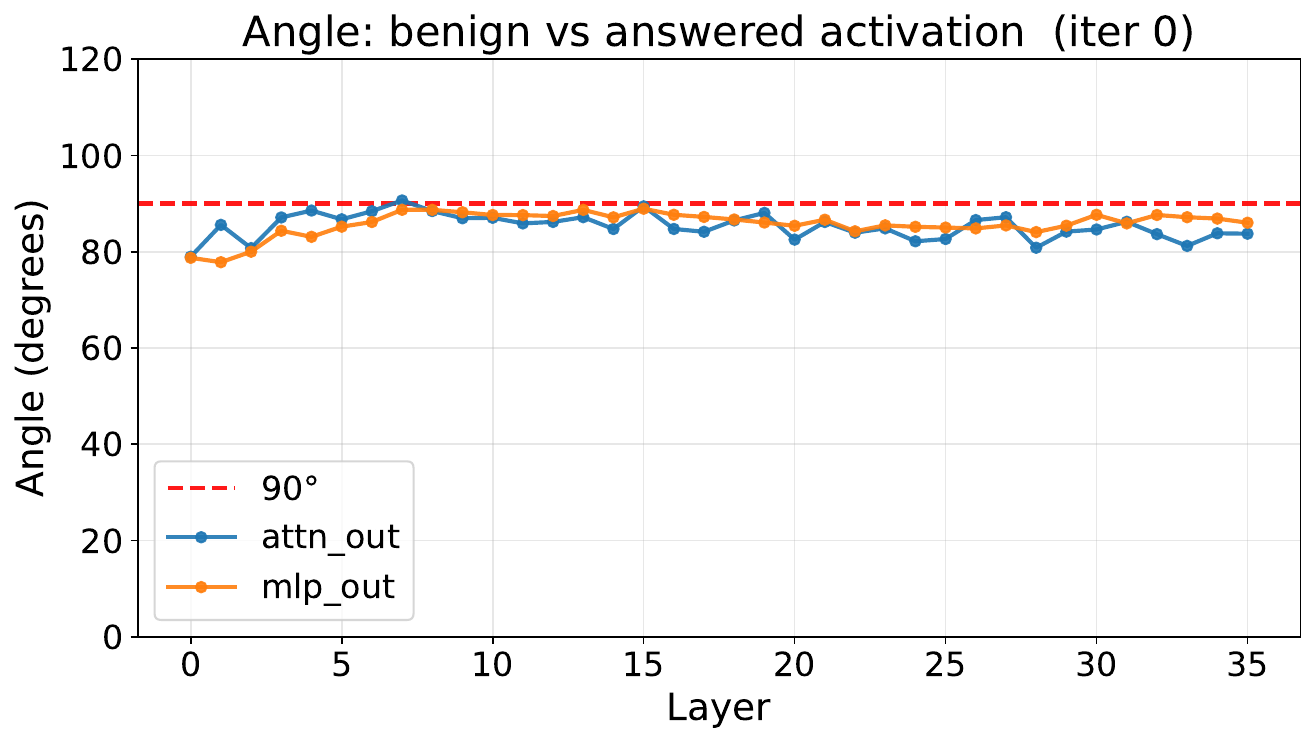}
        \caption*{Qwen2.5 (3B)}
    \end{subfigure}
    \begin{subfigure}
        {0.32\linewidth}
        \centering
        \includegraphics[width=\linewidth]{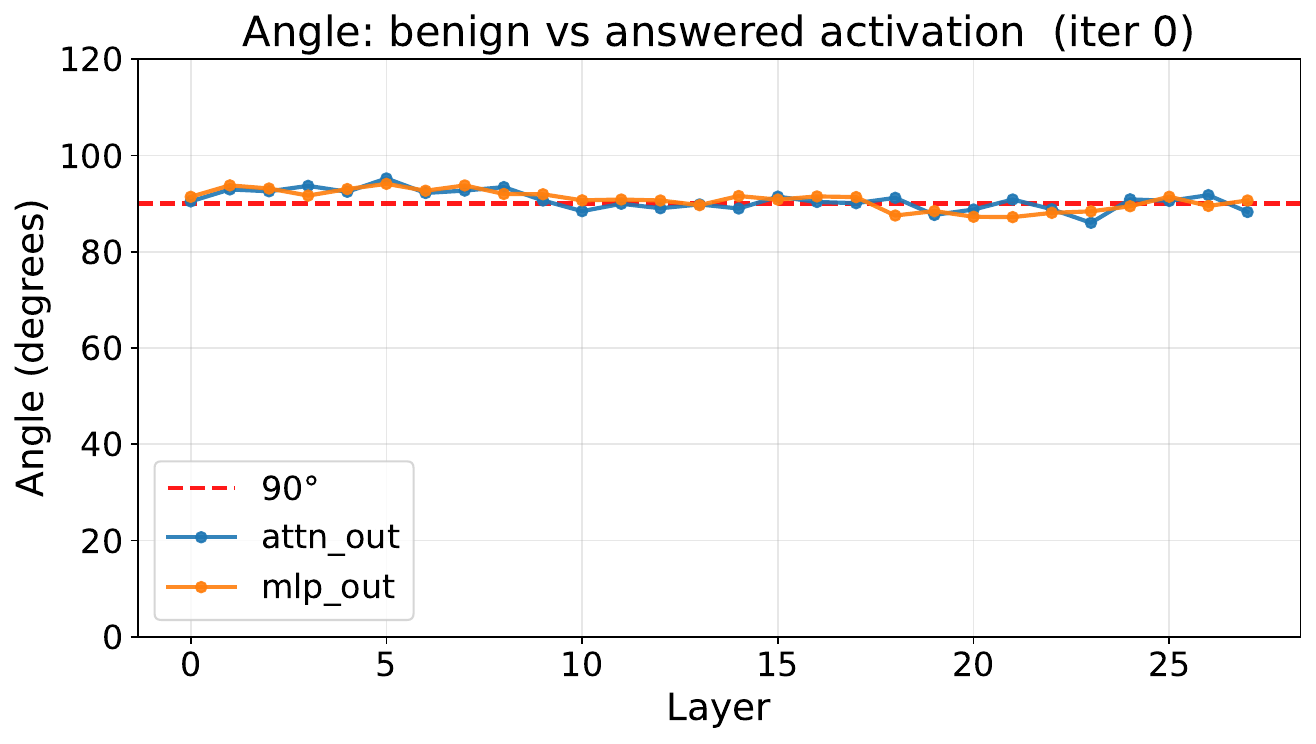}
        \caption*{Qwen2.5 (7B)}
    \end{subfigure}
    \begin{subfigure}
        {0.32\linewidth}
        \centering
        \includegraphics[width=\linewidth]{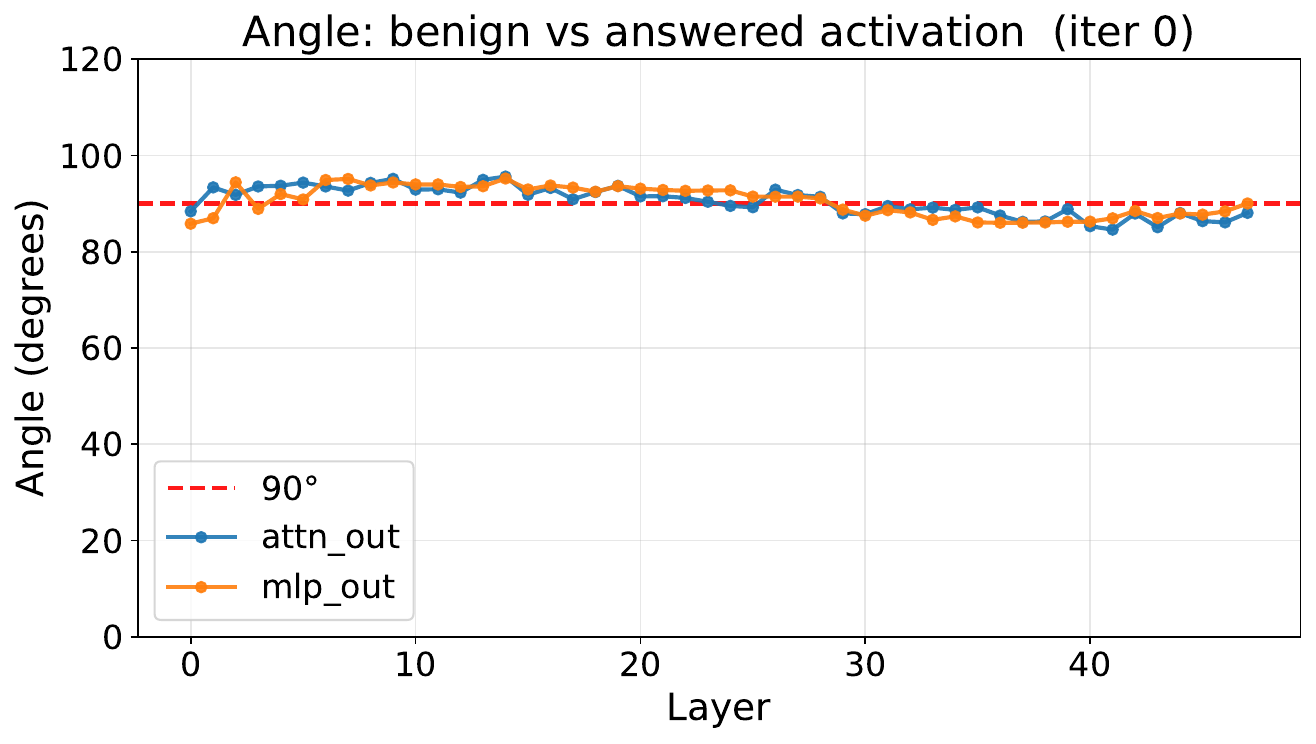}
        \caption*{Qwen2.5 (14B)}
    \end{subfigure}
    \begin{subfigure}
        {0.32\linewidth}
        \centering
        \includegraphics[width=\linewidth]{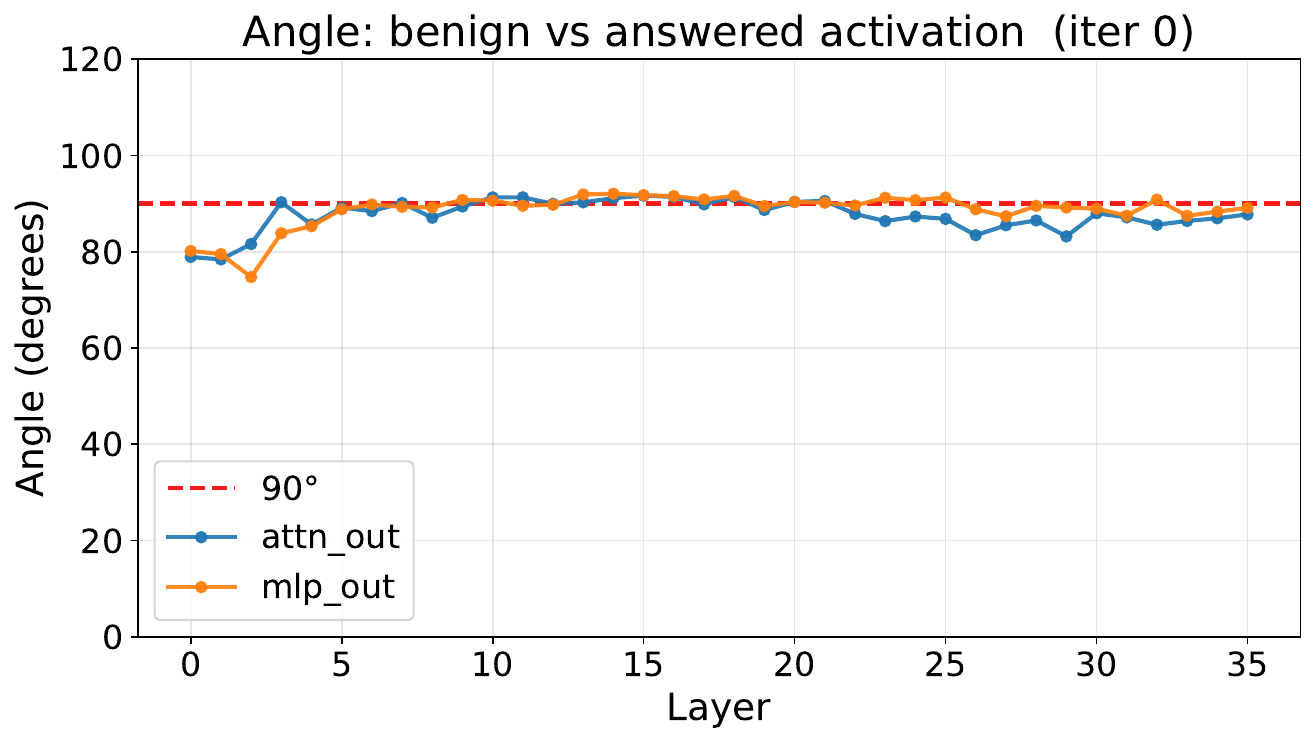}
        \caption*{Qwen3 (4B)}   
    \end{subfigure}
    \begin{subfigure}
        {0.32\linewidth}
        \centering
        \includegraphics[width=\linewidth]{figure/angle_qwen3_8b.pdf}
        \caption*{Qwen3 (8B)}
    \end{subfigure}
    \begin{subfigure}
        {0.32\linewidth}
        \centering
        \includegraphics[width=\linewidth]{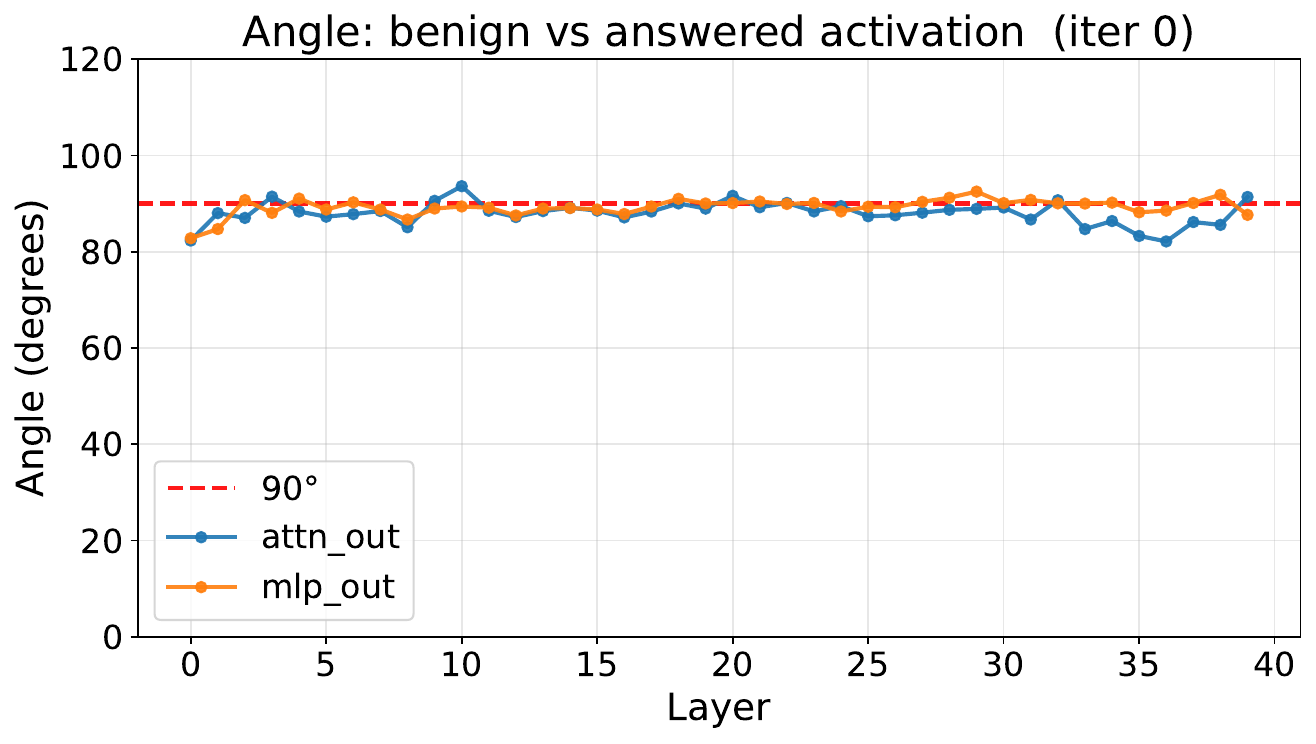}
        \caption*{Qwen3 (14B)}
    \end{subfigure}
    \caption{Angles between answer vectors and benign vectors of different LLMs.}\label{fig:app_angle_between_vectors}
\end{figure*}

As shown in \Figref{fig:app_angle_between_vectors}, the angles between the answer vectors and benign vectors of different LLMs are approximately $90^{\circ}$, indicating that they are nearly orthogonal.

\section{Discussion on Layer Type Selection}\label{app:layer_type_selection}
\begin{figure*}
    \centering
    \begin{subfigure}{0.32\linewidth}
        \includegraphics[width=\linewidth]{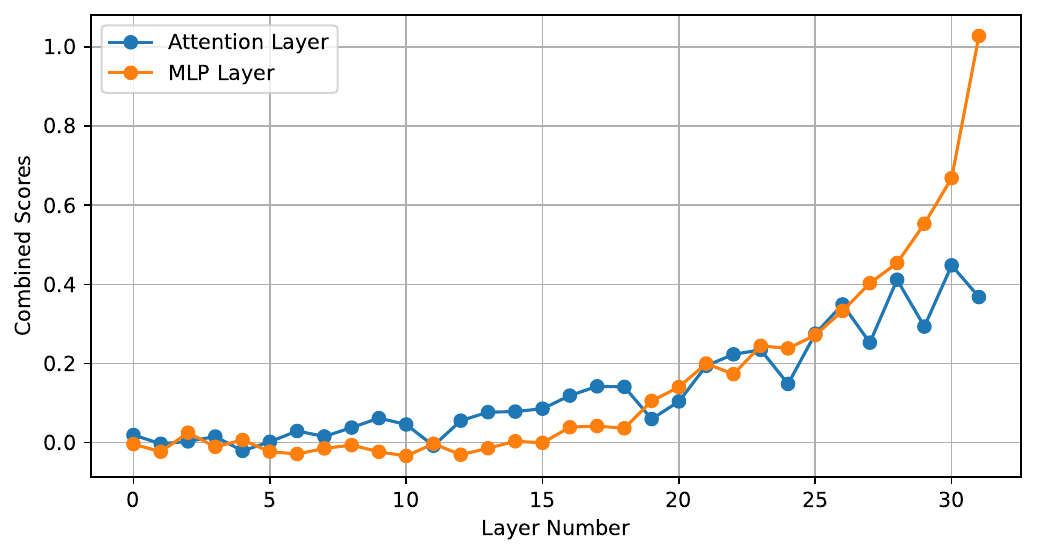}
        \caption*{Llama-3.1 (8B)}
    \end{subfigure}
    \begin{subfigure}{0.32\linewidth}
        \includegraphics[width=\linewidth]{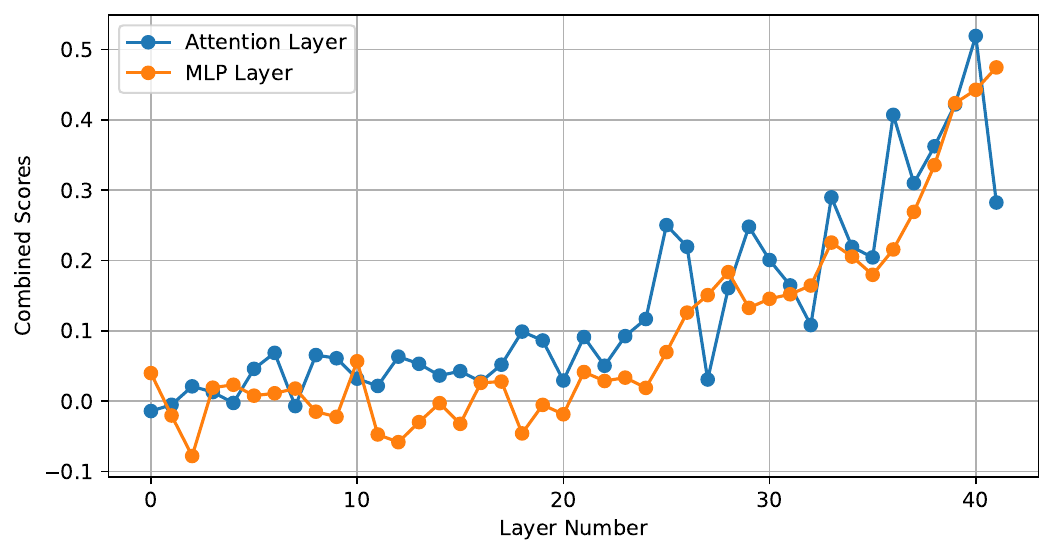}
        \caption*{gemma-2 (9B)}
    \end{subfigure}
    \begin{subfigure}{0.32\linewidth}
        \includegraphics[width=\linewidth]{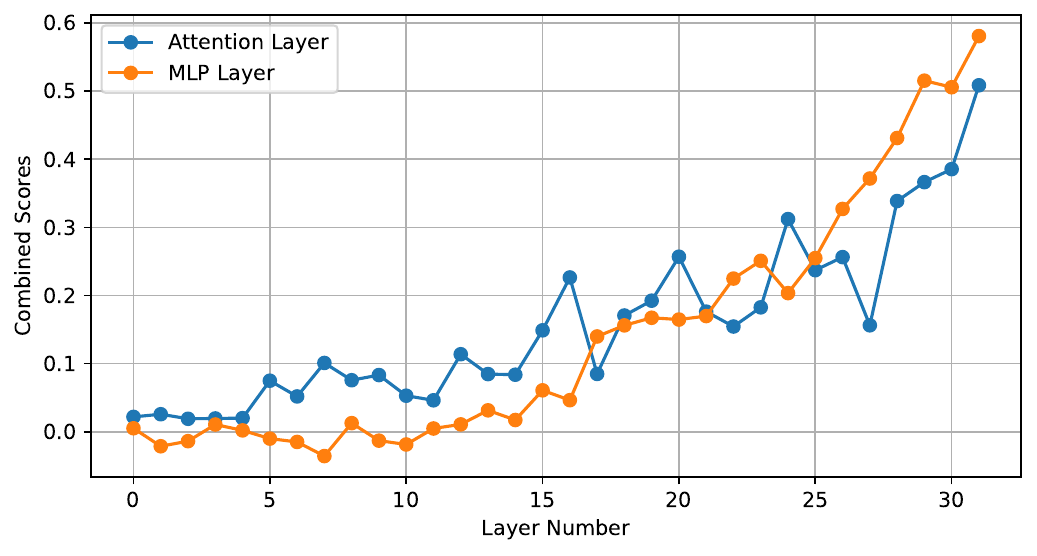}
        \caption*{Mistral-v0.3 (7B)}
    \end{subfigure}
    \begin{subfigure}{0.32\linewidth}
        \includegraphics[width=\linewidth]{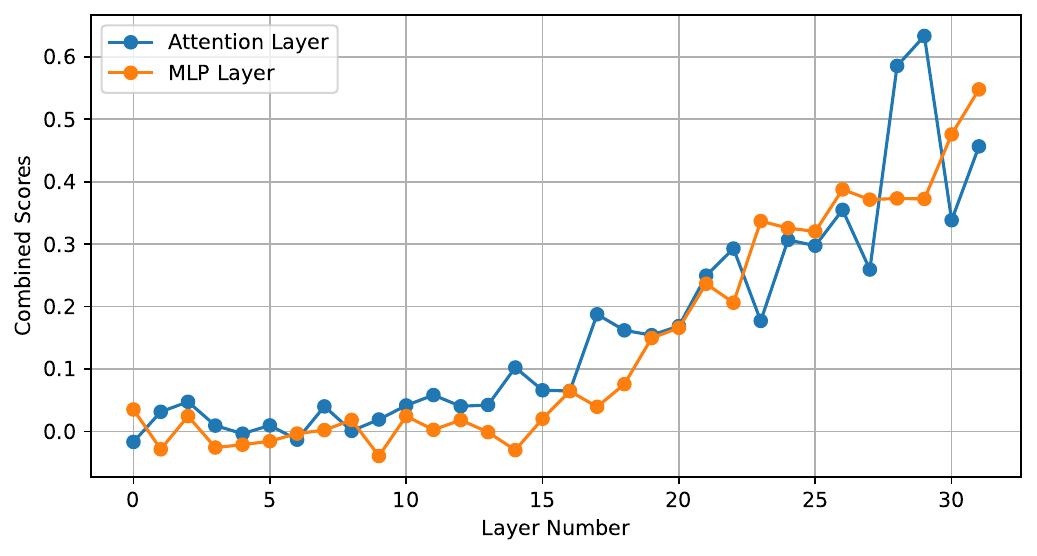}
        \caption*{Phi-3.5 (4B)}
    \end{subfigure}
    \begin{subfigure}{0.32\linewidth}
        \includegraphics[width=\linewidth]{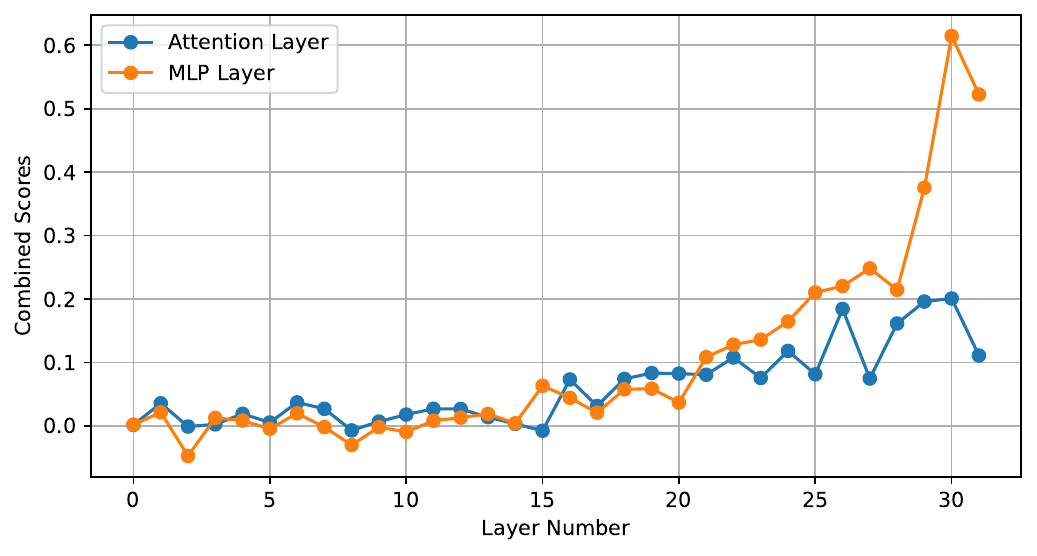}
        \caption*{Phi-4 (4B)}
    \end{subfigure}
    \begin{subfigure}{0.32\linewidth}
        \includegraphics[width=\linewidth]{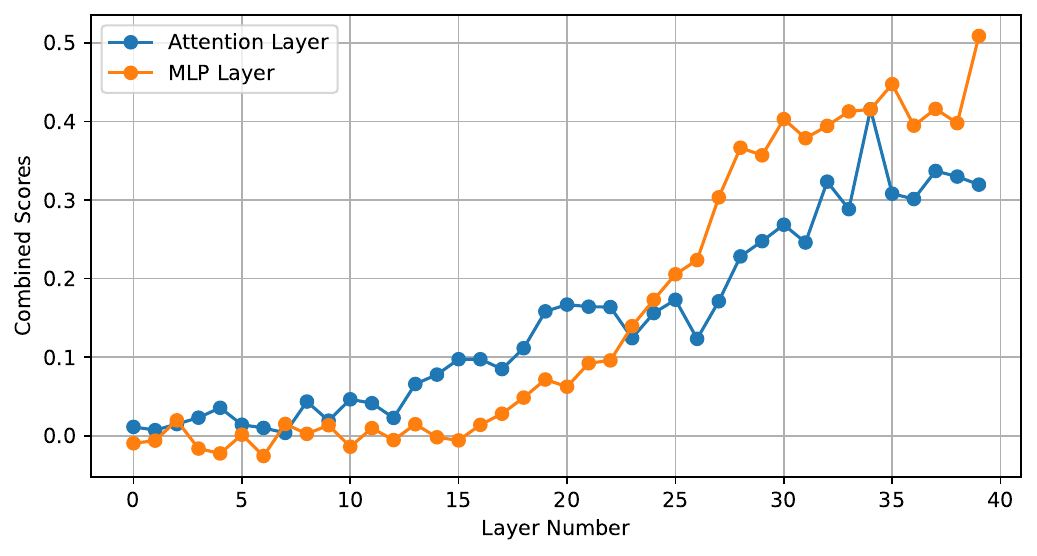}
        \caption*{Phi-4 (15B)}
    \end{subfigure}
    \begin{subfigure}{0.32\linewidth}
        \includegraphics[width=\linewidth]{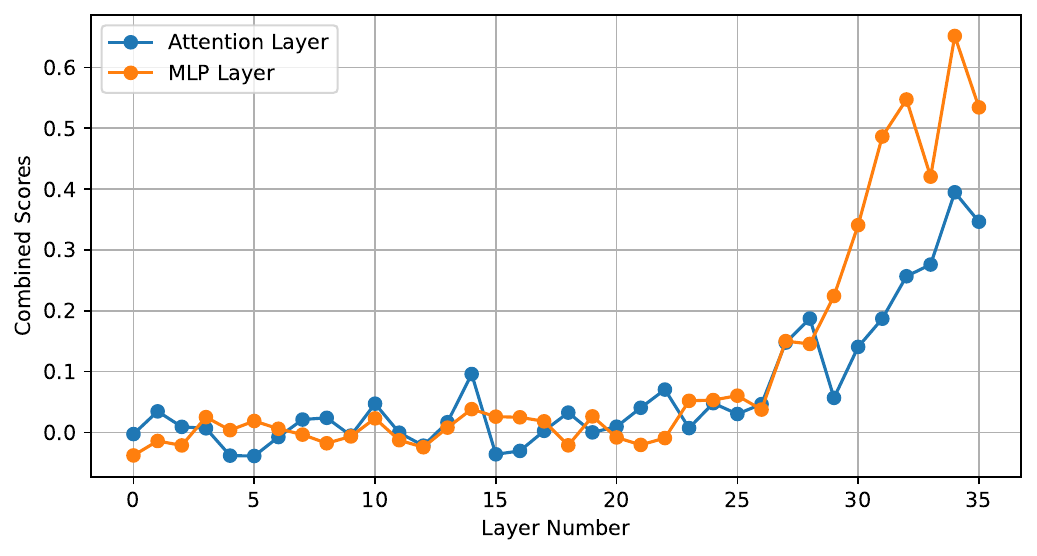}
        \caption*{Qwen2.5 (3B)}
    \end{subfigure}
    \begin{subfigure}{0.32\linewidth}
        \includegraphics[width=\linewidth]{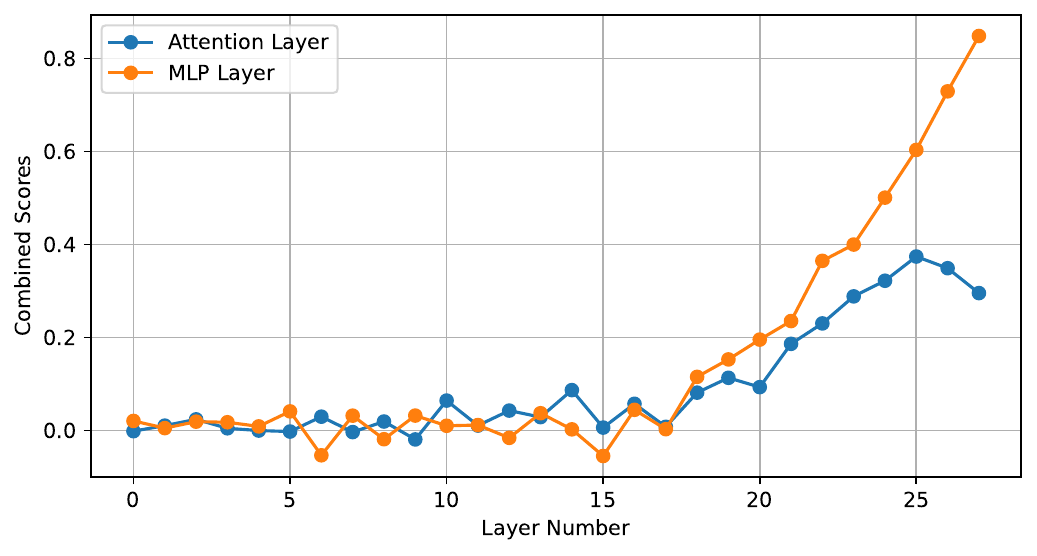}
        \caption*{Qwen2.5 (7B)}
    \end{subfigure}
    \begin{subfigure}{0.32\linewidth}
        \includegraphics[width=\linewidth]{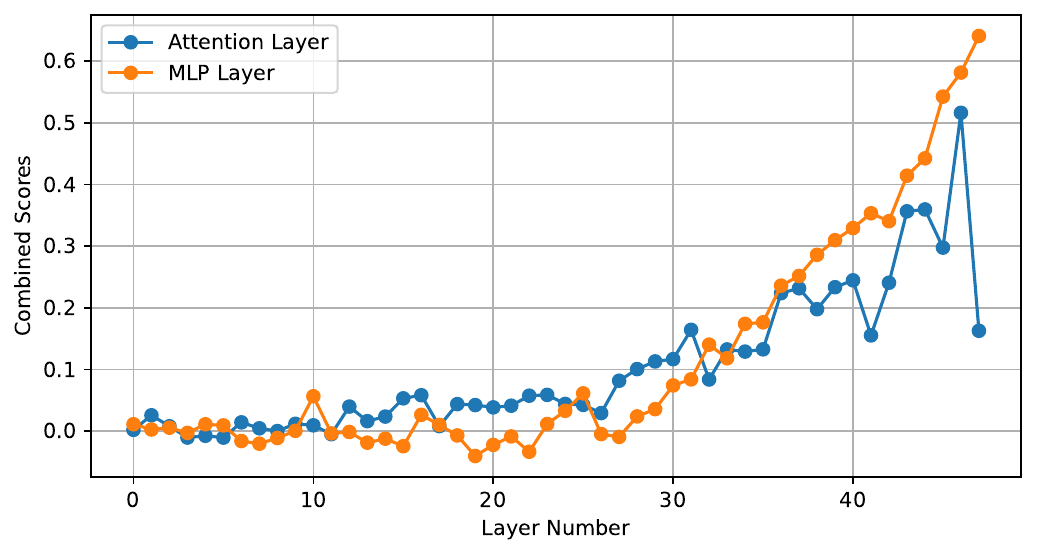}
        \caption*{Qwen2.5 (14B)}
    \end{subfigure}
    \begin{subfigure}{0.32\linewidth}
        \includegraphics[width=\linewidth]{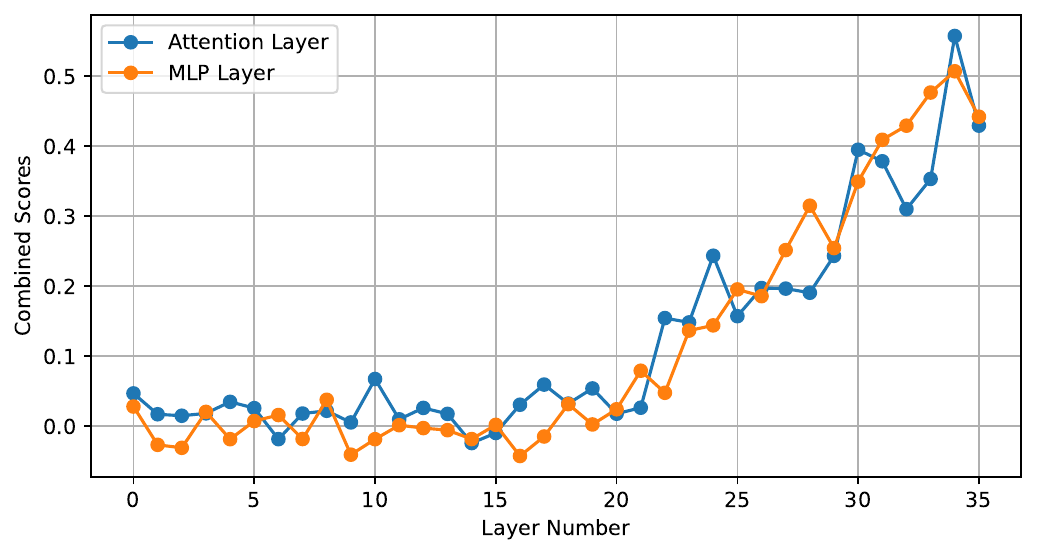}
        \caption*{Qwen3 (4B)}
    \end{subfigure}
    \begin{subfigure}{0.32\linewidth}
        \includegraphics[width=\linewidth]{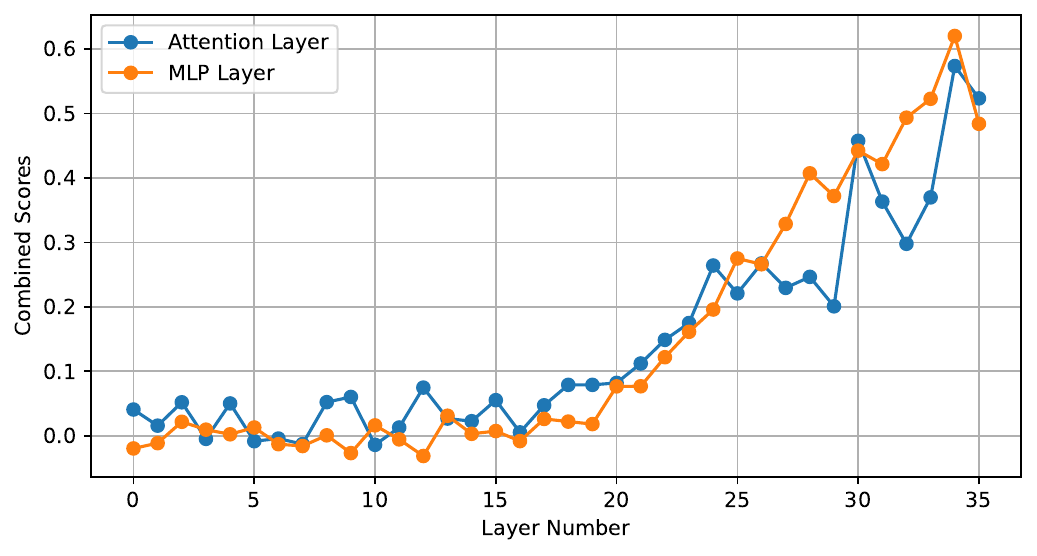}
        \caption*{Qwen3 (8B)}
    \end{subfigure}
    \begin{subfigure}{0.32\linewidth}
        \includegraphics[width=\linewidth]{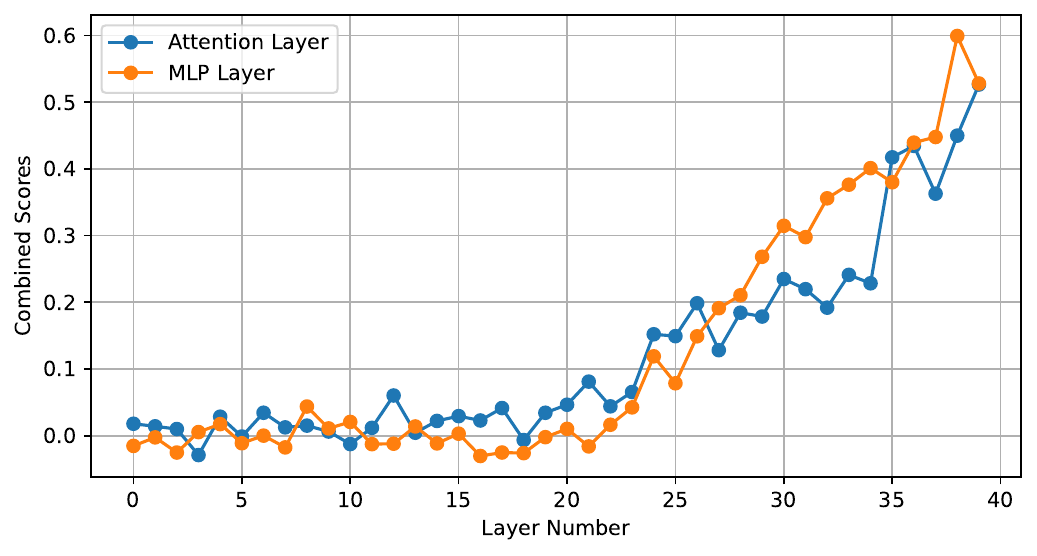}
        \caption*{Qwen3 (14B)}
    \end{subfigure}
    
    \caption{Comparison of combined scores between MLP and attention sublayers across different LLMs.}\label{fig:layer_type_comparison}
\end{figure*}

In \Secref{sec:layer_selection}, we mention that we treat both MLP and attention sublayers as ``layers'' for selection. This is because both types of sublayers contribute to the model's internal representations and decision-making processes. Modifying either type can influence the model's behavior regarding safety alignment.
Besides, we conduct preliminary experiments to compare the combined score (Eq.~\ref{eq:combined_score}) distributions of MLP and attention sublayers. The results are presented in \Figref{fig:layer_type_comparison}. The results show that both MLP and attention sublayers exhibit similar contribution score distributions across different LLMs. Later layers tend to have higher contribution scores, indicating their greater relevance to safety-related decisions. Therefore, we treat both MLP and attention sublayers equally in our layer selection process.

\section{Additional Instructions on General Ability Datasets}\label{app:additional_instructions_on_datasets}
In this section, we provide detailed instructions on the general ability datasets used in our experiments.
\begin{itemize}
    \item \textbf{Corpus of Linguistic Acceptability (COLA)} \citep{warstadt2018neural} is a dataset for evaluating the grammatical acceptability of sentences. Each sample consists of a sentence and a binary label indicating whether the sentence is grammatically acceptable or not.
    \item \textbf{Multi-Genre Natural Language Inference (MNLI)} \citep{williams2018broad} is a large-scale dataset for natural language inference. Each sample consists of a pair of sentences annotated with textual entailment labels.
    \item \textbf{Recognizing Textual Entailment (RTE)} \citep{bentivogli2009fifth} is a dataset for evaluating the ability of models to recognize textual entailment. Each sample consists of a pair of sentences where one sentence is the premise and the other is the hypothesis. 
    \item \textbf{Microsoft Research Paraphrase Corpus (MRPC)} \citep{dolan2005automatically} is a dataset for evaluating the ability of models to recognize paraphrases. Each sample consists of a pair of sentences extracted from online news sources, with human annotations indicating whether each pair is semantically equivalent or not.
    \item \textbf{Stanford Sentiment Treebank (SST)} \citep{socher-etal-2013-recursive} is a dataset for sentiment analysis. Each sample consists of a sentence and a binary label indicating whether the sentiment of the sentence is positive or negative.
    \item \textbf{GSM8K} \citep{cobbe2021gsm8k} is a dataset for evaluating the mathematical reasoning ability of models. Each sample consists of a math word problem and its corresponding solution. 
\end{itemize}

\section{Details about Experimental Setup}\label{app:experimental_setup}
\begin{table*}[ht]
\belowrulesep=0pt
\aboverulesep=0pt
  \centering
  \caption{Selected layer numbers of different models in \method.}
  \resizebox{\linewidth}{!}{
    \begin{tabular}{l|c|c|c|c|c|c|c|c|c|c|c|c}
    \toprule
    Model & \multicolumn{1}{c|}{Llama-3.1} & \multicolumn{1}{c|}{Gemma-2} & \multicolumn{1}{c|}{Mistral-v0.3} & \multicolumn{1}{c|}{Phi-3.5} & \multicolumn{2}{c|}{Phi-4} & \multicolumn{3}{c|}{Qwen2.5} & \multicolumn{3}{c}{Qwen3} \\
    \midrule
    Size  & \multicolumn{1}{c|}{8B} & \multicolumn{1}{c|}{9B} & \multicolumn{1}{c|}{7B} & \multicolumn{1}{c|}{4B} & \multicolumn{1}{c|}{4B} & \multicolumn{1}{c|}{15B} & \multicolumn{1}{c|}{3B} & \multicolumn{1}{c|}{7B} & \multicolumn{1}{c|}{14B} & \multicolumn{1}{c|}{4B} & \multicolumn{1}{c|}{8B} & \multicolumn{1}{c}{14B} \\
    \midrule
    \# Selected Layers &    42   &    60   &    42   &    30   &    48   &   54    &    60   &     36  &    54   &    48   &     60  &  48\\
    \bottomrule
    \end{tabular}%
  }
  \label{tab:layer_selection_numbers}%
\end{table*}%

We implement \method{} with max iteration number $T=30$. The final modified model is obtained by selecting the best model on the validation set during the iterations.
The numbers of selected layers $L_{select}$ of each model are shown in \tableref{tab:layer_selection_numbers}.
For the SVM-based vector identification, we use the default regularization parameter $C=1.0$ from scikit-learn~\citep{pedregosa2011scikit}.
For baseline methods, we follow the original papers and use the default hyperparameters.
If the original papers do not provide hyperparameter settings for certain models, we transfer the hyperparameters from similar models (e.g., models with the same architecture or in the same family).
All experiments are conducted on 2$\times$80 GB A100 GPUs. We use the default generation configurations in Hugging Face Transformers\footnote{\url{https://huggingface.co/docs/transformers/index}} for base LLMs during inference.
The temperature parameters of all models are set to 0.0 to ensure deterministic outputs.
We use a fixed random seed of 42 for reproducibility across all experiments.

\section{Details on Judge Model Selection}\label{app:judge_model}
\begin{figure}
    \centering
    \includegraphics[width=0.95\linewidth]{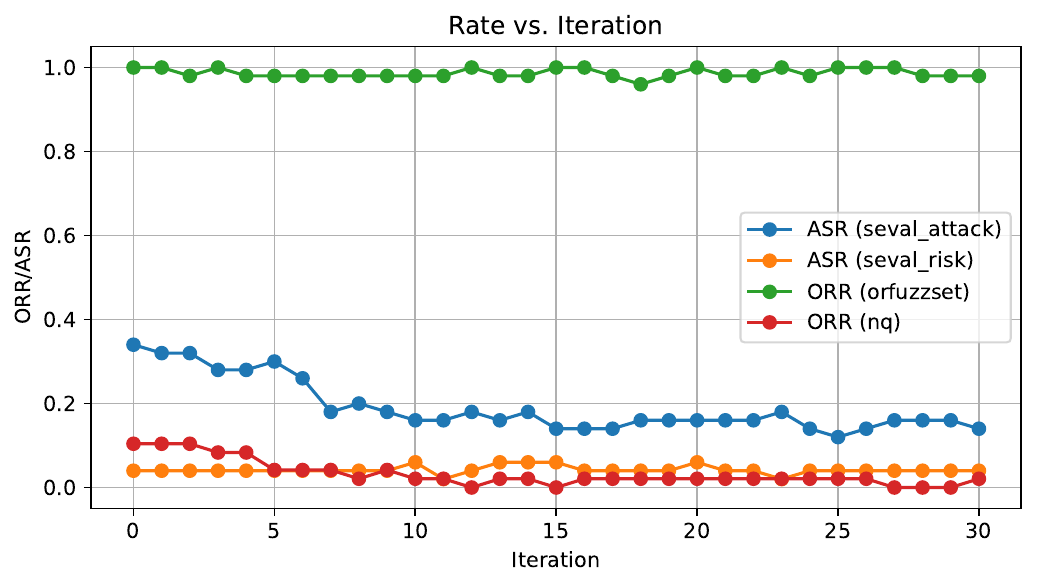}
    \caption{An example of evaluation results with combined judge models.}\label{fig:judge_example}
\end{figure}
As far as we know, Qwen3-Guard-Gen-8B~\citep{zhao2025qwen3guard} is currently the only open-source LLM specifically designed to evaluate jailbreak and over-refusal behaviors.
We also considered combining multiple judge models to realize the evaluation (e.g., LlamaGuard 3~\citep{chi2024llama} for jailbreak and OR-Judge~\citep{zhang2025orfuzzfuzzingotherside} for over-refusal).
However, due to their different judgement criteria, combining multiple judge models may lead to inconsistent evaluations.
As a result, \method{} will find incorrect vectors to align, leading to suboptimal performance.
\Figref{fig:judge_example} shows an example of such inconsistent evaluations.
The ORR evaluated by OR-Judge reaches 100\% due to the inconsistency between the two judge models.
Therefore, we choose Qwen3-Guard-Gen-8B as the sole judge model for a consistent evaluation of both jailbreak and over-refusal behaviors.

\section{Detailed Results on Iteration Number}\label{app:detailed_results_on_iteration_number}
\begin{figure*}
    \begin{subfigure}{0.32\linewidth}
        \centering
        \includegraphics[width=\linewidth]{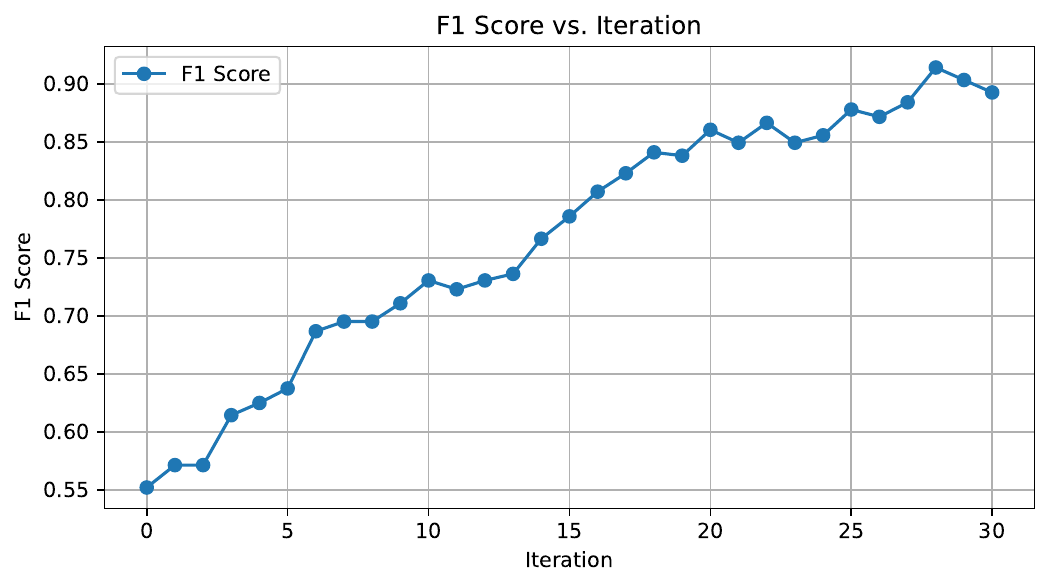}
        \caption*{gemma-2 (9B)}
    \end{subfigure}
    \begin{subfigure}{0.32\linewidth}
        \centering
        \includegraphics[width=\linewidth]{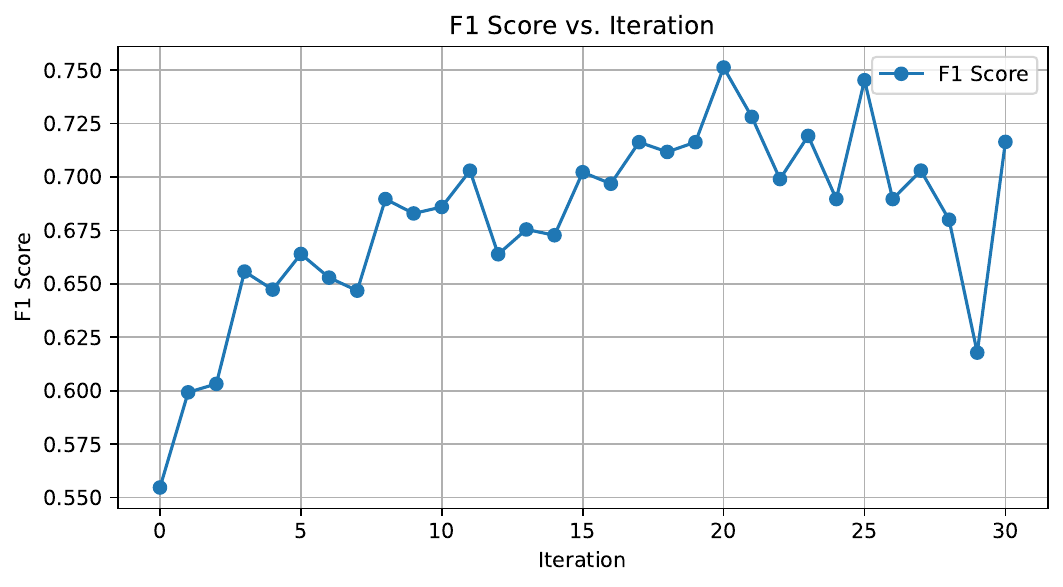}
        \caption*{Mistral-v0.3 (7B)}
    \end{subfigure}
    \begin{subfigure}{0.32\linewidth}
        \centering
        \includegraphics[width=\linewidth]{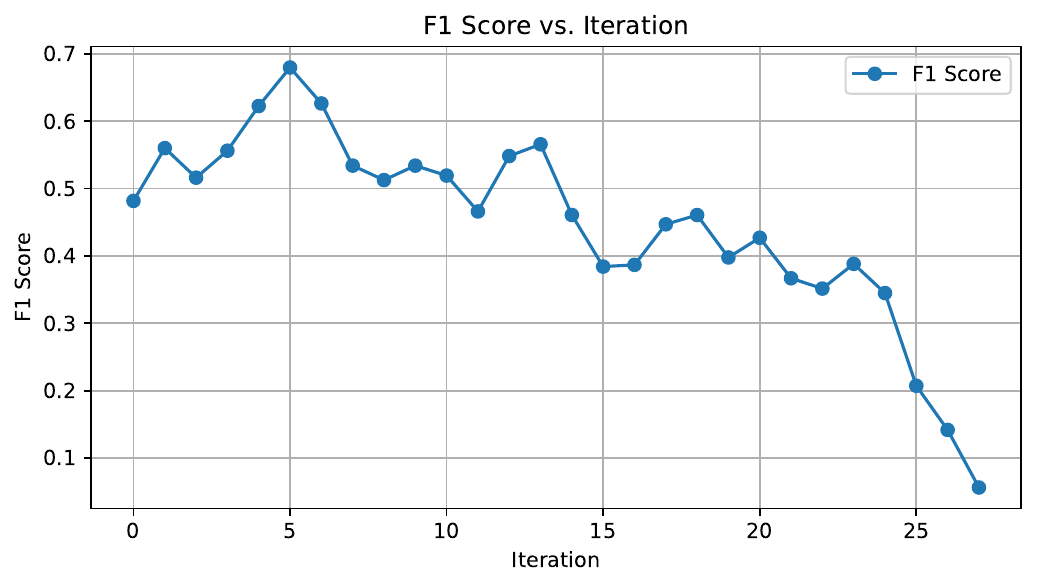}
        \caption*{Phi-3.5 (4B)}
    \end{subfigure}
    \begin{subfigure}{0.32\linewidth}
        \centering
        \includegraphics[width=\linewidth]{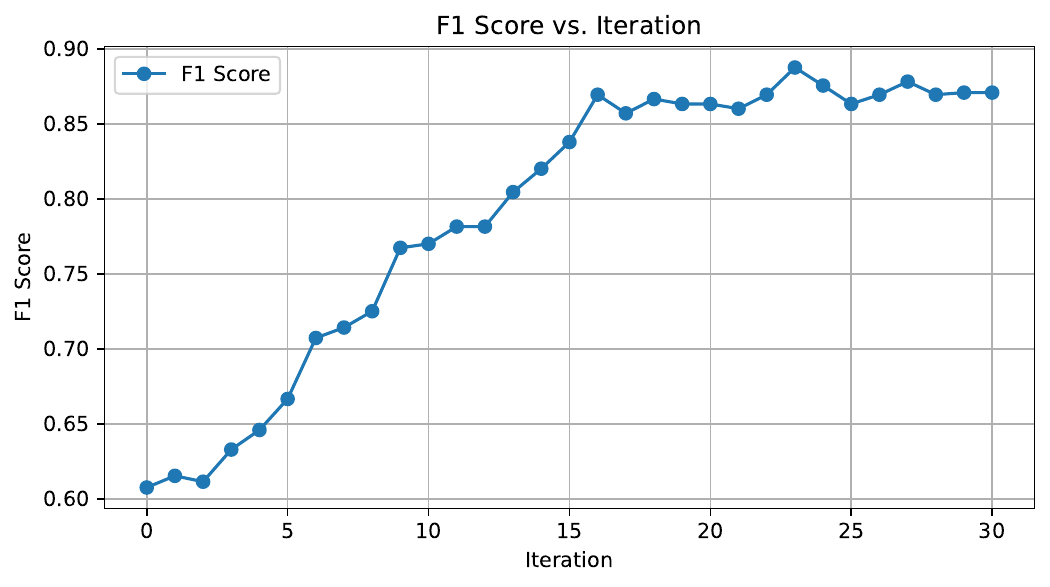}
        \caption*{Phi-4 (15B)}
    \end{subfigure}
    \begin{subfigure}{0.32\linewidth}
        \centering
        \includegraphics[width=\linewidth]{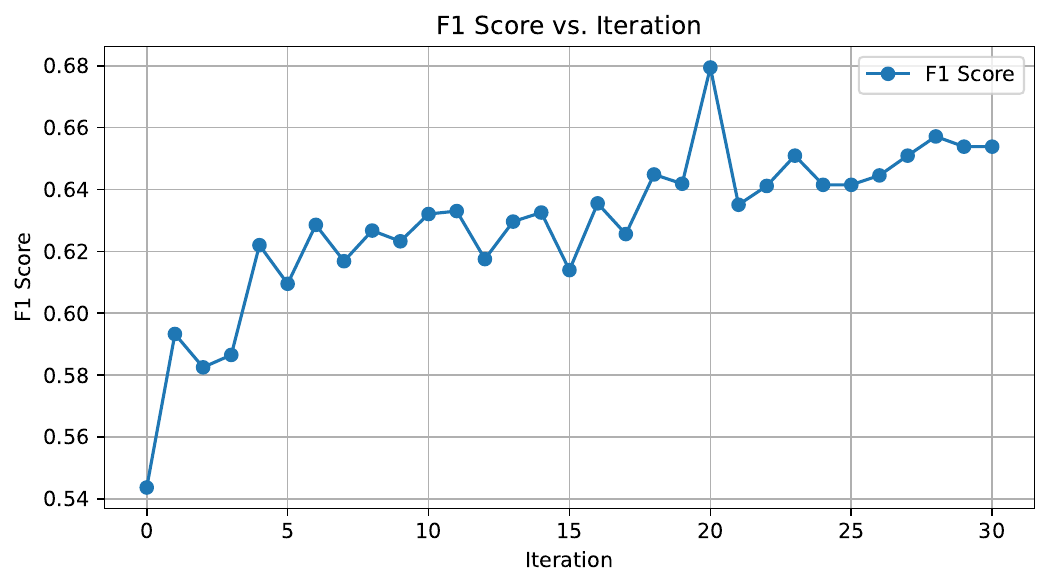}
        \caption*{Qwen2.5 (3B)}
    \end{subfigure}
    \begin{subfigure}{0.32\linewidth}
        \centering
        \includegraphics[width=\linewidth]{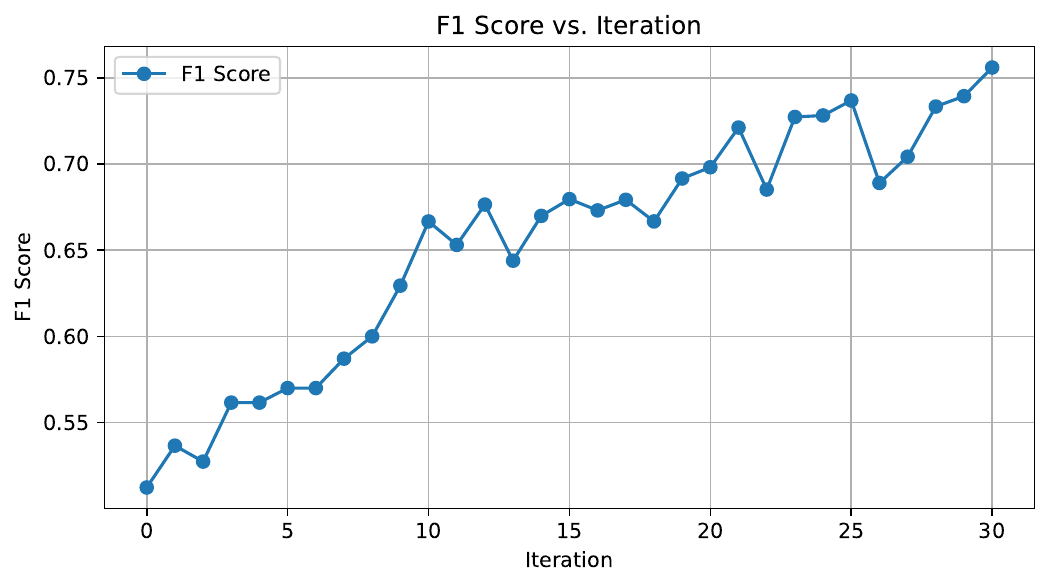}
        \caption*{Qwen2.5 (7B)}
    \end{subfigure}
    \begin{subfigure}{0.32\linewidth}
        \centering
        \includegraphics[width=\linewidth]{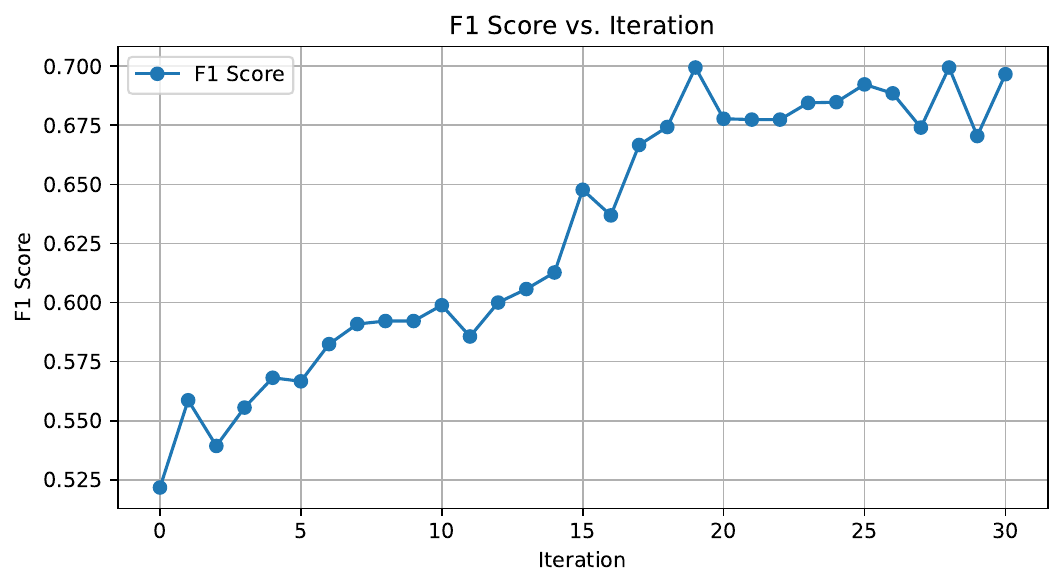}
        \caption*{Qwen2.5 (14B)}
    \end{subfigure}
    \begin{subfigure}{0.32\linewidth}
        \centering
        \includegraphics[width=\linewidth]{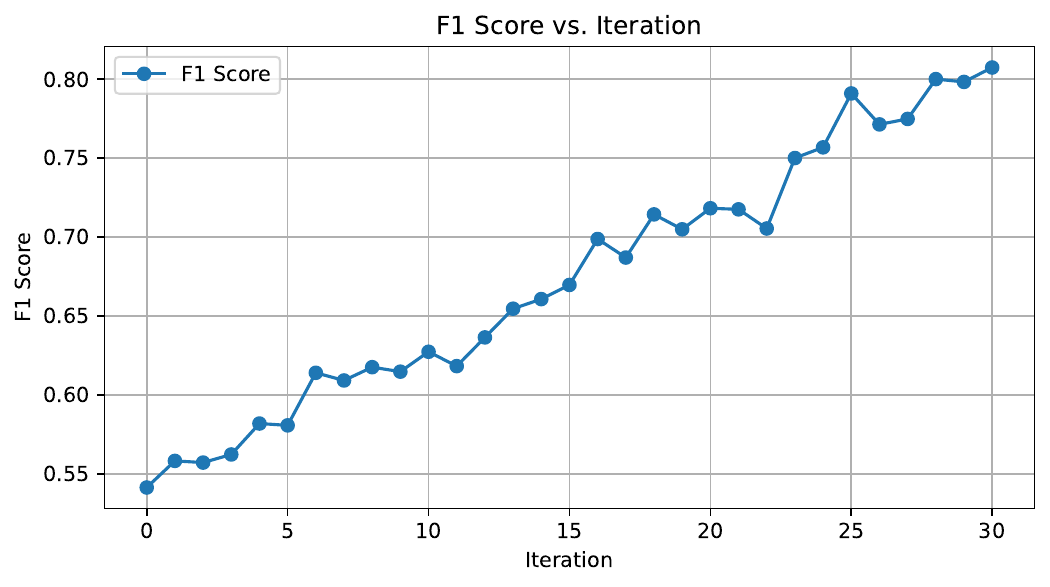}
        \caption*{Qwen3 (4B)}
    \end{subfigure}
    \begin{subfigure}{0.32\linewidth}
        \centering
        \includegraphics[width=\linewidth]{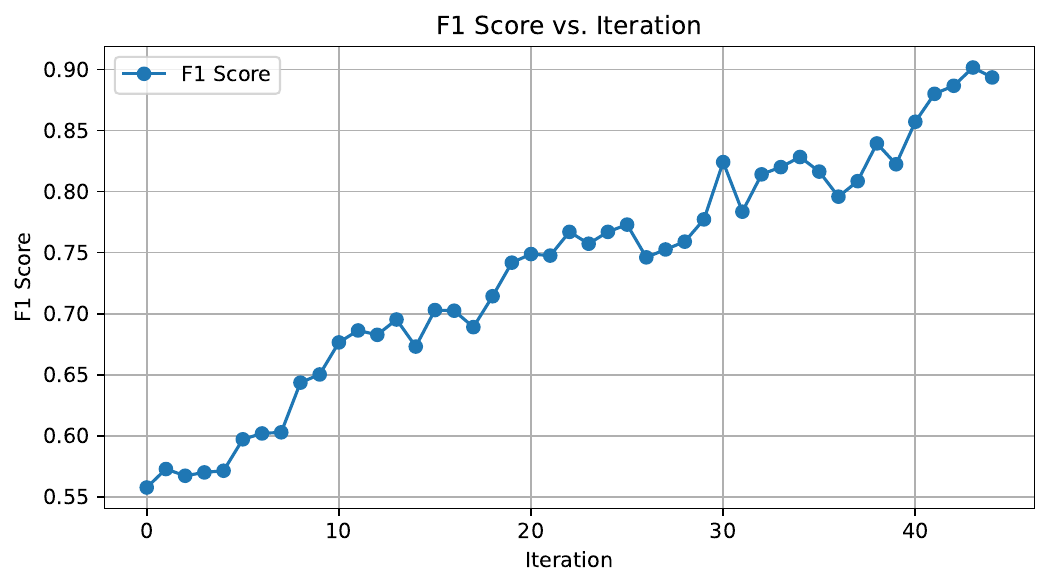}
        \caption*{Qwen3 (8B)}
    \end{subfigure}
    \caption{Detailed results on the impact of iteration number of each LLM.}\label{fig:detailed_results_on_iteration_number}
\end{figure*}

The detailed results on the impact of iteration number of each LLM are shown in \Figref{fig:detailed_results_on_iteration_number}.

\section{Transferability Experiments}\label{app:transferability_experiments}
To evaluate the transferability of \method{}, we assess how well the vector alignment learned on the training datasets generalizes to unseen datasets.
We evaluate the modified models on three additional jailbreak datasets (XSTest-Toxic~\citep{röttger2024xstesttestsuiteidentifying}, OR-Bench-Toxic~\citep{cui2025orbenchoverrefusalbenchmarklarge}, and AdvBench~\citep{zou2023universal}) and two over-refusal datasets (XSTest~\citep{röttger2024xstesttestsuiteidentifying} and OR-Bench~\citep{cui2025orbenchoverrefusalbenchmarklarge}) that are not included in the training set.
The results are shown in \tableref{tab:transferability}.

\begin{table*}[htbp]
    \belowrulesep=0pt
    \aboverulesep=0pt
  \centering
  \caption{Transferability results on unseen datasets.}
  \resizebox{\linewidth}{!}{
    \begin{tabular}{c|c|l|r|r|r|r|r|r||c|c|l|r|r|r|r|r|r}
    \toprule
    Model & Size  & Method & \multicolumn{1}{p{4.055em}|}{AdvBench\newline{}ASR↓} & \multicolumn{1}{p{4.055em}|}{OR-Bench-Toxic\newline{}ASR↓} & \multicolumn{1}{p{4.055em}|}{XSTest-Toxic\newline{}ASR↓} & \multicolumn{1}{p{4.055em}|}{OR-Bench\newline{}ORR↓} & \multicolumn{1}{p{4.055em}|}{XSTest\newline{}ORR↓} & \multicolumn{1}{p{4.055em}||}{Final\newline{}F1↑} & Model & Size  & Method & \multicolumn{1}{p{4.055em}|}{AdvBench\newline{}ASR↓} & \multicolumn{1}{p{4.055em}|}{OR-Bench-Toxic\newline{}ASR↓} & \multicolumn{1}{p{4.055em}|}{XSTest-Toxic\newline{}ASR↓} & \multicolumn{1}{p{4.055em}|}{OR-Bench\newline{}ORR↓} & \multicolumn{1}{p{4.055em}|}{XSTest\newline{}ORR↓} & \multicolumn{1}{p{4.055em}}{Final\newline{}F1↑} \\
    \midrule
    \multirow{6}{*}{Llama-3.1} & \multirow{6}{*}{8B} & Original & 0.58\% & 3.05\% & \textbf{0.00\%} & 48.22\% & 16.57\% & 0.7077  & \multirow{18}{*}{Qwen2.5} & \multirow{6}{*}{3B} & Original & 0.19\% & 2.29\% & 0.57\% & 55.42\% & 19.34\% & 0.6522  \\
\cmidrule{3-9}\cmidrule{12-18}          &       & AlphaSteer+ & 0.58\% & 3.05\% & \textbf{0.00\%} & 48.67\% & 17.13\% & 0.7038  &       &       & AlphaSteer+ & \textbf{0.00\%} & 2.44\% & \textbf{0.00\%} & 53.53\% & 21.55\% & 0.6649  \\
\cmidrule{3-9}\cmidrule{12-18}          &       & AlphaSteer & 0.19\% & 1.37\% & \textbf{0.00\%} & 61.87\% & 27.07\% & 0.5921  &       &       & AlphaSteer & 0.19\% & 1.53\% & \textbf{0.00\%} & 59.97\% & 21.55\% & 0.6144  \\
\cmidrule{3-9}\cmidrule{12-18}          &       & Steer & \textbf{0.00\%} & \textbf{0.15\%} & \textbf{0.00\%} & 85.60\% & 49.17\% & 0.3163  &       &       & Steer & \textbf{0.00\%} & \textbf{0.15\%} & \textbf{0.00\%} & 85.97\% & 37.57\% & 0.3313  \\
\cmidrule{3-9}\cmidrule{12-18}          &       & SCANS & 0.19\% & 1.37\% & \textbf{0.00\%} & 72.48\% & 26.52\% & 0.4945  &       &       & SCANS & 6.35\% & 16.64\% & 1.15\% & \textbf{40.11\%} & \textbf{13.81\%} & \textbf{0.7305} \\
\cmidrule{3-9}\cmidrule{12-18}          &       & Modified & 7.69\% & 5.80\% & 4.02\% & \textbf{46.85\%} & \textbf{6.63\%} & \textbf{0.7088} &       &       & Modified & 0.38\% & 4.12\% & 0.57\% & 54.21\% & 21.55\% & 0.6555  \\
\cmidrule{1-9}\cmidrule{11-18}    \multirow{6}{*}{gemma-2} & \multirow{6}{*}{9B} & Original & 0.58\% & 1.98\% & \textbf{0.00\%} & 80.52\% & 28.73\% & 0.4059  &       & \multirow{6}{*}{7B} & Original & \textbf{0.38\%} & 6.72\% & \textbf{0.00\%} & 24.64\% & 8.84\% & 0.8569  \\
\cmidrule{3-9}\cmidrule{12-18}          &       & AlphaSteer+ & 0.77\% & 0.92\% & 0.57\% & 81.58\% & \textbf{26.52\%} & 0.3985  &       &       & AlphaSteer+ & 0.77\% & 6.11\% & \textbf{0.00\%} & 24.87\% & 8.84\% & 0.8563  \\
\cmidrule{3-9}\cmidrule{12-18}          &       & AlphaSteer & \textbf{0.00\%} & 0.46\% & \textbf{0.00\%} & 88.86\% & 28.73\% & 0.3103  &       &       & AlphaSteer & 3.08\% & \textbf{3.66\%} & \textbf{0.00\%} & 34.50\% & 9.39\% & 0.8006  \\
\cmidrule{3-9}\cmidrule{12-18}          &       & Steer & \textbf{0.00\%} & \textbf{0.31\%} & \textbf{0.00\%} & 94.16\% & 56.35\% & 0.1882  &       &       & Steer & 8.65\% & 5.50\% & \textbf{0.00\%} & 61.03\% & 29.28\% & 0.5776  \\
\cmidrule{3-9}\cmidrule{12-18}          &       & SCANS & 3.08\% & 5.50\% & 0.57\% & \textbf{59.82\%} & 29.28\% & \textbf{0.5952} &       &       & SCANS & 1.54\% & 6.87\% & 1.72\% & 40.56\% & 11.60\% & 0.7552  \\
\cmidrule{3-9}\cmidrule{12-18}          &       & Modified & 2.88\% & 1.68\% & \textbf{0.00\%} & 82.41\% & 29.83\% & 0.3809  &       &       & Modified & 3.65\% & 9.92\% & \textbf{0.00\%} & \textbf{19.94\%} & \textbf{7.73\%} & \textbf{0.8714} \\
\cmidrule{1-9}\cmidrule{11-18}    \multirow{6}{*}{Mistral-v0.3} & \multirow{6}{*}{7B} & Original & 54.42\% & 48.70\% & 16.67\% & \textbf{7.28\%} & \textbf{3.87\%} & 0.7920  &       & \multirow{6}{*}{14B} & Original & \textbf{0.00\%} & 4.58\% & \textbf{0.00\%} & 21.15\% & 8.29\% & 0.8816  \\
\cmidrule{3-9}\cmidrule{12-18}          &       & AlphaSteer+ & 52.12\% & 48.85\% & 16.67\% & 7.35\% & 4.42\% & 0.7937  &       &       & AlphaSteer+ & 0.19\% & 5.34\% & \textbf{0.00\%} & 20.77\% & 8.29\% & 0.8817  \\
\cmidrule{3-9}\cmidrule{12-18}          &       & AlphaSteer & 45.38\% & 48.09\% & 14.37\% & 7.66\% & 4.97\% & \textbf{0.8021} &       &       & AlphaSteer & \textbf{0.00\%} & 3.21\% & \textbf{0.00\%} & 26.31\% & 9.39\% & 0.8551  \\
\cmidrule{3-9}\cmidrule{12-18}          &       & Steer & 42.69\% & 40.92\% & 5.75\% & 11.75\% & 11.05\% & 0.7970  &       &       & Steer & \textbf{0.00\%} & \textbf{0.15\%} & \textbf{0.00\%} & 57.01\% & 16.02\% & 0.6477  \\
\cmidrule{3-9}\cmidrule{12-18}          &       & SCANS & 74.42\% & 41.98\% & 22.99\% & 18.57\% & 7.73\% & 0.7209  &       &       & SCANS & 7.88\% & 16.95\% & 2.30\% & 30.40\% & 19.34\% & 0.7824  \\
\cmidrule{3-9}\cmidrule{12-18}          &       & Modified & \textbf{27.31\%} & \textbf{17.25\%} & \textbf{2.87\%} & 37.30\% & 10.50\% & 0.7195  &       &       & Modified & 0.77\% & 5.04\% & \textbf{0.00\%} & \textbf{17.74\%} & \textbf{6.08\%} & \textbf{0.8990} \\
    \midrule
    \multirow{6}{*}{Phi-3.5} & \multirow{6}{*}{4B} & Original & 2.12\% & 4.89\% & 1.72\% & 45.49\% & 13.26\% & 0.7234  & \multirow{18}{*}{Qwen3} & \multirow{6}{*}{4B} & Original & 0.96\% & 4.73\% & 0.57\% & 44.35\% & 6.63\% & 0.7402  \\
\cmidrule{3-9}\cmidrule{12-18}          &       & AlphaSteer+ & \textbf{1.15\%} & 4.43\% & 1.15\% & 43.90\% & 13.81\% & \textbf{0.7365} &       &       & AlphaSteer+ & 7.88\% & 24.58\% & 4.02\% & \textbf{21.08\%} & \textbf{4.42\%} & \textbf{0.8307} \\
\cmidrule{3-9}\cmidrule{12-18}          &       & AlphaSteer & \textbf{1.15\%} & 3.66\% & 1.15\% & 48.98\% & 13.81\% & 0.7022  &       &       & AlphaSteer & 33.85\% & 37.10\% & 8.62\% & 22.14\% & 7.73\% & 0.7634  \\
\cmidrule{3-9}\cmidrule{12-18}          &       & Steer & \textbf{1.15\%} & 3.51\% & 1.15\% & 56.41\% & 22.10\% & 0.6373  &       &       & Steer & \textbf{0.77\%} & \textbf{1.83\%} & \textbf{0.00\%} & 66.49\% & 19.89\% & 0.5583  \\
\cmidrule{3-9}\cmidrule{12-18}          &       & SCANS & 1.54\% & \textbf{1.37\%} & 1.15\% & 79.83\% & 37.57\% & 0.3994  &       &       & SCANS & 3.65\% & 6.11\% & 0.57\% & 46.93\% & 11.60\% & 0.7107  \\
\cmidrule{3-9}\cmidrule{12-18}          &       & Modified & 6.54\% & 6.41\% & \textbf{0.00\%} & \textbf{43.21\%} & \textbf{12.71\%} & 0.7306  &       &       & Modified & 2.31\% & 4.27\% & 0.57\% & 36.69\% & 7.73\% & 0.7880  \\
\cmidrule{1-9}\cmidrule{11-18}    \multirow{12}{*}{Phi-4} & \multirow{6}{*}{4B} & Original & 0.58\% & 2.14\% & \textbf{0.00\%} & 58.83\% & 17.68\% & 0.6265  &       & \multirow{6}{*}{8B} & Original & 0.96\% & 3.21\% & 1.15\% & 44.12\% & 9.39\% & 0.7419  \\
\cmidrule{3-9}\cmidrule{12-18}          &       & AlphaSteer+ & \textbf{0.19\%} & \textbf{1.53\%} & \textbf{0.00\%} & 59.97\% & 18.23\% & 0.6182  &       &       & AlphaSteer+ & 7.31\% & 14.20\% & 7.47\% & 62.70\% & 22.10\% & 0.5560  \\
\cmidrule{3-9}\cmidrule{12-18}          &       & AlphaSteer & \textbf{0.19\%} & \textbf{1.53\%} & \textbf{0.00\%} & 55.42\% & 18.23\% & 0.6551  &       &       & AlphaSteer & \textbf{0.58\%} & 4.89\% & 1.15\% & \textbf{34.65\%} & \textbf{7.18\%} & \textbf{0.8025} \\
\cmidrule{3-9}\cmidrule{12-18}          &       & Steer & \textbf{0.19\%} & 2.60\% & \textbf{0.00\%} & 46.70\% & 16.57\% & 0.7201  &       &       & Steer & 0.96\% & 4.12\% & 0.57\% & 40.03\% & 9.39\% & 0.7677  \\
\cmidrule{3-9}\cmidrule{12-18}          &       & SCANS & 10.19\% & 12.98\% & 0.57\% & 36.01\% & \textbf{8.84\%} & 0.7621  &       &       & SCANS & \textbf{0.58\%} & \textbf{2.29\%} & \textbf{0.00\%} & 70.96\% & 20.99\% & 0.5147  \\
\cmidrule{3-9}\cmidrule{12-18}          &       & Modified & 0.58\% & 7.63\% & 1.15\% & \textbf{18.73\%} & 11.60\% & \textbf{0.8841} &       &       & Modified & 0.96\% & 3.82\% & \textbf{0.00\%} & 40.71\% & 8.29\% & 0.7651  \\
\cmidrule{2-9}\cmidrule{11-18}          & \multirow{6}{*}{15B} & Original & 0.19\% & 3.97\% & \textbf{0.00\%} & 72.71\% & 17.13\% & 0.5007  &       & \multirow{6}{*}{14B} & Original & 0.19\% & 4.43\% & 0.57\% & 40.33\% & 7.18\% & 0.7683  \\
\cmidrule{3-9}\cmidrule{12-18}          &       & AlphaSteer+ & \textbf{0.00\%} & 3.97\% & \textbf{0.00\%} & 72.63\% & 15.47\% & 0.5039  &       &       & AlphaSteer+ & 5.96\% & 22.44\% & 3.45\% & \textbf{13.65\%} & 6.63\% & \textbf{0.8743} \\
\cmidrule{3-9}\cmidrule{12-18}          &       & AlphaSteer & \textbf{0.00\%} & 3.36\% & \textbf{0.00\%} & 75.97\% & 16.57\% & 0.4704  &       &       & AlphaSteer & 0.77\% & 12.21\% & 0.57\% & 20.09\% & \textbf{5.52\%} & 0.8719  \\
\cmidrule{3-9}\cmidrule{12-18}          &       & Steer & \textbf{0.00\%} & \textbf{2.29\%} & \textbf{0.00\%} & 84.23\% & 20.99\% & 0.3762  &       &       & Steer & \textbf{0.00\%} & \textbf{0.31\%} & 0.57\% & 72.33\% & 20.99\% & 0.5052  \\
\cmidrule{3-9}\cmidrule{12-18}          &       & SCANS & 1.35\% & 5.95\% & \textbf{0.00\%} & 67.78\% & \textbf{11.60\%} & 0.5490  &       &       & SCANS & 29.04\% & 26.11\% & 9.20\% & 35.94\% & 25.41\% & 0.6955  \\
\cmidrule{3-9}\cmidrule{12-18}          &       & Modified & 0.77\% & 3.82\% & \textbf{0.00\%} & \textbf{53.37\%} & \textbf{11.60\%} & \textbf{0.6727} &       &       & Modified & 4.81\% & 3.05\% & \textbf{0.00\%} & 51.86\% & 11.05\% & 0.6801  \\
\bottomrule
    \end{tabular}%
 }
  \label{tab:transferability}%
\end{table*}%

The results show that the performance of \method{} on unseen datasets varies across different models.
While \method{} maintains reasonable safety alignment on most unseen datasets, the performance degradation compared to the training datasets indicates that further research is needed to improve the generalization of vector steering methods.

\end{document}